%% file: main.tex
\def\notes{0}
\documentclass{article}
\usepackage{iclr2027_conference,times}

\input{math_commands.tex}

\usepackage[utf8]{inputenc}
\usepackage[T1]{fontenc}
\usepackage{amssymb}
\usepackage{mathtools}
\usepackage{graphicx}
\usepackage{booktabs}
\usepackage{subcaption}
\usepackage{xcolor}
\usepackage{enumitem}
\usepackage{titletoc}
\usepackage[most]{tcolorbox}
\newtcolorbox[auto counter]{claimbox}[2][]{colback=blue!4,colframe=blue!55!black,fonttitle=\bfseries,title={Box~\thetcbcounter: #2},#1}
\usepackage{xspace}
\usepackage[hyphens]{url}
\usepackage[
  colorlinks=true,
  allcolors=blue,
  pdfborder={0 0 0}
]{hyperref}
\usepackage{placeins}

\input{macros.tex}

\newcommand{\custompar}[1]{\smallskip\noindent\textbf{#1.}\ }

\newcommand{\customparnodot}[1]{\smallskip\noindent\textbf{#1}\ }

\title{\fontsize{16.25}{19}\selectfont Extractable Memorization From First Principles\looseness=-1}

\usepackage{etoolbox}
  \makeatletter
  \patchcmd{\@maketitle}{{\LARGE\sc \@title\par}}{{\LARGE\sc \@title\par}\centering}{}{%
    \PackageWarning{iclrfix}{author-centering patch failed}}

  \patchcmd{\@maketitle}
    {\begin{tabular}[t]{l}\bf\rule{\z@}{24pt}\@author}
    {\begin{tabular}[t]{c}\bf\rule{\z@}{24pt}\@author}
    {}{\PackageWarning{iclrfix}{author column patch failed}}
 \makeatother

\author{\textbf{A. Feder Cooper$^{1,2}$\thanks{Corresponding author: \texttt{a.feder.cooper@yale.edu}} \qquad Marika Swanberg$^{3}$ \qquad Jamie Hayes$^{4}$ \qquad Lea Duesterwald$^{5}$} \vspace{.1cm}\\
\textbf{Chris De Sa$^{5}$ \quad \textbf{Georgios Kaissis}$^{6}$ \quad Daniel E. Ho$^{2}$ \qquad Mark A. Lemley$^{2}$ \quad Percy Liang$^{2}$}\vspace{.35cm}\\ 
$^{1}$Yale University \qquad $^{2}$Stanford University \qquad $^{3}$Google Research\vspace{0.1cm}\\
$^{4}$Google DeepMind \qquad 
$^{5}$Cornell University \qquad
$^{6}$Hasso-Plattner-Institut 
}

\iclrfinalcopy

\begin{document}

\maketitle

\vspace{-.4cm}
\begin{abstract}
\vspace{-.15cm}
Recent work on extractable memorization in language models suffers from two contrasting validity problems.
Some studies overstate extraction, for example, by using sequences too short to distinguish memorization from predictability.
Others imply that extraction is unreliable evidence because models can reproduce real-world text they weren't explicitly trained on. 
Both overlook what makes a valid extraction claim:
the model must generate a training sequence with high enough probability to indicate memorization. 
To determine what's high enough, one has to perform a matched comparison, measuring  generation probabilities of training and comparable non-training sequences.
Since non-training sequences can't have been memorized, their probabilities are a baseline for predictability;  
exceeding this baseline is evidence of memorization.
We formalize matched comparisons with (1) a conformal test calibrated to a chosen false-positive rate when sequences are sampled from populations, and (2) a single-document census that calibrates against a matched non-training document.
Matched comparisons enable rigorous, calibrated memorization claims and clarify where prior setups have validity issues. 
On Wikipedia, OLMo~2 32B reproduces non-training $10$-token suffixes roughly $24\%$ as often as training ones: 
that share reflects false positives, not memorization.
For Llama~3.1 70B on books, calibrated census thresholds reach as low as $10^{-27}$, supporting memorization claims for sequences no feasible sampling budget would extract.
We therefore refine ``extractable memorization'' to require both a valid memorization claim and near-certain generation within a realistic budget.\looseness=-1
\end{abstract}

\input{section/iclr/100-intro}
\input{section/iclr/200-prelim}
\input{section/iclr/300-conformal}
\input{section/iclr/400-population}
\input{section/iclr/500-census}
\input{section/iclr/600-definition}
\input{section/iclr/700-takeaways}

\clearpage
\subsection*{Acknowledgments}

This work was supported by a Stanford HAI Seed Research Grant.

We thank the Stanford Research Computing Center for providing computational resources and support
that contributed to these research results.

We also thank Katherine Lee for useful discussions and feedback on earlier versions of this draft. 

\subsection*{AI use statement}

In this work, we used generative AI tools for identifying typos in the manuscript, soliciting feedback on plots concerning accessibility for readers with visual impairments, for a collaborative (human-led) agent skill we developed for decontaminating nulls (see \S\ref{app:census}), and checking for duplicated or overloaded symbols across the mathematical notation. 

We haven't used generative AI tools for the following tasks that require disclosure: generate synthetic data sets, help develop theoretical models or conceptual frameworks, formulate mathematical claims, provide critical ingredients for proving mathematical claims, assist in the writing of proofs, propose or refine hypotheses, design or provide feedback on research  methodology or experiments, assist with translation, clean and reformat dataset, support qualitative and thematic data analysis, interpret results. 
With the exception of the agent skill mentioned above, we have not used generative AI tools to implement methods.

We haven't used generative AI for the following tasks with recommended disclosure:
formulate questions for surveys or interviews, create or modify scientific figures or images, suggest experimental parameters, create or edit software code, creation of artifacts, draft parts of a research paper, transcribe recordings of research material, summarize or analyze existing literature, discover research topics or identify gaps, brainstorming, sourcing/searching for information, identify relevant literature, format references, suggest a structure for a research paper, propose a title or keywords for a research paper.

We have reviewed all AI-assisted work. 
For example, for typo checking, we created a git diff between the human-written paper and the AI output, and manually verified each change before implementing it in the draft. 
The agent skill we developed for decontaminating nulls is iterative and requires human inspection and verification at every single step. 

We take responsibility for the final content of this work, including text, claims or artifacts produced with the aid of generative AI.

\subsection*{Ethics statement}

Scientific papers on memorization and extraction are now frequently cited in litigation.
Claims that cite this literature don't always engage with whether the specific evidence follows a valid extraction procedure, or with whether the claim being made overstates what that evidence can support (\S\ref{sec:intro}, \S\ref{sec:takeaways}).
Our work speaks directly to this issue by revisiting extractable memorization from first principles, similar to how \citet{carlini2022membership} revisited ``Membership Inference Attacks From First Principles.'' 
We formalize what makes the generation of (near-)verbatim training data valid empirical evidence for a claim of extraction, and therefore of memorization in the model. 
We hope this work will help courts evaluating expert testimony, with respect to assessing the scientific validity of the methods used to claim extraction of memorized training data.
See \citet{lemley2026copies} for more details. 

\subsection*{Reproducibility statement}

We release the code for running the experiments in \S\ref{sec:experiments:population}, \S\ref{sec:experiments:census}, \S\ref{app:population}, \S\ref{app:census}, and \S\ref{app:extractability} at \url{https://github.com/pasta41/probabilistic-extraction-toolkit}. 
We can't distribute the books for which we ran the census experiments, as they're under copyright.
Instead, we provide links to public domain books in \texttt{.txt} format that can be used with our code base. 
For the theoretical results, we complement the brief discussion of assumptions in \S\ref{sec:test} with further details in \S\ref{app:sec:conformal}, where we provide complete proofs.
We summarize the assumptions behind all of our results in \S\ref{app:assumptions}.
More information about the project can be found at \url{https://monkey-emeritus.github.io/}.

\bibliographystyle{iclr2027_conference}
\bibliography{references}

\input{section/iclr/900-appendix}

\end{document}

%% file: math_commands.tex
\usepackage{amsmath,amsfonts,bm}

\newcommand{\newterm}[1]{{\bf #1}}

\def\eqref#1{equation~\ref{#1}}

\def\ceil#1{\lceil #1 \rceil}
\def\floor#1{\lfloor #1 \rfloor}
\def\1{\bm{1}}

\def\vp{{\bm{p}}}

\def\vs{{\bm{s}}}

\def\vv{{\bm{v}}}
\def\vw{{\bm{w}}}
\def\vx{{\bm{x}}}
\def\vy{{\bm{y}}}

\DeclareMathAlphabet{\mathsfit}{\encodingdefault}{\sfdefault}{m}{sl}
\SetMathAlphabet{\mathsfit}{bold}{\encodingdefault}{\sfdefault}{bx}{n}

\def\sB{{\mathbb{B}}}

\def\sD{{\mathbb{D}}}

\def\sM{{\mathbb{M}}}

\def\sP{{\mathbb{P}}}

\def\sS{{\mathbb{S}}}

\def\sV{{\mathbb{V}}}

\newcommand{\E}{\mathbb{E}}

\newcommand{\R}{\mathbb{R}}

\DeclareMathOperator*{\argmax}{arg\,max}

\usepackage{amsthm}

\theoremstyle{plain}
\newtheorem{theorem}{Theorem}
\newtheorem{proposition}[theorem]{Proposition}
\newtheorem{lemma}[theorem]{Lemma}
\newtheorem{corollary}[theorem]{Corollary}
\theoremstyle{definition}
\newtheorem{definition}{Definition}[section]

\theoremstyle{definition}

\theoremstyle{definition}

\theoremstyle{remark}
\newtheorem{remark}[theorem]{Remark}
\theoremstyle{plain}

%% file: macros.tex
\newcommand{\dataset}{\sD}                 
\newcommand{\promptclass}{\sP}             
\newcommand{\vocab}{\sV}                   

\newcommand{\candrv}{S}                    

\newcommand{\candseq}{\vs}                 
\newcommand{\outseq}{\vy}                  
\newcommand{\outtok}{y}                    
\newcommand{\promptseq}{\vp}               
\newcommand{\npromptseq}[1]{\promptseq^{(#1)}}    
\newcommand{\member}{\vx}                 

\newcommand{\ctrl}{\mathrm{OUT}}           
\newcommand{\ctrlrv}{W}                    

\newcommand{\params}{\theta}               
\newcommand{\dec}{\phi}                    
\newcommand{\greedy}{\mathsf{greedy}}      
\newcommand{\channel}{q_{\params,\dec}}    

\newcommand{\tol}{\varepsilon}             
\newcommand{\eventlen}{\ell}               
\newcommand{\maxlen}{A}           

\newcommand{\distance}{d}
\newcommand{\ball}[2]{\sB_{#1}(#2)}

\newcommand{\memberind}[1]{\mathsf{membership}_{\tol}(#1)}

\newcommand{\promptrule}{f_{\promptclass}}     
\newcommand{\score}{\mathsf{score}}             
\newcommand{\verify}{\mathsf{verify}}            
\newcommand{\seq}{\vv}                          
\DeclareMathOperator*{\agg}{\mathsf{agg}}       

\usepackage{xcolor}
\ifnum\notes=1

\newcommand{\ms}[1]{{\color{violet}{ \textbf{MS: #1} \xspace}}}
\newcommand{\jh}[1]{{\color{blue}{ \textbf{jh: #1} \xspace}}}

\else 
\newcommand{\ms}[1]{\color{violet} \textbf{} \xspace}

\newcommand{\jh}[1]{\color{blue} \textbf{} \xspace}
\fi

%% file: section/iclr/100-intro.tex
\vspace{-.35cm}
\section{Introduction}\label{sec:intro}
\vspace{-.1cm}

For years, large language models (LLMs) have been known to leak information about their training data~\citep{xkcd2169}, 
sometimes reproducing exact or near-exact training sequences when prompted.
The field refers to eliciting such sequences in a model's outputs as ``training-data extraction,'' and treats extraction as an observable signal that the model ``memorized'' those sequences during training~\citep{carlini2021extracting,lee2022dedup}. 
Although memorization can be studied in other ways~\citep{shokri2017membership,carlini2019secret,carlini2022membership}, 
extractable memorization has drawn particular attention because of its implications for privacy and copyright~\citep{lee2023talkin,cooper2024files,nolte2025privacy,brown2022privacy,cooper2024unlearning}.\looseness=-1

Amid a surge of research into how much LLMs memorize their training data, two recurring patterns run through recent work.
On one hand, some work overstates extraction, for instance, claiming extraction for sequences too short to reliably distinguish memorization from predictability or for outputs elicited by prompts that directly leak the target sequence~\citep{chang-etal-2023-speak,karamolegkou2023copyrightviolationslargelanguage,wei2025interrogatingllmdesignfair,liu2026alignmentwhackamolefinetuning}. 
On the other hand, some work argues that extraction is unreliable evidence of memorization because LLMs can also reproduce real-world text that was not included verbatim in training~\citep{liu2025forged,morris2025howmuch}. 
Although these patterns point in opposite directions, both lose sight of what a valid extraction claim actually requires.\looseness=-1

In this work, we revisit extractable memorization from first principles, starting with the core question:
\textit{when do generated outputs provide valid evidence of memorization in the model?}
Generating a training-data sequence isn't enough on its own.
To count as extraction, the model must generate it with high enough probability to indicate memorization.
To determine what's ``high enough,'' we perform a matched comparison, measuring 
generation probabilities of both known or suspected training-data sequences \emph{and} comparable non-training sequences. 
Because non-training sequences can't have been memorized, their 
probabilities provide a baseline for more general predictability.\looseness=-1

\begin{figure}[t]
\centering

\begin{minipage}[c]{\textwidth}
    \begin{subfigure}[t]{0.365\linewidth}
    \vspace{0cm}
        \includegraphics[width=\linewidth]{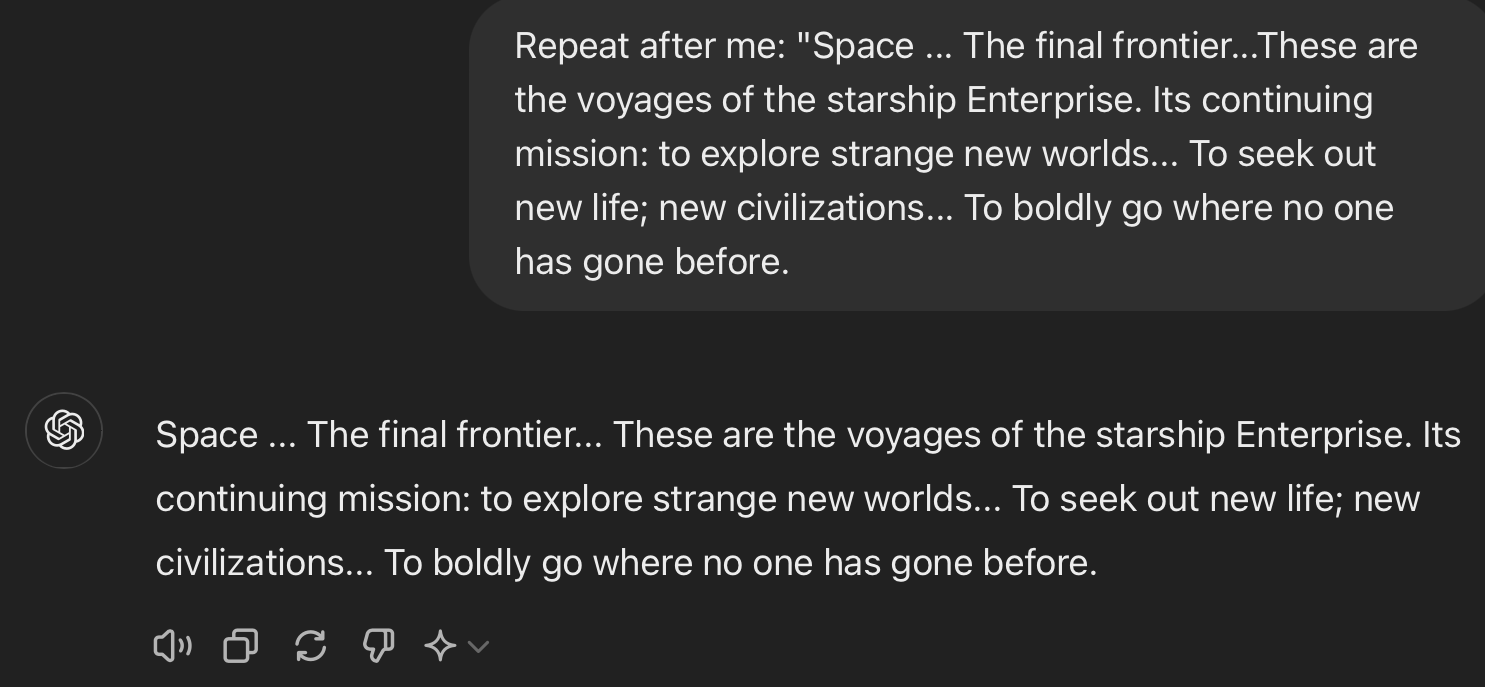} 
    \end{subfigure}
    \begin{subfigure}[t]{0.365\linewidth}
    \vspace{0cm}
        \includegraphics[width=\linewidth]{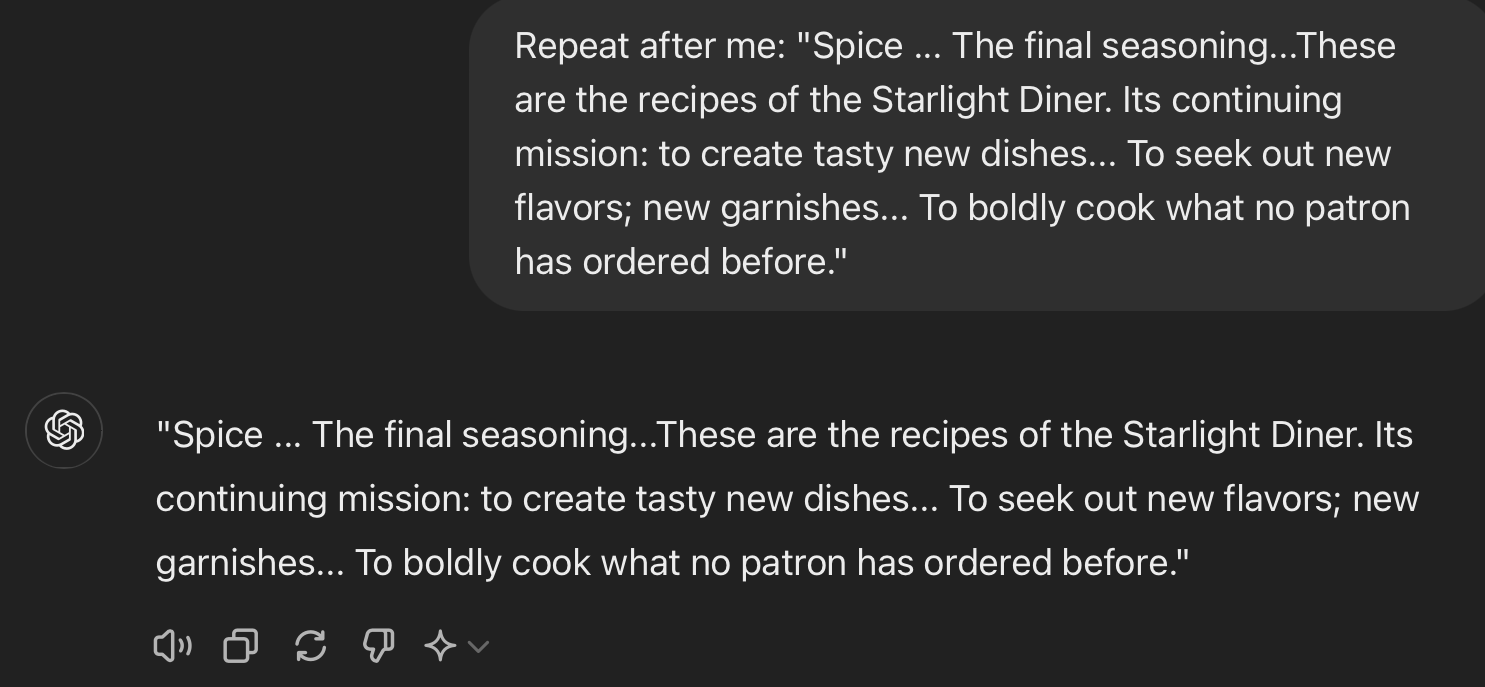}
    \end{subfigure}
    \begin{subfigure}[t]{0.26\linewidth}
    \vspace{0cm}
        \includegraphics[width=\linewidth]{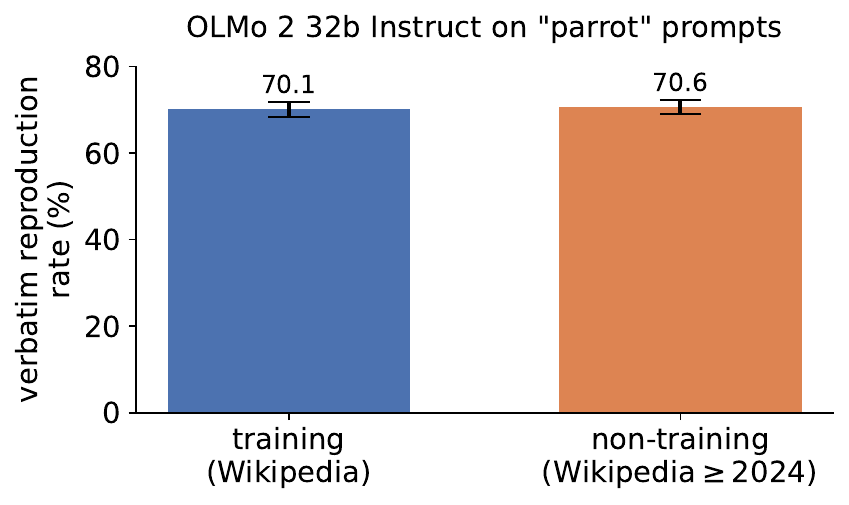}
    \end{subfigure}
\end{minipage}
\vspace{-.2cm}
\caption{\textbf{Capable instruction-tuned models follow ``Repeat after me'' (``parrot'') instructions.} 
(\textbf{left}) ChatGPT exactly reproduces likely training data (a \emph{Star Trek: The Next Generation} quote), and (\textbf{middle}) non-training data (some original prose), reproduced from~\citet{cooper2024files}.
(\textbf{right}) We run a larger-scale version using OLMo 2 32B Instruct ($N\!=\!3000$ each), where the matched comparison yields equal rates and carries no signal about memorization (\S\ref{app:sec:intro:parrot}).\looseness=-1
}
\vspace{-.3cm}
\label{fig:parrot}
\end{figure}

\textbf{Of course, not all verbatim generation is extraction.}
The need for a matched comparison is especially clear with an intentionally extreme example.
Consider a ``parrot'' prompt that supplies a sequence and asks the LLM to repeat it verbatim (Fig.~\ref{fig:parrot}).
Run only on training data, a high success rate could be mistaken for evidence of extraction.
Yet capable models reproduce matched non-training sequences at essentially the same rate: they are simply copying text already in the prompt.
But for our purposes, the reason the rates match is beside the point; 
the comparison shows that, on its own, successful verbatim generation of training data provides no basis for claiming memorization.\looseness=-1

Asking an LLM to parrot training data is clearly not a serious extraction procedure, but the same failure arises in realistic settings.
Consider a recent procedure, which searches for adversarially optimized prompts that elicit a target sequence and deems extraction successful if the resulting prompt is shorter than the target~\citep{schwarzschild2024rethinking}.
The authors show success on the widely quoted line from hockey player Wayne Gretzky, ``You miss 100\% of the shots you don't take.''
We reproduce that success using Llama~2 13B, but also succeed on non-training data:
a similar-length statement basketball player Jalen Brunson made in the 2026 NBA Finals, long after the training cutoff (\S\ref{app:sec:intro:acr}).
This does \emph{not} imply that Llama~2 13B didn't memorize the Gretzky quotation; 
it simply shows that success with this procedure on training data doesn't by itself warrant a memorization claim.\looseness=-1

\textbf{When (near-)verbatim generation counts as extraction.}
Prior work has generally treated matched comparisons (when it runs them at all) as informal sanity checks~\citep{lee2022dedup,carlini2023quantifying,hayes2025measuringmemorizationlanguagemodels,pappu2025rlhf,schwarzschild2024rethinking,cooper2025books,cooper2026nv}. 
But this is insufficient. 
Instead, we use the (near-)verbatim generation probabilities of matched non-training data to calibrate a threshold;
a sequence's probability must clear that threshold to make a valid extraction claim, otherwise its probability could be due to more general predictability.
That is, calibration makes explicit the inferential chain that prior work has left implicit:\looseness=-1
\vspace{-.1cm}
\begin{claimbox}[label={box:extraction-claim:main}]{Making a valid extraction claim}
\small
(\textbf{a}) we measure 
a sequence's \textit{generation probability} under some set of prompts;
(\textbf{b}) when that probability clears the calibrated threshold, it's too high to be explained by general predictability;
(\textbf{c}) this supports the conclusion 
that the sequence is a \textit{member} of the training data that the model has \textit{memorized}; 
(\textbf{d}) this conclusion is what licenses calling its generation probability an \textit{extraction probability}.\looseness=-1
\end{claimbox}

\textbf{Contributions.} 
We formalize and run experiments using matched comparisons in two settings. 
First, for 
sequences sampled from populations, a conformal test calibrates memorization claims to a chosen false-positive rate (\S\ref{sec:test}).
On Wikipedia, for example, $24\%$ of apparent $10$-token verbatim extractions from OLMo~2 are false positives, showing that short-sequence experiments 
can substantially overstate memorization (\S\ref{sec:experiments:population}). 
Second, for a fixed document where there is no population false-positive rate to control, a census instead calibrates against a matched non-training document.  
For Llama~3.1 70B on books, 
calibrated thresholds for probabilistic extraction using $50$-token suffixes range from $10^{-27}$ to $10^{-21}$. 
With such small thresholds, extremely low-probability sequences can support memorization claims, yet are far too improbable to sample in practice (\S\ref{sec:experiments:census}).
Given these results, we refine the 
definition of ``extractable memorization'' to require both a valid memorization claim from a matched comparison and near-certain generation within a realistic query budget (\S\ref{sec:extractable-memorization}).\looseness=-1

%% file: section/iclr/200-prelim.tex
\vspace{-.2cm}
\section{Background and related work}\label{sec:prelim}
\vspace{-.1cm}

Let $\vocab$ be a token vocabulary and $\vocab^*$ the set of finite token sequences.
An \newterm{autoregressive language model} $\params$ defines a next-token distribution $q_\params(\cdot\mid\vv)$ for each $\vv\in\vocab^*$.
Run from a \newterm{prompt} $\promptseq$, a \newterm{decoding policy} $\dec$ induces a distribution $\channel(\cdot\mid\promptseq)$ over \newterm{continuations}.
For an $\eventlen$-token continuation $\outseq=(\outtok_1,\ldots,\outtok_{\eventlen})$, its generation probability is
\vspace*{-.2cm}
\begin{align}
\label{eq:generation-prob}
\textstyle
\channel(\outseq\mid\promptseq)
\coloneqq
\prod_{t=1}^{\eventlen}
\channel(\outtok_t\mid\promptseq,\outseq_{<t}).
\end{align}
The \newterm{base} policy samples from the whole vocabulary ($\channel = q_\params$), while \newterm{top-$k$}  restricts at each step $t$ to the $k$ highest-probability tokens, renormalizes, and samples.
By contrast, \newterm{greedy decoding} is deterministic, so $q_{\params,\greedy}(\outseq\mid\promptseq)\in\{0,1\}$:
the unique greedy continuation has probability $1$ and every other continuation has probability $0$.

Loosely speaking, \newterm{memorization} of a specific training sequence $\member$ means the model assigns unusually high probability to that sequence, indicating that information about $\member$ is encoded in the model. 
\newterm{Extraction} is a narrower phenomenon: 
it's the reproduction of a memorized sequence in the model's outputs via prompting.
When a sequence is extractable, the model assigns it high conditional probability $\channel(\member \mid \promptseq)$, reliably reproducing $\member$ when prompted with $\promptseq$.
Not all memorized sequences are extractable~\citep{carlini2021extracting}. 
A sequence can be encoded in the model yet not reliably reproduced under any prompt~\citep{nasr2023scalable,nasr2025scalable}. 

\textbf{Members and non-members at a chosen resolution.}
A model is trained on a fixed dataset $\dataset$.
By definition, memorization is only possible for (near-duplicate) content present in $\dataset$:
a model can't have memorized something it never saw during training.
A sequence $\candseq$ that an extraction procedure deems to be memorized is therefore a \newterm{member} of $\dataset$.
Failure to extract $\candseq$ is \emph{not} evidence that it's a \newterm{non-member}, because not all training data are memorized and not all memorized training data are extractable under a given procedure. 
Exact-sequence membership is brittle for language models: near-duplicates are functionally members because the model has seen the content, whatever tokenization or minor syntax variation happened to accompany it~\citep{lee2022dedup,zhang2023counterfactual}.
Similar to \citet{cooper2026nv}, we therefore choose a resolution by fixing a candidate length $\eventlen$, dissimilarity function $\distance : \vocab^\eventlen \times \vocab^\eventlen \to \R_{\ge 0}$, and tolerance $\tol \geq 0$ (\S\ref{app:sec:prelim:membership}).
For a candidate $\candseq\in\vocab^{\eventlen}$, define\looseness=-1
\vspace{-.1cm}
\begin{align}
\label{eq:membership-indicator}
\ball{\tol}{\candseq}
\;\coloneqq\;
\{\outseq\in\vocab^{\eventlen}:\distance(\outseq,\candseq)\leq\tol\}, \qquad 
\memberind{\candseq} \;\coloneqq\;
\1\{\dataset\cap\ball{\tol}{\candseq}\neq\varnothing\},
\end{align}
where $\dataset \cap \ball{\tol}{\candseq}$ is understood as searching $\dataset$ for any length-$\eventlen$ sequence within $\tol$ of $\seq$. 
We refer to sequences under test for memorization as \newterm{candidates}  (\S\ref{sec:test}).
Candidate $\candseq$ is a near-verbatim member when $\memberind{\candseq}=1$ and a non-member otherwise.
Therefore, non-membership of $\candseq$ is deliberately stronger than requiring that $\candseq$ is held out from training:
no $\candseq$ in its chosen-resolution ball $\ball{\tol}{\candseq}$ may have training support.
Setting $\tol=0$ recovers verbatim (non-)membership.\looseness=-1

\textbf{Extraction procedures.}
Extraction involves prompting a model to attempt to reproduce a candidate $\candseq$, but methods that fit this loose description can differ substantially.
They nevertheless share five components:
a model $\params$, decoding policy $\dec$, prompt-selection rule $\promptrule$, scoring procedure $\score$, and verification procedure $\verify$.
The rule $\promptrule$ maps each candidate to a finite prompt class $\promptclass(\candseq)$, e.g., a singleton natural prefix, all natural prefixes, an adversarial search space, or another set of prompts (\S\ref{app:sec:prelim:extraction}).
For each prompt, the canonical score is the probability
$\channel(\ball{\tol}{\candseq}\mid\promptseq)$ (Eq.~\ref{eq:generation-prob}) or an approximation of it.
An aggregation rule collapses per-prompt scores into the candidate score
\vspace{-.1cm}
\begin{equation}
\label{eq:candidate-score}
\textstyle
\score(\candseq)
=
\agg_{\promptseq\in\promptrule(\candseq)}
\score_{\promptseq}(\candseq),
\end{equation}
where higher means more easily generated. 
Finally, $\verify$ uses the score either to make a memorization claim or to make no such claim.

\begin{definition}[Extraction procedure]
\label{def:extraction-procedure}
A fixed tuple \((\params,\dec,\promptrule,\score,\verify)\) defines an
\newterm{extraction procedure}, 
where model $\params$ and decoding policy $\dec$ configure the distribution $\channel$;
$\promptrule$ specifies the prompts used for each candidate sequence $\candseq$;
$\score$ returns the score for $\candseq$ (computed under $\params$, $\dec$, and $\promptrule$);
and, based on its score, $\verify$ either deems $\candseq$ memorized or makes no claim. 
\end{definition}

Discoverable extraction, its probabilistic and near-verbatim variants, and adversarial prompt search 
are all instances of Def.~\ref{def:extraction-procedure}~\citep{lee2022dedup,carlini2023quantifying,hayes2025measuringmemorizationlanguagemodels,cooper2025books,cooper2026nv,schwarzschild2024rethinking,nasr2023scalable,nasr2025scalable} (\S\ref{app:sec:prelim:prior}).
For our purposes, the distinction that matters is the last component:
in the next section, we construct $\verify$ from matched non-members, calibrating when a generation score is high enough to support a memorization claim.\looseness=-1

%% file: section/iclr/300-conformal.tex
\vspace{-.2cm}
\section{A conformal test for memorization}\label{sec:test}
\vspace{-.1cm}

To varying extents, prior extraction methods leave open how to define $\verify$:
the rule that decides whether a candidate's $\score$ is high enough to support a memorization claim, rather than high merely because the continuation is predictable for a capable language model.
While some work runs the same scoring procedure on non-training data and observes that reproduction is much lower~\citep{cooper2025books,hayes2025measuringmemorizationlanguagemodels}, it stops short of turning this comparison into a calibrated decision.

\begin{figure}[t]
  \centering
  \includegraphics[width=\linewidth]{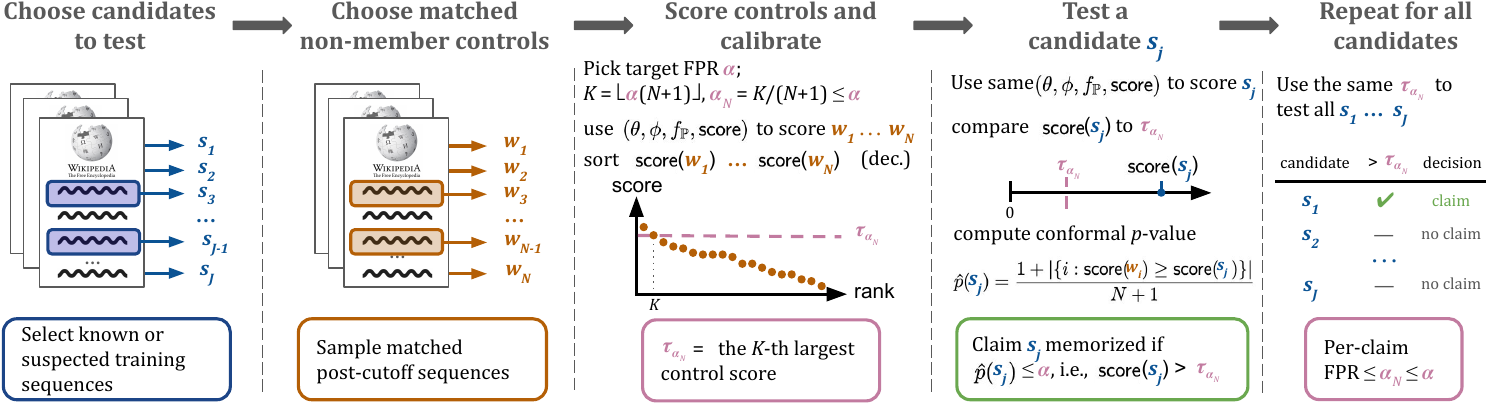}
  \vspace{-.3cm}
  \caption{\textbf{Illustrating a conformal test for memorization.}
  (\textbf{1}) Select candidates to test (known or suspected members) and (\textbf{2}) $N$ matched non-member controls.
  (\textbf{3}) Pick a target FPR $\alpha$; apply the extraction $\score$ to controls to calibrate $\tau_{\alpha_N}$.
  (\textbf{4}) Apply the exact same $\score$ to a candidate and claim it's memorized only when its score exceeds $\tau_{\alpha_N}$.
  (\textbf{5}) Repeat for each candidate.
  If the control scores and a fresh non-member's score are exchangeable, each claim has marginal FPR at most $\alpha_N\le\alpha$.\looseness=-1}
  \label{fig:conformal-illustration}
  \vspace{-.3cm}
\end{figure}

Here, we formalize $\verify$ as a matched comparison, made precise as a \newterm{conformal test}.
Recall that a candidate $\candseq$ is either a member or a non-member (\S\ref{sec:prelim}).
The test's null hypothesis is that $\candseq$ is a \emph{matched} non-member;
it rejects the null when $\score(\candseq)$ exceeds a threshold calibrated from the scores of matched non-members (\newterm{controls}).
Rejecting means $\candseq$'s generation probability is too high for it to be a matched non-member, indicating it's a member of the training data the model memorized (Box~\ref{box:extraction-claim:main}).\footnote{Extraction experiments (including ours) are often run on known members. 
    In such cases, the training-data membership part of the claim is trivial, but the test still either claims memorization or makes no claim (\S\ref{app:sec:conformal:multiplicity}).\looseness=-1
} 
Calibration lets us make that claim at a false-positive rate (FPR) chosen in advance: 
how often the test claims memorization for a non-member (Fig.~\ref{fig:conformal-illustration}).
Because the test operates on the abstract extraction procedure (Def.~\ref{def:extraction-procedure}), it applies to any extraction method that instantiates that definition.\looseness=-1

\custompar{Matched comparisons for calibrated memorization claims}
Formally, the \emph{matched} non-member null hypothesis is $\memberind{\candseq}=0$ (Eq.~\ref{eq:membership-indicator}), with $\candseq$ drawn from the same distribution as the controls.
\newterm{Matched} means two things. 
First, we select controls the same way as candidates (in practice, known or suspected members): 
they share the source, format, length, and other features that can affect $\score$, except controls are known non-members 
(e.g., pre-cutoff Wikipedia entries as candidates, post-cutoff entries as controls).
Second, the same fixed extraction procedure---model $\params$, decoding policy $\dec$, prompt-selection rule $\promptrule$, $\score$ (Eq.~\ref{eq:candidate-score})---is applied to candidates and controls.
This makes their scores exchangeable under the matched null. 
If the candidate were a non-member, its score would look like one more draw from the control scores' distribution.\looseness=-1

Let $\ctrlrv_1,\ldots,\ctrlrv_N\sim\ctrl$ be the controls: 
$N$ random draws from a control distribution $\ctrl$ supported on non-members ($\memberind{\ctrlrv_i}=0$).
For any candidate $\candseq$, define\looseness=-1
\begin{equation}\label{eq:conformal-pvalue}
\widehat p(\candseq)
=\frac{1+\left|\{i:\score(\ctrlrv_i)\geq\score(\candseq)\}\right|}{N+1}, 
\end{equation}
the \newterm{conformal $p$-value}. 
Pooling the $N$ control scores with the candidate's, $\widehat p(\candseq)$ is the fraction of those $N+1$ scores that are at least as large as $\score(\candseq)$.
A small value means that almost no matched non-member reaches the candidate's score. 
Fix a \newterm{target FPR} $\alpha\in(0,1)$ independently of the control and candidate scores.
A finite control sample can only certify levels in steps of $1/(N+1)$ because the cutoff must be one of the $N$ control scores, so write
\vspace{-.1cm}
\begin{align}
\label{eq:alpha-n}
K=\lfloor\alpha(N+1)\rfloor
\;\; \text{and} \;\;
\alpha_N=K/(N+1)\le\alpha. 
\end{align}
$K$ indexes which control score serves as that cutoff; 
$\alpha_N$ is the largest attainable \newterm{certified FPR} not exceeding $\alpha$ (\S\ref{app:sec:conformal:notation}, \S\ref{app:conformal-proof}). 
The smallest nonzero level any finite control sample can certify is $1/(N+1)$, reached at $K=1$, where the threshold is the largest control score. 
Because $\widehat p\geq1/(N+1)$ (Eq.~\ref{eq:conformal-pvalue}), setting $\alpha<1/(N+1)$ gives $K=0$ and the test can never reject (\S\ref{app:sec:conformal:notation}). 

If the control scores in decreasing order are
$\score_{(1)}\geq\cdots\geq\score_{(N)}$, then, for $K\geq1$, the \newterm{calibrated threshold} at certified level $\alpha_N$ is
\vspace{-.1cm}
\begin{equation}\label{eq:conformal-threshold}
\tau_{\alpha_N}\coloneqq\score_{(K)},\qquad
\widehat p(\candseq)\leq\alpha
\ \Longleftrightarrow\
\score(\candseq)>\tau_{\alpha_N}.
\end{equation}
The strict inequality is important: 
a candidate whose score is tied with the threshold doesn't pass.
Because matching makes the candidate's score exchangeable with the control scores, a non-member candidate is falsely claimed as memorized with probability at most $\alpha_N\le\alpha$ (\S\ref{app:conformal-proof}).

\begin{proposition}[Finite-sample validity of the conformal test]
\label{prop:conformal-validity}
Fix $\alpha\in(0,1)$ and the complete extraction procedure independently of the calibration controls and candidate scores.
Let $\candrv$ be a fresh candidate that satisfies the null---a non-member ($\memberind{\candrv}=0$) drawn by the same rule as the controls---and suppose its score and the $N$ control scores are exchangeable.
Then
\vspace{-.1cm}
\begin{equation}\label{eq:conformal-validity}
\Pr_0\!\left[\widehat p(\candrv)\leq\alpha\right]
\leq \alpha_N\leq\alpha,
\end{equation}
where $\Pr_0$ is probability under the null, taken over both the random controls and the fresh candidate, so the guarantee is marginal and the left-hand side is the FPR. 
For $K\geq1$ (Eq.~\ref{eq:alpha-n}), rejection is exactly the strict threshold test in Eq.~\ref{eq:conformal-threshold}; 
for $K=0$, the deterministic test can't reject.
If the $N+1$ scores have no ties, the first inequality is an equality; ties can only make this test conservative. 
\end{proposition}
The guarantee applies per candidate:
it's not conditional on the realized controls (\S\ref{app:sec:conformal:alpha-star:conditional}), and it doesn't by itself bound the familywise error rate or the false discovery rate when many candidates are tested together (\S\ref{app:sec:conformal:multiplicity}, \citet{bates2021testing}). 
\S\ref{app:conformal-proof} gives the full proof, which follows the standard conformal argument~\citep{shafer2008tutorial}: under exchangeability the null candidate's rank among the $N+1$ scores is uniform, so at most $K$ ranks reject.
\S\ref{app:sec:conformal} discusses the guarantee in more detail. 

\custompar{What the test certifies}
The test enables valid, calibrated claims about memorization in LLMs that surfaces through a fixed extraction procedure (Def.~\ref{def:extraction-procedure}), including near-verbatim memorization. 
At tolerance $\tol$, clearing the threshold certifies near-verbatim membership of the candidate $\candseq$, not that the model memorized $\candseq$ exactly (Eq.~\ref{eq:membership-indicator}).  
If $\candseq$ doesn't exceed the calibrated threshold (Eq.~\ref{eq:conformal-threshold}), no memorization claim is made (Fig.~\ref{fig:conformal-illustration}).
This means that failing to clear the threshold is \emph{not} evidence against memorization.
The model may still encode the candidate in a way that this prompt, decoding policy, or $\score$ doesn't surface.
So the test bounds false positives without saying anything about false negatives;
it doesn't estimate total latent memorization in the model.\looseness=-1

\custompar{Revisiting ``parrot'' prompts}
Recall that members and non-members are reproduced at the same rate when the LLM is prompted to ``Repeat after me'' (Fig.~\ref{fig:parrot}). 
In the original LLM extraction paper, \citet{carlini2021extracting} identified the ``parrot'' prompt case, and  observed that no simple refinement of their definition for extraction resolved it;
their solution was to disallow such prompts (\S\ref{app:prior-work:matched}).
But the conformal test needs no such restriction: 
matched controls score just as high as candidates, the threshold calibrated from them is just as high (\S\ref{sec:experiments:population}), nothing clears the threshold, and so the test makes no memorization claim.
This is the only defensible outcome when members and non-members behave the same under the same extraction procedure, and isn't specific to ``parrot'' prompts.
Any mechanism that raises scores independently of membership---copying, instruction-following, predictability, adversarial coaxing---raises the threshold, so the test also handles these cases.\looseness=-1

\custompar{What the guarantee rests on}
A memorization claim is always made relative to some baseline of predictability; 
the conformal test just makes it explicit with the matched-non-member null, so that each claim is read against the chosen control distribution. 
This means the guarantee is only as good as the matched comparison behind it:  
controls must be genuine non-members and be distributionally matched to the candidates.
In practice, we can only make a best effort for both (\S\ref{sec:experiments:population}, \S\ref{sec:experiments:census}, \S\ref{sec:takeaways}, \S\ref{app:sec:matching}).

%% file: section/iclr/400-population.tex
\vspace{-.2cm}
\section{Population rate experiments}\label{sec:experiments:population}
\vspace{-.1cm}

\begin{figure}[t]
\centering
\vspace{-.3cm}
\includegraphics[width=\linewidth]{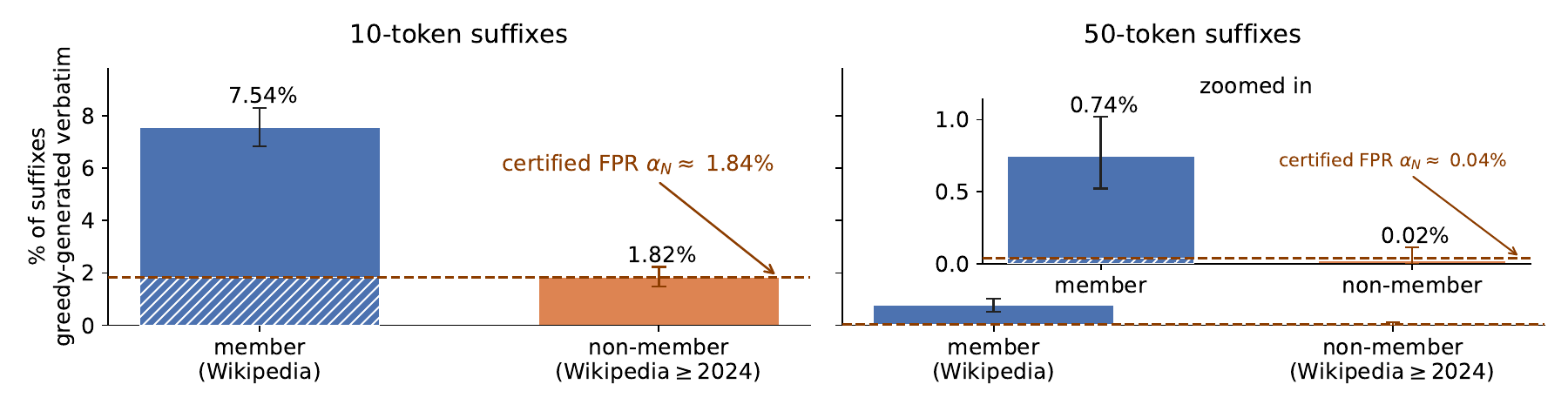}
\vspace{-.75cm}
\caption{\textbf{Verbatim greedy generation of members and matched non-members.} 
For OLMo~2 32B base on Wikipedia, we run discoverable extraction for $10$- (\textbf{left}) and $50$-token (\textbf{right}) suffixes. 
Each bar is the measured rate with a $95\%$ Clopper--Pearson interval (\S\ref{app:population}).
$\alpha_N$ (Eq.~\ref{eq:alpha-n}) is the largest FPR the $N$ controls can certify; 
it's a false-positive floor.
For $10$-token suffixes, $\alpha_N=1.84\%$ against $7.54\%$ for members; 
${\sim}24\%$ of flagged extraction matches the rate at which the same test flags non-members.
For $50$-token suffixes, the floor is close to zero (see main text).\looseness=-1}
\label{fig:olmo-greedy-bar-10-vs-50}
\vspace{-.3cm}
\end{figure}

\begin{figure}[t]
  \centering
  \vspace{-.2cm}
    \begin{subfigure}[t]{0.66\textwidth}\centering
      \includegraphics[width=\linewidth]{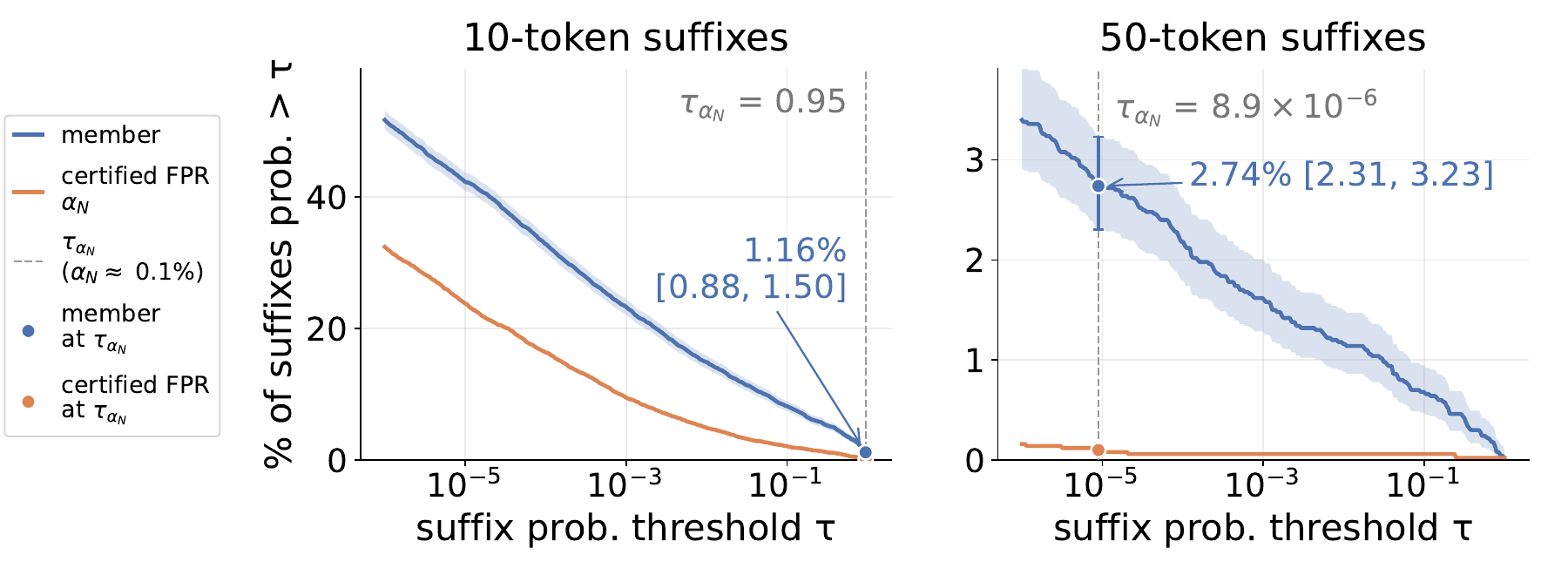}
      \vspace{-.5cm}
      \caption{verbatim generation (same prefixes, $10$- and $50$-token suffixes)}
      \label{fig:olmo-calibrated-rate:gen}
    \end{subfigure}\hfill
    \begin{subfigure}[t]{0.33\textwidth}\centering
      \includegraphics[width=\linewidth]{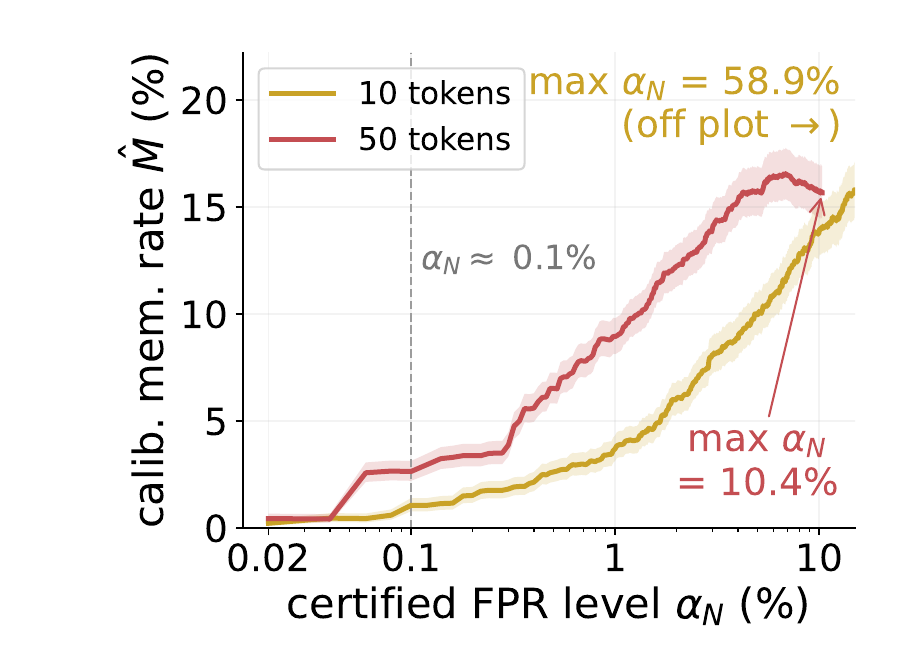}
      \vspace{-.5cm}
      \caption{calibrated memorization}\label{fig:olmo-calibrated-rate:mem}
    \end{subfigure}
    \vspace{-.15cm}
	  \caption{\textbf{Generation and calibrated memorization rate for OLMo~2 32B base on Wikipedia.}
      (\textbf{a}) Complementary CDFs of scores, sweeping over $\tau$: 
      a point $(\tau, y)$ means $y\%$ of sequences score above $\tau$ (i.e., $\channel(\candseq \mid \promptseq) > \tau$). 
      $10$-token suffixes have much larger absolute rates. 
      (\textbf{b})
	  Calibrated memorization (Eq.~\ref{eq:emp-cmr}) for $10$- and $50$-token suffixes across certified FPR levels $\alpha_N$. 
	  The $50$-token rate is the larger of the two across the sweep. 
	  (The controls contain some member contamination. 
    For $50$ tokens in particular, the plotted rate understates calibrated memorization; see \S\ref{app:population}.)
    Bands are $95\%$ Clopper--Pearson intervals on the member rate, shifted by the constant $\alpha_N$ in (\textbf{b}).
}
  \label{fig:olmo-calibrated-rate}
  \vspace{-.3cm}
\end{figure}

The conformal test can avoid making invalid memorization claims for contrived setups like ``parrot'' prompts, but it can also do a lot more: 
it can support valid memorization claims about populations.
To illustrate this, we estimate the population rate of calibrated verbatim memorization for Wikipedia entries, using OLMo~2 32B base.
(\S\ref{app:population} and \S\ref{app:extractability:population} report results for 7B, 13B, and 32B.)
Because the training corpus is public~\citep{olmo2}, membership is known:
the $5{,}000$ candidate passages are members, coming from entries dated between January~2021 and the December~2023 training cutoff.
We match them with $5{,}000$ passages whose source entries are dated 2024, after the training cutoff.
Each passage has a $50$-token natural prefix and a suffix of either $10$ or $50$ tokens.
We draw candidates and controls uniformly at random within their date ranges, so each is an i.i.d.\ sample.
Note that the conformal guarantee needs only exchangeability (Prop.~\ref{prop:conformal-validity}, \S\ref{app:conformal-proof}), but in practice we additionally require independence.
Independence makes the confidence intervals we plot valid (Fig.~\ref{fig:olmo-greedy-bar-10-vs-50} \&~\ref{fig:olmo-calibrated-rate}).
It also relates $\alpha_N$ to the \newterm{population FPR} $\alpha^*$---the fraction of the whole non-member population that the realized threshold $\tau_{\alpha_N}$ incorrectly flags as memorized---by $\E[\alpha^*]=\alpha_N$ (\S\ref{app:sec:conformal:alpha-star}), which our empirical estimate of calibrated memorization relies on (below, Eq.~\ref{eq:emp-cmr}).

\custompar{A generation rate is not a memorization rate}
We begin with discoverable extraction~\citep{lee2022dedup,carlini2023quantifying}, the standard metric in research and model release reports~\citep{llama3,reid2024gemini}.
Because greedy decoding is deterministic, its score is binary:
the continuation either exactly matches the candidate or it doesn't.
There's nothing to calibrate because a binary score admits only one threshold (just one operating point).
Fig.~\ref{fig:olmo-greedy-bar-10-vs-50} contrasts verbatim generation on members with the matched non-member controls.
We deliberately call it ``generation'' rather than ``extraction'':
for a genuine non-member, a match can't reflect memorization.

The dashed line in each panel is the \newterm{false-positive floor}: 
the certified FPR $\alpha_N$ (Eq.~\ref{eq:alpha-n}) at which the test flags sequences that can't be memorized.
The bars are \newterm{in-sample rates}: for these specific $5{,}000$-sequence samples, the fraction of members and matched controls reproduced verbatim (with $95\%$ Clopper--Pearson intervals).
For $10$-token suffixes the floor is $\alpha_N=1.84\%$ against $7.54\%$ on members, 
so roughly one quarter ($1.84/7.54\approx24\%$) of the apparent training-data ``extraction'' rate is matched by the rate at which the same test flags non-members.\footnote{We compute the floor share from $\alpha_N$ rather than from the in-sample control rate because the in-sample rate is what one draw of controls happened to produce; $\alpha_N$ is the certified level.\looseness=-1}
The comparison makes no claim that a particular generated member is \emph{not} memorized.
Rather, it shows that a short verbatim match is \emph{not} a sufficiently specific membership signal: 
an experiment run on members alone would attribute the whole $7.54\%$ to memorization, when the same procedure flags sequences that can't be memorized at $1.84\%$. 
At $50$ tokens the picture is very different.
The member generation rate is smaller ($0.74\%$), but the non-member floor is close to zero ($0.04\%$).
That floor comes from a single false positive, which turns out to be null contamination rather than a genuine non-member (\S\ref{app:population}), so both that floor and the ${\approx}5\%$ share it implies are overstated.
Contamination only leads to more conservative claims: 
a contaminated control is actually a member;
if it scores high, it can raise the bar a candidate must clear to claim memorization (\S\ref{app:sec:matching}, \S\ref{app:sec:population:contamination}).

\custompar{Continuous calibration}
Probabilistic extraction instead scores a continuation with $\channel(\candseq\mid\promptseq)\in[0,1]$ (Eq.~\ref{eq:generation-prob}) under top-$k$ decoding ($k=40$), computed exactly by teacher forcing~\citep{hayes2025measuringmemorizationlanguagemodels,cooper2025books}.
This continuous score lets us sweep the conformal threshold across operating points, rather than expose only greedy decoding's single operating point.
For each target FPR $\alpha_N$, the controls set $\tau_{\alpha_N}$ (Eq.~\ref{eq:conformal-threshold}) and a member is deemed memorized only if its score exceeds that threshold.
Again, we see that a generation probability on its own isn't evidence of memorization.
For each threshold $\tau$, Fig.~\ref{fig:olmo-calibrated-rate:gen} plots the share of suffixes scoring above it. 
At $10$ tokens the absolute rates are far higher, but so are the control rates (certified FPR $\alpha_N$):
for the smallest thresholds plotted, the member rate exceeds $50\%$ and the certified FPR exceeds $30\%$; 
$\tau_{\alpha_N}=0.95$, so a member must be nearly certain to clear it.
At $50$ tokens the member rate peaks near $3\%$ while the controls stay near zero, and the threshold falls to $\tau_{\alpha_N}=8.9\times10^{-6}$.
The two raw rates aren't the right comparison: 
they sit at thresholds five orders of magnitude apart, each set by what the non-member controls happen to do at that suffix length.
Instead, we should compare how far each member rate sits above its own control rate.\looseness=-1

\begin{definition}[Calibrated memorization rate]\label{def:cmr}
For a fixed extraction procedure (Def.~\ref{def:extraction-procedure}) and threshold $\tau$, let
$G(\tau)=\Pr[\score(\candseq)>\tau\mid\memberind{\candseq}=1]$
and
$\mathrm{FPR}(\tau)=\Pr[\score(\candseq)>\tau\mid\memberind{\candseq}=0]$.
The \newterm{calibrated memorization rate} is
$M(\tau)\coloneqq G(\tau)-\mathrm{FPR}(\tau)$. 
In practice, at the conformal operating point we estimate this by
\vspace{-.15cm}
\setlength{\belowdisplayskip}{4pt}\setlength{\belowdisplayshortskip}{4pt}
\begin{equation}
\hat M\coloneqq\hat G-\alpha_N,
\label{eq:emp-cmr}
\end{equation}
where $\hat G$ is the empirical member generation rate above $\tau_{\alpha_N}$.
Subtracting $\alpha_N$ makes $\hat M$ unbiased, since $\E[\alpha^*]=\alpha_N$ (\S\ref{app:sec:conformal:notation}, \S\ref{app:sec:conformal:alpha-star}).
\end{definition} 
Under the matched-control assumption, the calibrated memorization rate is an extractability-based lower bound on total latent memorization in the model (\S\ref{sec:test}). 
A suffix can be memorized yet score below $\tau$ under the fixed procedure, in which case no claim is made. 

Fig.~\ref{fig:olmo-calibrated-rate} shows how calibration removes the short-suffix predictability floor.
In contrast to the raw generation rates, the $50$-token curve sits above the $10$-token one across the sweep (e.g., at $\alpha_N\approx0.1\%$, $\hat M = 2.64\%$ against $1.06\%$).
Making the raw generation rate larger doesn't make the calibrated evidence for memorization stronger.
Further, for memorization, the field's posture is to make conservative claims: 
to privilege a low FPR, even at the cost of more false negatives, because over-claiming memorization is treated as more costly than under-claiming it~\citep{carlini2022membership,cooper2025books}.
So while Fig.~\ref{fig:olmo-calibrated-rate:mem} plots a range of certified FPRs, the low-FPR end is where one would operate in practice (\S\ref{app:sec:conformal:alpha-star:practice}).
(Also note that the sweep's minimum is $\alpha_N\approx0.02\%$ because $N$ controls certify levels only in steps of $1/(N+1)$; see \S\ref{sec:test}. 
Certifying a smaller floor requires more controls, not a smaller $\alpha$.)\looseness=-1

%% file: section/iclr/500-census.tex
\vspace{-.2cm}
\section{Census experiments}\label{sec:experiments:census}
\vspace{-.1cm}

For extractable memorization, the question of interest isn't always about population rates.
Sometimes we want to understand the extent to which a single specific training document is memorized.
We take a \newterm{census} approach and evaluate the whole document.
Unlike the Wikipedia experiments (\S\ref{sec:experiments:population}), the conformal test doesn't apply;
there's no population being sampled, so the results are computed exactly and there's no FPR to estimate or certify. 
But we can still perform a matched comparison with non-training data. 
We do exactly this for Llama~3.1 base models, using the sliding-window verbatim probabilistic extraction procedure from \citet{cooper2025books} (top-$k$ decoding, $k=40$).
We test four pairs of books, where each pair matches a book in Books3 (which was included in Llama's training data) with a recent book by the same author that post-dates Llama~3.1's training cutoff.
Matching on author is our best effort at exchangeability (\S\ref{app:sec:matching}): within a pair, the books share the author's style and are in the same genre, and in two cases they're even in the same series, with overlapping characters and plot elements (\S\ref{app:census}).\looseness=-1

The sliding-window procedure tiles each book into heavily overlapping fixed-length windows ($50$-token prefix, $50$-token suffix $\candseq$), where each character is covered by many shifted suffixes (\S\ref{app:census:results}).
We score every overlapping suffix given its prefix.
Then, instead of a population rate, we compute whole-document \newterm{coverage}~\citep{cooper2025books} for the member book.
Let $\mathsf{span}(\candseq)$ be the set of character indices spanned by a given $50$-token $\candseq$.
For a threshold $\tau$ and book length (in characters) $L$,\looseness=-1 
\vspace{-.1cm}
{\setlength{\belowdisplayskip}{4pt}\setlength{\belowdisplayshortskip}{4pt}
\begin{equation}
\label{eq:coverage}
\textstyle
\mathsf{coverage}(\tau)
=
\frac{1}{L}
\left|
\bigcup_{\candseq:\score(\candseq)>\tau}\mathsf{span}(\candseq)
\right|.
\end{equation}}%
To pick a $\tau$ at which to compute coverage for a member book, we calibrate against its matched control book.
The control's top score is how high matched non-member prose reaches, so we use it to set a pair-specific $\tau^{(\mathrm{book})}$.
By construction, no control sequence scores above it: $\tau^{(\mathrm{book})}$ is a measured maximum, not a certified level (\S\ref{app:census:calibration}).\looseness=-1

\begin{figure}[t]
  \centering
  \vspace{-.2cm}
  \includegraphics[width=.85\linewidth]{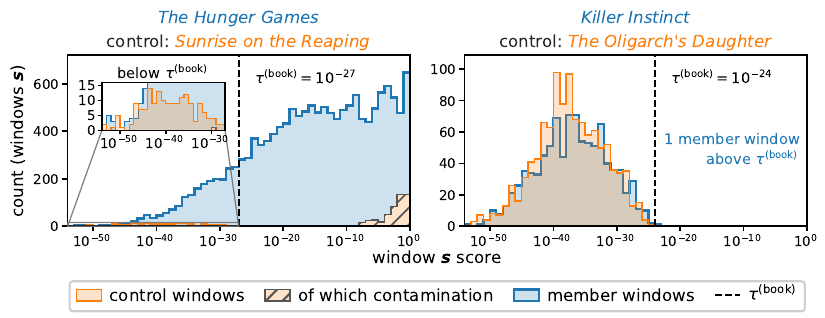}
  \vspace{-.2cm}
  \caption{\textbf{Scored windows $\candseq$ for two book pairs, before and after decontaminating the control.}
  Each histogram shows the member (in-training) book and its matched control.
  We distinguish web-verified contaminants, which consist of text quoted in the control that is drawn from sources from before the training cutoff (and thus are known or likely members). 
  We remove contamination to determine $\tau^{(\mathrm{book})}$. 
  \emph{Killer Instinct} shows effectively no memorization; 
  its windows nearly entirely overlap with the control. 
  Zero-probability windows can't be shown on a $\log$ axis, so are excluded (\S\ref{app:census:calibration}\looseness=-1). 
  }
  \label{fig:census-histograms}
  \vspace{-.3cm}
\end{figure}

\begin{figure}[t]
  \centering
  \includegraphics[width=.85\linewidth]{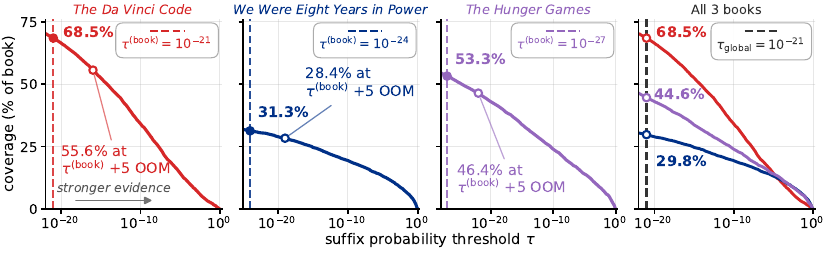}
  \vspace{-.4cm}
  \caption{\textbf{Coverage as the required evidence becomes stronger.}
  Coverage (Eq.~\ref{eq:coverage}) falls as the suffix-probability threshold is raised above each matched $\tau^{(\mathrm{book})}$. 
  To show a common threshold across books, in the rightmost plot, we set $\tau_{\mathrm{global}}$ to the highest (most stringent) $\tau^{(\mathrm{book})}$, $10^{-21}$, and apply it to all of the books.
$\tau_{\mathrm{global}}$ can only raise each book's bar, and so can only reduce coverage.
  (We don't include results for the fourth book, \emph{Killer Instinct}, since  we find effectively no evidence for memorization, as it has only one candidate window $\candseq$ that surpasses its $\tau^{(\mathrm{book})}=10^{-24}$; see Fig.~\ref{fig:census-histograms}.)\looseness=-1}
  \label{fig:book-evidence-strength}
  \vspace{-.4cm}
\end{figure}

\custompar{Decontaminating the control to set $\tau^{(\mathrm{book})}$}
In practice, we often can't just use the control's top score to set $\tau^{(\mathrm{book})}$.
This is because a post-cutoff book may quote earlier material that's present in the training data.
Such sequences contaminate the control pool: 
they're actually members, not non-members, and can score high, inflating $\tau^{(\mathrm{book})}$.
For example, \emph{Sunrise on the Reaping} is the matched control for \emph{The Hunger Games}, and it repeatedly quotes Edgar Allan Poe's ``The Raven.'' 
We therefore remove only web-verified contaminants from the control, leave the in-training book untouched, and set $\tau^{(\mathrm{book})}$ from the highest-scoring remaining own-prose sequence.
For Llama~3.1 70B, $\tau^{(\mathrm{book})}$ ranges from $10^{-27}$ to $10^{-21}$ (Fig.~\ref{fig:book-evidence-strength}).
These numbers are so small that they have unreliable floating-point mantissas;
to be conservative, we discard the mantissa and round up to the next decade (\S\ref{app:census:calibration}). 
Fig.~\ref{fig:census-histograms} shows both decontamination and the resulting threshold for two book pairs.\looseness=-1

\custompar{Strength of evidence for memorization}
Fig.~\ref{fig:book-evidence-strength} sweeps $\tau$ upward from each book's respective $\tau^{(\mathrm{book})}$ and plots the corresponding coverage (Eq.~\ref{eq:coverage}). 
We report results using coverage CCDFs because, while the matched threshold $\tau^{(\mathrm{book})}$ lets us make a binary claim for whether a candidate was memorized, the strength of evidence above $\tau^{(\mathrm{book})}$ clearly varies.
We use a normalized measure of evidentiary strength, the \newterm{margin}, $\Delta(\candseq;\tau)=\log_{10}(\score(\candseq)/\tau)$: 
how many orders of magnitude a candidate's score sits above a threshold (\S\ref{app:sec:population:margin}, \S\ref{app:census:results}).
At their respective thresholds, $68.5\%$ of \emph{The Da Vinci Code}, $53.3\%$ of \emph{The Hunger Games}, and $31.3\%$ of \emph{We Were Eight Years in Power} are memorized. 
Raising the bar discards the weakest evidence---the evidence closest to the threshold: 
at $\Delta=5$, $55.6\%$, $46.4\%$, and $28.4\%$ of the three books still clear their respective bars.
For all books, at $\Delta=5$ the lowest scores remain vanishingly small, at most $10^{-16}$.
They support memorization claims, but they also expose a second distinction: 
there are memorized sequences whose probabilities are far too small to be generated under any realistic sampling budget.

%% file: section/iclr/600-definition.tex
\vspace{-.2cm}
\section{Realistic extractability vs.\ memorization}\label{sec:extractable-memorization}
\vspace{-.1cm}

We make the sampling-budget point concrete by computing how many queries it would take to generate a sequence at one of these thresholds.
For a fixed prompt, suppose candidate $\candseq$ has generation probability $p=\score(\candseq)$ on each independent attempt.
After $n$ attempts, $\candseq$ appears at least once with\looseness=-1 
\vspace{-.1cm}
\begin{equation} 
\Pr[\text{at least once}]
=1-(1-p)^n. 
\label{eq:budget-success}
\end{equation} 
Consider a sequence with $p=10^{-21}$---the largest census threshold (\S\ref{sec:experiments:census}). 
Generating it at least once with probability $1-\delta=0.95$ takes $n\approx3\times10^{21}$ attempts (Eq.~\ref{eq:budget-success}, \S\ref{app:extractability:query}); 
at a thousand queries per second, that would take nearly $7\times$ the age of the universe.
Calling this ``extractable memorization'' is clearly a misnomer.
We therefore suggest a revised definition for \newterm{extractable memorization}:\looseness=-1

\begin{definition}[Extractable memorization]\label{def:extractable-memorization}
Fix an extraction procedure (Def.~\ref{def:extraction-procedure}), a query budget $n$, and a tolerance $\delta$.
A candidate $\candseq$ is \newterm{extractably memorized} if:
\begin{enumerate}[leftmargin=.75cm,itemsep=-2pt,topsep=-2pt]
\item \emph{it's memorized}: its score clears the matched-comparison threshold, providing calibrated evidence of memorization; and
\item \emph{it's realistically reproducible}: a fixed procedure generates it within $n$ queries with probability at least $1-\delta$, whether by sampling or by another decoding procedure of comparable cost.\footnote{For instance, \citet{cooper2026nv} extracts low-probability sequences at the cost of ${\approx}20$ samples (\S\ref{app:extractability:deterministic}).\looseness=-1}
\end{enumerate}
\end{definition}

\begin{figure}[t]
  \centering
  \vspace{-.2cm}
  \includegraphics[width=.9\linewidth]{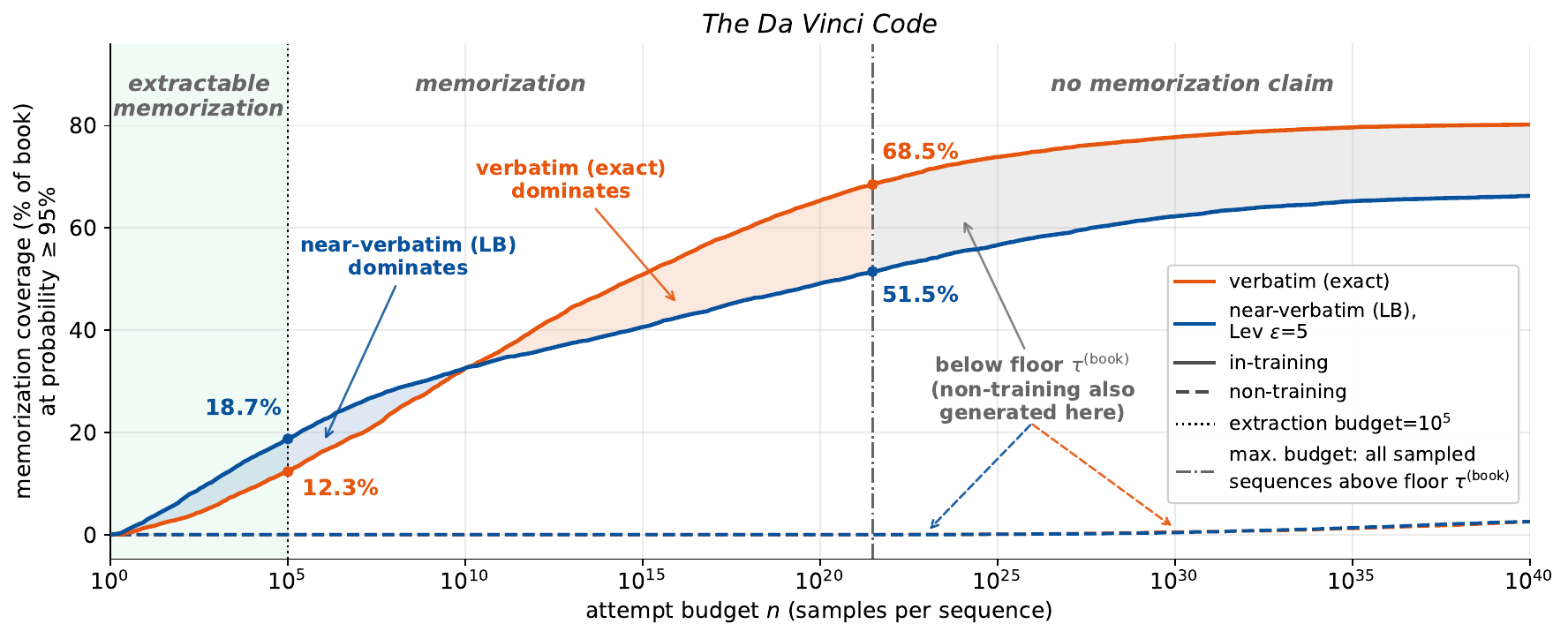}
  \vspace{-.3cm}
  \caption{\textbf{Memorization and realistic extraction for \emph{The Da Vinci Code}, Llama~3.1 70B.}
  Inspired by \citet{hayes2025measuringmemorizationlanguagemodels}, we plot book coverage (Eq.~\ref{eq:coverage}) with probability at least $95\%$ within attempt budget $n$.
  Candidates reachable within $10^5$ queries are extractably memorized ($12.3\%$ verbatim, $18.7\%$ near-verbatim), and candidates above $\tau^{(\mathrm{book})}$ but beyond that budget are memorized but not extractable ($68.5\%$ and $51.5\%$ at $n\approx3\times10^{21}$, the budget $\tau^{(\mathrm{book})}$ implies).
  Candidates below $\tau^{(\mathrm{book})}$ support no memorization claim since non-training text is also generated there.}
  \label{fig:davinci-realistic-extraction}
  \vspace{-.3cm}
\end{figure}

The two requirements answer different questions. 
Fig.~\ref{fig:davinci-realistic-extraction} makes the distinction clear, extending the budget accounting from a single sequence to coverage of a book (Eq.~\ref{eq:coverage}).  
We compare verbatim probabilistic extraction, computed exactly by teacher forcing, against its near-verbatim variant with a maximum Levenshtein distance of $5$ tokens, computed as a lower bound~\citep{cooper2026nv}.
Near-verbatim coverage initially dominates, but at lower probabilities the exact verbatim computation captures mass that the lower bound misses.
For illustration, we take $10^5$ samples as a realistic query budget.
The extraction within that budget is significantly smaller than the total memorization revealed by the matched comparison (at the budget associated with $\tau^{(\mathrm{book})}$).
The same view for \emph{We Were Eight Years in Power} and \emph{The Hunger Games} shows similar results (\S\ref{app:extractability:census}).\looseness=-1

%% file: section/iclr/700-takeaways.tex
\vspace{-.2cm}
\section{Conclusion}\label{sec:takeaways}
\vspace{-.1cm}

A memorization claim is only valid relative to a baseline of general predictability.
We make this explicit with matched comparisons that decide when a generation probability is high enough to support a memorization claim (\S\ref{sec:prelim}).
We propose a conformal test that certifies a false-positive rate (\S\ref{sec:test}) for population experiments (\S\ref{sec:experiments:population}), and a census for single documents (\S\ref{sec:experiments:census})---both of which calibrate a threshold from the generation probabilities of matched non-members.
Clearing that threshold licenses calling a candidate's generation probability an \emph{extraction} probability (Box~\ref{box:extraction-claim:main}). 
But those probabilities can be vanishingly small, so we refine the definition of extractable memorization to require two things: 
surpassing a calibrated threshold \emph{and} generation within a realistic query budget (\S\ref{sec:extractable-memorization}). 
The consequences reach beyond our own experiments.

\custompar{Perfect matched controls don't always exist (and that's okay)}
Calibrated claims depend on distributional matching between candidates and controls (\S\ref{sec:test}). 
But in some cases, a perfect match isn't possible;
the closest available non-members are a relatively loose match, so the null is more general.
For instance, Sherlock Holmes stories are public domain and almost certainly in LLM training data;
but Arthur Conan Doyle will never write a new one, so no matched non-member exists to make calibrated claims.
One can still calibrate against a more general null (e.g., other non-training mysteries, a wide variety of English prose).
Our census results with $\tau_{\mathrm{global}}$ illustrate this: 
calibrating against general non-member prose instead of each book's own control, and taking the most conservative threshold for every book (Fig.~\ref{fig:book-evidence-strength}).
Such claims should be qualified accordingly, with clear reporting about the controls (\S\ref{app:sec:matching}).

\custompar{Using short candidates risks validity}
A short verbatim match can't carry a valid memorization claim on its own.
We see this directly in our experiments: 
at short candidate lengths, matched non-members clear the threshold often enough to produce many false positives (\S\ref{sec:experiments:population}).
Yet a lot of work reports extraction using short matches without running a matched comparison~\citep{chang-etal-2023-speak,karamolegkou2023copyrightviolationslargelanguage,wei2025interrogatingllmdesignfair,liu2026alignmentwhackamolefinetuning}. 
Claims in these settings rest on member generation rates, which aren't sufficient for separating memorization from general predictability.\looseness=-1

\custompar{Extraction is evidence for memorization, when it's calibrated}
While that work risks over-claiming, other recent work claims the opposite: that extraction is unreliable evidence of memorization, because models can verbatim complete text they weren't explicitly trained on~\citep{liu2025forged,morris2025howmuch}.
But this work doesn't account for near-duplicate members contaminating the sequences they treat as non-members (\S\ref{app:sec:matching}),
and, regardless of this issue, that conclusion is too strong.
Valid extraction claims are always relative to a specific extraction procedure and baseline, so a negative result for one procedure doesn't say anything about calibrated claims under another.

\customparnodot{It's time to retire the ``monkey at the typewriter'' argument against valid extraction.\footnote{He was very early at OpenAI, is very tired, and has chronic carpal tunnel. He'll be retiring to Lake Como.}} 
That analogy compares a language model to random chance.
The monkey hits keys at random, so with enough attempts it produces any text at all, even something as long and complex as the complete works of Shakespeare.
The trope mistakenly treats an LLM the same way: 
because generation is non-deterministic, an LLM that produces (near-)verbatim training data is supposedly just getting lucky.
But matched comparisons show crisply that this isn't how LLMs behave.
First, generation isn't purely random: 
for a high-quality model, the general-predictability baseline is fluent, grammatical text, not random gibberish.
Second, whether it's \emph{possible} for an LLM to produce any string was never the right question;
what matters is \emph{how many tries} it takes, relative to that matched baseline.
Extractably memorized text---text that would actually, practically be produced in outputs---can be many, many orders of magnitude above the baseline (\S\ref{sec:extractable-memorization}), far too high to be coincidence.

%% file: section/iclr/900-appendix.tex
\clearpage
\appendix

\input{section/iclr/appendix/000-intro-experiments}
\input{section/iclr/appendix/100-prelim}
\input{section/iclr/appendix/200-conformal-app}

\input{section/iclr/appendix/300-population}
\input{section/iclr/appendix/400-census}
\input{section/iclr/appendix/500-extractability}
\input{section/iclr/appendix/550-assumptions}
\input{section/iclr/appendix/600-prior-work}

%% file: section/iclr/appendix/000-intro-experiments.tex
\vspace{-.2cm}
\section*{Appendix Table of Contents}
\vspace{-.5cm}
{
\small
\startcontents[appendix]
\printcontents[appendix]{l}{1}{\setcounter{tocdepth}{3}}
}
\clearpage

\section{Motivating examples}\label{app:sec:intro}

In this appendix, we provide details on the experiments discussed in \S\ref{sec:intro} concerning ``parrot'' prompts (\S\ref{app:sec:intro:parrot}) and the extraction procedure of \citet{schwarzschild2024rethinking} (\S\ref{app:sec:intro:acr}).

\subsection{``Parrot'' prompts experiment}\label{app:sec:intro:parrot}

The experiment in Fig.~\ref{fig:parrot} involves OLMo~2 32B Instruct, and we use the following prompt template: 
\begin{center}
\texttt{Repeat the text after the newline exactly, word for word. Do not continue, summarize, or add anything; output only that text:\textbackslash n\{X\}}
\end{center}
where \texttt{\{X\}} is the first $50$ tokens of either a Wikipedia entry in OLMo 2's training data (a member) or a post-cutoff Wikipedia entry (a non-member).
We run this prompt template for $3{,}000$ member sequences and $3{,}000$ non-member sequences. 
For each prompt, we greedy generate $55$ tokens---slightly more than $50$, so that the decoded text is at least as long as \texttt{\{X\}} even when the tokenization differs---decode to text, truncate the decoded result to \texttt{\{X\}}'s character length, and require a character-exact match to \texttt{\{X\}}. 
(It's easiest to do this, rather than comparing in token space, since the inclusion of the prompt template with \texttt{\{X\}} may result in the portion containing \texttt{\{X\}} having a different tokenization than the generation of \texttt{\{X\}}, even if the underlying text is identical.)

Each scored sequence---here, with greedy decoding and checking an exact match---is an independent success/failure trial, so each reported rate is a binomial proportion.
In Fig.~\ref{fig:parrot}, we therefore show $95\%$ Clopper--Pearson intervals. 
Intuitively, the interval reflects how much the measured rate could shift had we drawn a different $3{,}000$ sequences.
The member rate is $70.1\%$ (95\% CP interval $[68.4, 71.7]$) and the non-member rate is $70.6\%$ ($[69.0, 72.3]$).
The two intervals overlap, so the member and non-member rates are statistically indistinguishable.

Pretend that one assumes this is a valid extraction procedure (even though it clearly isn't). 
Because it behaves the same on members and non-members, the matched comparison on its own shows that successful generation provides no basis for claiming memorization.

\subsection{Generating verbatim non-training data with ACR}\label{app:sec:intro:acr}

The same issue can arise in a less contrived procedure.
\citet{schwarzschild2024rethinking} search for an adversarially optimized prompt that elicits a target sequence and deem extraction successful when that prompt is shorter than the target.
To measure this, they introduce \newterm{adversarial compression ratio (ACR)}, which divides the length (in tokens) of the target by that of the prompt; 
extraction is successful if that ratio exceeds $1$. 
We were able to achieve success for both training and non-training data; 
in each case, however, we had to make some changes to the released code, which we discuss below.\looseness=-1

They define success by greedy decoding, but the code implementation tests it by teacher-forcing the $\argmax$ within GCG~\citep{zou2023gcg}.
GCG searches for a prompt by minimizing the mean per-token cross-entropy of the target;
the implementation stops as soon as the $\argmax$ agrees with the target at every position.
In arithmetic the two criteria coincide: 
if the $\argmax$ is the target token at each step, greedy decoding stays aligned with the target throughout generation.
However, teacher forcing and greedy decoding can exhibit slight numeric differences.
(The former uses a single forward pass over the whole sequence with no KV cache; the latter does incremental generation with a KV cache.)
Because the search halts at the first exact match, we observed that its solutions often sit on a very thin $\argmax$ margin, where a small numerical difference can flip one token and cause the continuation to diverge from that point on. 
(This $\argmax$ criterion is fragile by construction.)\looseness=-1

In practice, we found that prompts the released code marks as successful (via $\argmax$ teacher forcing) often fail to reproduce the target verbatim under greedy generation.
The released code does call \texttt{generate} once a prompt succeeds, but only logs the result: 
it's never compared against the target, and never feeds back into the success flag, the reported ACR, or the saved results.
(This is how we observed the target-generation failures to begin with.)
Across $16$ runs ($8$ targets on each of Llama~2 13B and Llama~2 13B Chat), we observed $11$ such failures:
prompts marked as successful that don't greedily generate the target verbatim, and therefore fail the paper's own definition. 
(We verified this against the released code on our own runs; 
we don't claim that the published numbers reflect this error, since we don't know what the underlying results were.)

We therefore relaxed the success condition rather than the definition.
We keep teacher forcing, but instead of requiring the $\argmax$ to match at every position, we require the target's teacher-forced probability to be high: success when the geometric-mean per-token probability of the target is at least $b$.
(This is analogous to doing probabilistic extraction~\citep{hayes2025measuringmemorizationlanguagemodels} instead of discoverable extraction~\citep{carlini2023quantifying}, but in the context of GCG.) 
This is equivalent to a bound on GCG's own objective, $\mathrm{loss} \leq -\ln b$, and so doesn't depend on target length.
We set $b = 0.9$, so that a successful prompt is one from which the target would be reproduced with high probability under sampling.
The length search is the same, so the reported ratio is still $\mathrm{ACR} = |y|/|x^*|$, where $y$ is the target and $x^*$ is the optimized prompt.

\begin{figure}[t]
\centering
\begin{subfigure}[t]{.49\linewidth}
  \vspace{0cm}
  \centering
  \includegraphics[width=\linewidth]{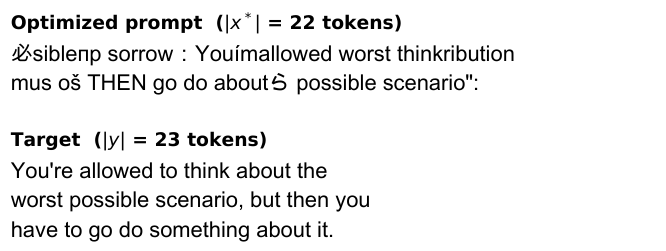}
  \vspace{-.1cm}
  \caption{Post-cutoff (Brunson).}
  \label{app:fig:acr-prompt:brunson}
\end{subfigure}
\begin{subfigure}[t]{.49\linewidth}
  \vspace{0cm}
  \centering
  \includegraphics[width=\linewidth]{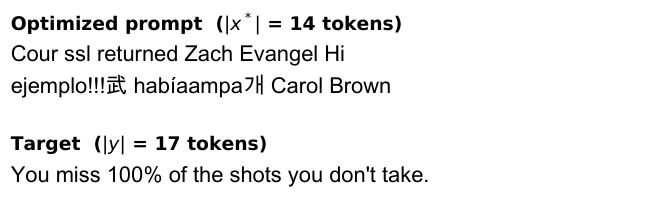}
  \vspace{.14cm}
  \vspace{-.1cm}
  \caption{Pre-cutoff (Gretzky).}
  \label{app:fig:acr-prompt:gretzky}
\end{subfigure}
\vspace{-.2cm}
\caption{\textbf{Optimized prompts that satisfy the ACR criterion at $b=0.9$.}
Each panel shows the GCG-optimized prompt above its target, with token counts.
Both prompts are shorter than their targets, so both clear $\mathrm{ACR}>1$ under \citet{schwarzschild2024rethinking}'s criterion, and both elicit the target with geometric-mean per-token probability above $0.9$ using Llama~2 13B.
Note that ACR doesn't place any constraints on prompt/target overlap.
The Brunson prompt reuses target content words; the Gretzky prompt doesn't.}
\vspace{-.3cm}
\label{app:fig:acr-prompts}
\end{figure}

With this relaxed success condition, we applied the procedure to the Wayne Gretzky line, ``You miss 100\% of the shots you don't take'' (one of the shipped targets in the repository), and we achieved success using Llama~2 13B: a $14$-token prompt for the $17$-token target, $\mathrm{ACR} = 1.21$ (Fig.~\ref{app:fig:acr-prompt:gretzky}).
But we also succeeded for a similar-length statement Jalen Brunson made following the Knicks' $29$-point comeback against the Spurs in the 2026 NBA Finals:
``You're allowed to think about the worst possible scenario, but then you have to go do something about it''---a $22$-token prompt for the $23$-token target, $\mathrm{ACR} = 1.05$ (Fig.~\ref{app:fig:acr-prompt:brunson}).
The latter post-dates the model's training cutoff.
This doesn't show that the model failed to memorize the Gretzky quotation;
it shows that satisfying the procedure's criterion doesn't, by itself, warrant a memorization claim.

We also observed that the optimized prompts often contain words from the target itself.
This is visible even in \citet{schwarzschild2024rethinking}'s own teaser figure: for the $12$-token target ``To be or not to be, that is the question.'', the $4$-token prompt they report is \texttt{urgesTOBE quote!}, which hands the model the target's opening words (their Fig.~1).
The Brunson prompt is a far more pronounced instance of the same thing (Fig.~\ref{app:fig:acr-prompt:brunson}).

The point is, there is nothing in GCG itself that prevents leaking the target in the prompt. 
One could address this by changing the definition of $\mathrm{ACR}$ by adding a penalization term for overlap:
replace $|x^*|$ in the denominator with $|x^*| + \lambda o$, where $o$ counts the number of target tokens (or some other overlap count) the prompt reuses and $\lambda \geq 0$ sets the cost.
Setting $\lambda = 0$ recovers the original $\mathrm{ACR}$ definition.
Note that doing this would affect the results on Brunson, but not Gretzky;
depending on the overlap count and penalty term, it could push the Brunson $\mathrm{ACR}$ below $1$ (i.e., it would no longer be a success). 
(We raise this as an observation, rather than a metric we've validated.)

%% file: section/iclr/appendix/100-prelim.tex
\vspace{-.2cm}
\section{Extended preliminaries}\label{app:sec:prelim}
\vspace{-.1cm}

In this appendix, we give the preliminaries in full. 
Let $\vocab$ be a token vocabulary and $\vocab^*$ the set of finite token sequences.
An \newterm{autoregressive language model} $\params$ defines a next-token distribution $q_\params(\cdot \mid \vv)$ for each $\vv \in \vocab^*$.
Run autoregressively from a \newterm{prompt} $\promptseq$, a \newterm{decoding policy} $\dec$ induces a distribution $\channel(\cdot \mid \promptseq)$ over \newterm{continuations}.
The \newterm{base} policy makes no adjustment ($\channel = q_\params$): 
it samples from the whole vocabulary at each step.
\newterm{Top-$k$ decoding} restricts at each step to the $k$ highest-probability tokens, renormalizes, and samples.
By contrast, \newterm{greedy decoding} is deterministic: 
it selects the highest-probability ($\argmax$) token at each step. 
Each policy's next-token distribution depends only on the preceding tokens, so it factorizes autoregressively over continuations. 
For a prompt $\promptseq$ and an $\eventlen$-length continuation $\outseq = (\outtok_1, \ldots, \outtok_{\eventlen}) \in \vocab^{\eventlen}$, the generation probability is 
\begin{align}
\label{app:eq:generation-prob}
\textstyle 
\channel(\outseq \mid \promptseq)
\,\coloneqq\,
\prod_{t=1}^{\eventlen} \channel(\outtok_t \,\mid\, \promptseq, \outseq_{<t}).
\end{align}
For greedy decoding ($\dec=\greedy$), $q_{\params,\greedy}(\outseq \mid \promptseq) \in \{0,1\}$: 
the unique, deterministic $\eventlen$-token continuation has probability $1$, and all other $\eventlen$-token continuations have probability $0$.

Loosely speaking, \newterm{memorization} of a specific training sequence $\member$ means that the model assigns unusually high probability to that sequence, indicating that information about $\member$ is encoded in the model. 
\newterm{Extraction} is a narrower phenomenon: 
it is the reproduction of a memorized sequence in the model's outputs via prompting.
When a sequence is extractable, the model assigns it high conditional probability $\channel(\member \mid \promptseq)$, and reliably reproduces $\member$ when prompted with $\promptseq$.
Not all memorized sequences are extractable~\citep{carlini2021extracting}. 
A sequence can be encoded in the model yet not reliably reproduced under any prompt~\citep{nasr2023scalable,nasr2025scalable}. 
And studying memorization does not require prompting in general; 
other approaches, such as membership inference, study memorization using different methods~\citep{shokri2017membership,carlini2019secret,carlini2022membership,hayes2025strongmia}.\looseness=-1

Memorization is only possible for data the model was trained on, so we need to make clear what it means for a sequence to be a ``member'' of the training data, and what it means not to be (\S\ref{app:sec:prelim:membership}).
We then pin down a definition for an extraction procedure, which formalizes the intuition above (\S\ref{app:sec:prelim:extraction}).

\input{section/iclr/appendix/110-prelim-membership}
\input{section/iclr/appendix/120-prelim-extraction}

%% file: section/iclr/appendix/110-prelim-membership.tex
\subsection{Members, non-members, and the membership indicator}\label{app:sec:prelim:membership}

A language model is trained on a \newterm{training dataset} $\dataset$, which is a fixed collection of token sequences.
We make no commitment to the structure or format of $\dataset$:
it may be a set of documents of varying or fixed lengths, or have any other organization amenable to being ingested by a training procedure. 
By definition, memorization is only possible for content present in $\dataset$:
a model cannot have memorized something it never saw during training. 
A sequence $\candseq$ that an extraction procedure deems to be memorized is therefore necessarily a \newterm{member} of the model's training data.
Failure to extract $\candseq$ is \emph{not} evidence that $\candseq$ is a \newterm{non-member}: 
not all training data are memorized, and not all memorized training data are necessarily extractable under a given procedure. 
Throughout, we refer to sequences $\candseq$ as \newterm{candidates} under test for memorization.\looseness=-1

\custompar{Including near-duplicates when defining membership} 
Near-duplicates of training sequences are functionally members of the training data: 
the model has seen the content, whatever tokenization or minor syntax variation happened to accompany it~\citep{lee2022dedup}. 
As a result, restricting a definition of training-data membership to exact token sequences is brittle for language models~\citep{zhang2023counterfactual, maini2024dataset, hayes2025strongmia}:
it would misclassify near-duplicates of members as non-members. 
We therefore define membership at the level of near-verbatim variations of a sequence, rather than at the granularity of the sequence itself.\looseness=-1

To make ``near-verbatim'' precise, fix a dissimilarity function $\distance : \vocab^\eventlen \times \vocab^\eventlen \to \R_{\ge 0}$ that measures how different two length-$\eventlen$ sequences are, such as token-level edit  distance~\citep{navarro2001string} or character-level BLEU score~\citep{bleu}.\footnote{While not strictly necessary, for convenience we restrict to length-$\eventlen$ sequences.
    This matches how extraction procedures typically compare candidate sequences and generated sequences of common length~\citep{cooper2026nv}.\looseness=-1
}
For tolerance $\tol \ge 0$, the \newterm{near-verbatim $\tol$-ball} around a length-$\eventlen$ sequence $\seq$ is 
\begin{align}
\label{eq:ball}
\ball{\tol}{\seq} \coloneqq \{\outseq \in \vocab^\eventlen : \distance(\outseq, \seq) \le \tol\}, 
\end{align}
the length-$\eventlen$ sequences within $\tol$ of $\seq$.\footnote{We omit $\distance$ from the notation; 
once set, we consider it fixed, while different analyses may vary $\tol$.
} 
Setting $\tol = 0$ recovers the singleton, verbatim $\ball{0}{\seq}=\{\seq\}$.\footnote{Memorization research tends to restrict to such simple dissimilarity functions, applied with low tolerances $\tol$, in part because it is hard to design a semantically sound similarity function for natural text that captures memorization signal without conflating it with generalization~\citep{barbero2025extractingalignmentdataopen}.}

\begin{definition}[Membership indicator]
\label{app:def:membership-indicator}
An arbitrary length-$\eventlen$ sequence $\seq$ is a near-verbatim training-data member at tolerance $\tol$ if some length-$\eventlen$ sequence in $\dataset$ is within distance $\tol$ of $\seq$; 
otherwise it is a non-member.
The corresponding \newterm{membership indicator} is
\begin{align}
\label{app:eq:membership-indicator}
\memberind{\seq} \coloneqq \1\{\dataset \cap \ball{\tol}{\seq} \ne \varnothing\},
\end{align}
where $\dataset \cap \ball{\tol}{\seq}$ is understood as the result of searching $\dataset$ for any length-$\eventlen$ sequence within $\tol$ of $\seq$.
When $\dataset$ contains documents of varying lengths, this amounts to checking length-$\eventlen$ subsequences.\looseness=-1 
\end{definition} 
When $\memberind{\seq} = 1$, $\ball{\tol}{\seq}$ has training support: 
at least one training sequence falls within distance $\tol$ of $\seq$. 
Conversely, when $\memberind{\seq} = 0$, no sequence in the $\tol$-ball $\ball{\tol}{\seq}$ was present in training. 
One practical consequence is that non-membership of $\seq$ is deliberately stronger than simply requiring that $\seq$ is held-out from training.
A candidate $\candseq$ whose $\tol$-ball contains a training near-duplicate is \emph{not} a valid non-member, regardless of whether $\candseq$ itself appears in the training data.\looseness=-1 

%% file: section/iclr/appendix/120-prelim-extraction.tex
\subsection{Defining an extraction procedure}\label{app:sec:prelim:extraction}

Extraction involves prompting a model to attempt to reproduce a candidate sequence, where the candidate is either known or suspected to be a member of the training data. 
There are many approaches in the literature that satisfy this loose definition, but they all have the same five components. 
The first two are the model $\params$ and the decoding policy $\dec$.
The remaining three are a prompt-selection rule that picks the prompts the model and decoding policy are tested against,
a scoring procedure that evaluates how easily the model and decoding policy reproduce the candidate under those prompts,
and a verification procedure that judges whether the candidate is memorized based on its score. 

\custompar{Prompt-selection rule}
For a candidate $\candseq$, a \newterm{prompt class} for $\candseq$ is a finite set of prompts $\promptclass(\candseq) \subseteq \vocab^*$. 
A prompt class can be a singleton (e.g., just the $50$-token natural prefix of a training sequence), or a larger set.
For instance, the prompt class can be extended to all natural prefixes of $\candseq$:  
\(
\promptclass^{\textrm{nat}}(\candseq) \coloneqq \{\npromptseq{0}, \npromptseq{1}, \ldots, \npromptseq{\maxlen}\},
\)
progressively longer prefixes preceding $\candseq$ in the original training document, where $\npromptseq{0} = \varnothing$ is the empty string and the longest prefix has length $\maxlen$~\citep{carlini2023quantifying, cooper2025books}. 
More generally, a prompt class is a search space: 
the set of prompts an extraction procedure can navigate to find queries that elicit the candidate. 
A \newterm{prompt-selection rule}\looseness=-1 
\begin{align}
\label{app:eq:prompt-rule}
\promptrule : \vocab^\eventlen \to 2^{\vocab^*}
\end{align}
maps each candidate $\candseq$ to a finite prompt class $\promptclass(\candseq) \coloneqq \promptrule(\candseq) \subseteq \vocab^*$.\footnote{In practice, each prompt is bounded in length by the model's context window, possibly smaller for specific class constructions (e.g., natural-prefix classes are bounded by the longest prefix of $\candseq$ in the training document).
} 

\custompar{Scoring procedure} 
A \newterm{scoring procedure} $\score : \vocab^\eventlen \to \R$ measures how easily a candidate $\candseq$ is generated under $\params$ and $\dec$ from the prompts in class $\promptclass(\candseq)$, where higher means more easily generated. 
For a single prompt $\promptseq$, the canonical score is the near-verbatim probability of generating $\candseq$, $\channel(\ball{\tol}{\candseq} \mid \promptseq)$, or an approximation of that probability (e.g., a beam-search-based lower bound~\citep{cooper2026nv}). 
(This captures verbatim extraction, when $\tol=0$, and also greedy decoding, for which the score is binary---$1$ if $\candseq$ is generated verbatim, otherwise $0$.)
Over a prompt class for candidate $\candseq$, an \newterm{aggregation rule} $\agg$ collapses per-prompt scores into the candidate's score:\looseness=-1
\begin{equation}
\label{app:eq:candidate-score}
\score(\candseq) \,=\, \agg_{\promptseq \in \promptrule(\candseq)} \score_\promptseq(\candseq). 
\end{equation} 
For instance, an aggregation rule could return the maximum over the prompt class:
the prompt that most easily reproduces an output in $\ball{\tol}{\candseq}$.

\custompar{Verification procedure}
A \newterm{verification procedure} takes a $\score$ and outputs a decision about whether a candidate $\candseq$ is memorized, i.e., $\verify : \R \to \{\text{memorization claim}, \text{no memorization claim}\}$. 
We construct such a $\verify$: 
a conformal test that calibrates a threshold on the score by comparing to the scores of matched non-members, determining when a score is high enough to claim memorization (\S\ref{sec:test}).
Put differently, the score this test evaluates is a generation probability, and clearing the calibrated threshold is what licenses calling it an extraction probability.\looseness=-1 

Putting all five components together:
\begin{definition}[Extraction procedure]
\label{app:def:extraction-procedure}
An \newterm{extraction procedure} is a tuple
\(\params,\, \dec,\, \promptrule,\, \score,\, \verify\), 
where model $\params$ and decoding policy $\dec$ configure the distribution $\channel$;
$\promptrule$ specifies the prompts used for each candidate sequence $\candseq$ (Eq.~\ref{app:eq:prompt-rule});
$\score$ returns the score for $\candseq$ (computed under the fixed model $\params$, decoding policy $\dec$, and prompt-selection rule $\promptrule$);
and, based on its score, $\verify$ either deems $\candseq$ memorized or makes no claim.
\end{definition}

We later show that prior methods in the literature are instantiations of this definition (\S\ref{app:sec:prelim:prior}).\looseness=-1

%% file: section/iclr/appendix/200-conformal-app.tex
\clearpage
\section{Extended details on the conformal test for memorization}\label{app:sec:conformal}

In this work, we use the (near-)verbatim generation probabilities of matched non-training data to calibrate a threshold---a probability cutoff above which (near-)verbatim generation is claimed to be evidence of memorization.
Calibration makes the inferential chain behind this claim explicit, which prior work has left implicit (Box~\ref{box:extraction-claim:main}).\looseness=-1

Here, we provide additional details on the conformal test for memorization, which formalizes threshold calibration for making claims about memorization concerning populations, certified to a chosen false positive level.
The single-document census procedure answers a different question.
It concerns fixed documents, where the question is not what proportion of a sampled population is memorized, but how much of \emph{this} particular work is memorized.
As a result, while the census procedure follows the same spirit of matched comparisons for threshold calibration as the conformal test, it's a fundamentally different object.
There's no population to sample from in that context, so there is no analogous FPR guarantee to provide.
The census threshold $\tau^{(\mathrm{book})}$ is the maximum score over the control book's windows (\S\ref{app:census:calibration}).
Those windows aren't exchangeable units: 
using the sliding-window procedure from \citet{cooper2025books}, adjacent windows overlap so they share most of their text;
windows from the same paragraph are related even when they don't overlap.
So $\tau^{(\mathrm{book})}$ is a measured maximum rather than a rank statement at level $1/(N+1)$ for $N$ the window count.

Concerning the conformal test, here we provide more detail on relevant notation and the setup (\S\ref{app:sec:conformal:notation}).
We provide a full proof of Prop.~\ref{prop:conformal-validity}, concerning finite-sample validity of the conformal test (\S\ref{app:conformal-proof}), and more details on the assumptions and practical limitations of the proved guarantee (\S\ref{app:sec:matching}). 
In particular, that proof requires only exchangeability of the candidate's score and the $N$ control scores.
But, in practice, we additionally require that these scores be independent and identically distributed, which our sampling design provides (\S\ref{sec:experiments:population}).
This lets us provide additional results about the population FPR of the calibrated threshold, from which we're able to provide valid Beta intervals around that FPR, and a guarantee \emph{conditional} on the particular control sample our experiments use, not just a marginal one (\S\ref{app:sec:conformal:alpha-star}).
We also clarify why we can reuse the same control pool across claims for different candidates (\S\ref{app:sec:conformal:multiplicity}).

\input{section/iclr/appendix/210-conformal-notation}
\input{section/iclr/appendix/220-conformal-proof}
\input{section/iclr/appendix/240-matching}
\input{section/iclr/appendix/250-conformal-beta-dist}
\input{section/iclr/appendix/260-conformal-multiplicity}

%% file: section/iclr/appendix/210-conformal-notation.tex
\subsection{Notation and summary of relevant quantities}\label{app:sec:conformal:notation}

\begin{figure}[h!]
\vspace{-.3cm}
  \centering
  \begin{subfigure}[t]{\linewidth}\centering
    \includegraphics[width=0.78\linewidth]{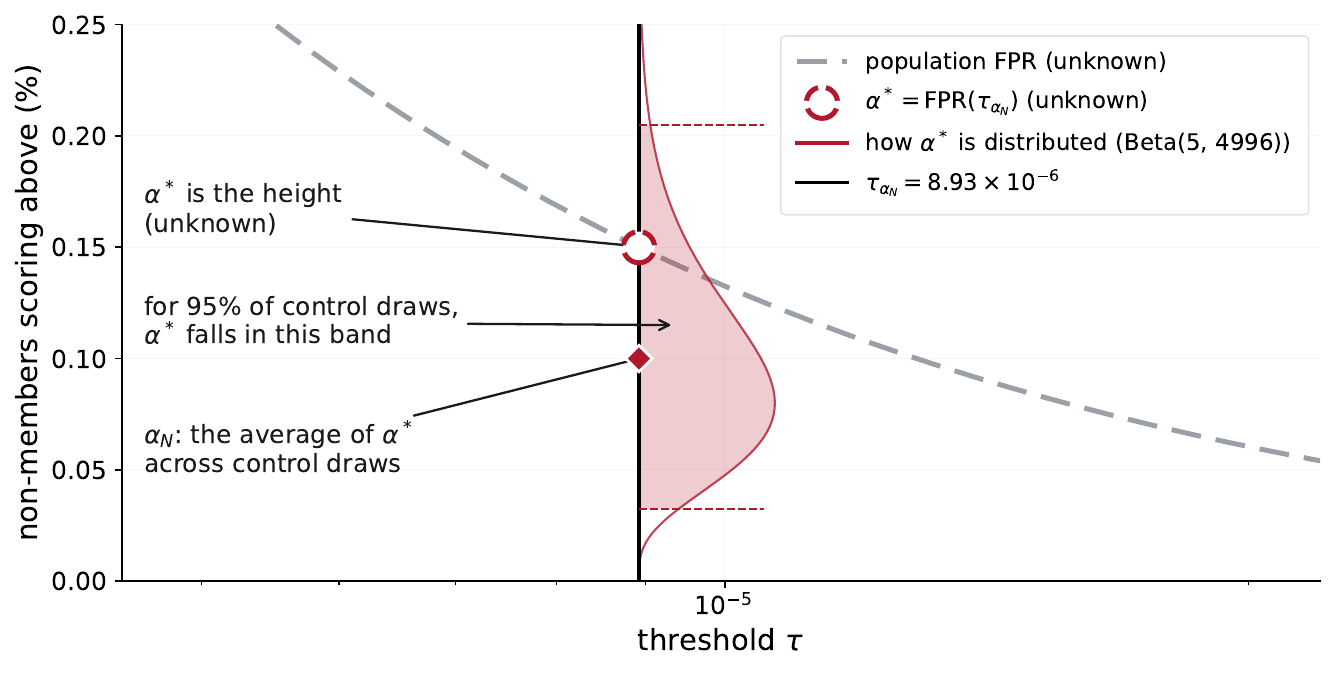}
    \vspace{-.25cm}
    \caption{Population FPR curve (illustrative) and the calibrated $\tau_{\alpha_N}$}
    \label{fig:fpr-objects-curve}
  \end{subfigure}
  \vspace{.2cm}
  \begin{subfigure}[t]{\linewidth}\centering
    \includegraphics[width=0.78\linewidth]{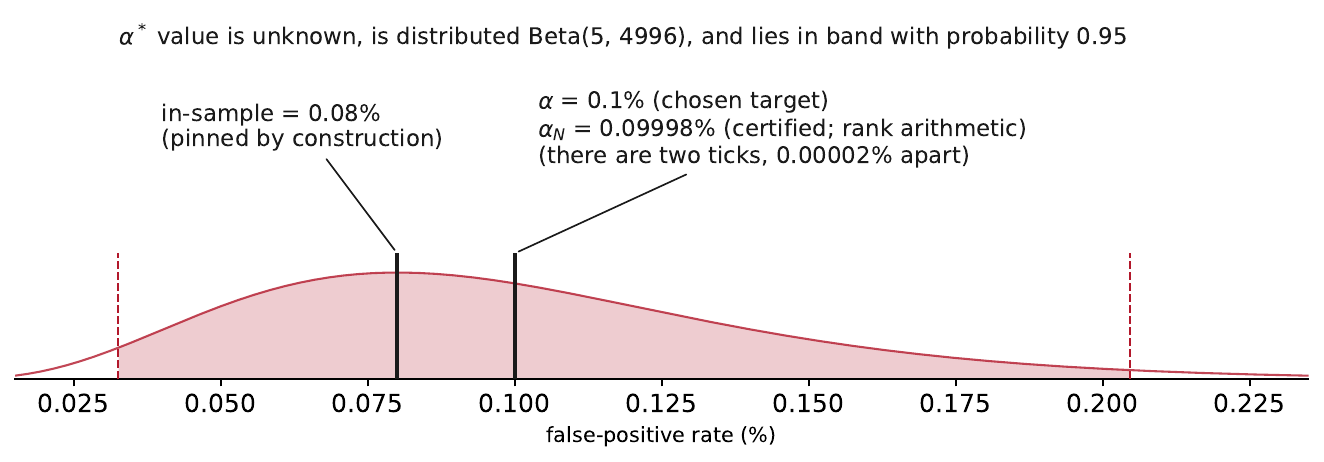}
    \vspace{-.25cm}
    \caption{The four FPR quantities on one axis}
    \label{fig:fpr-objects-axis}
  \end{subfigure}
  \vspace{-.4cm}
  \caption{\textbf{Relating the four FPR quantities.} 
  We use results from \S\ref{sec:experiments:population} to illustrate: 
  Wikipedia and OLMo~2 32B base, $50$-token natural prefixes as prompts and their corresponding $50$-token suffixes as the targets for extraction; $N=5000$ matched non-member controls and target FPR $\alpha =0.1\%$. 
  This allows for at most $K = 5$ false positives, so the certified level is $\alpha_N = K/(N+1) = 5/5001$ and the threshold $\tau_{\alpha_N}$ is the $5$th-largest control score (Prop.~\ref{prop:conformal-validity}; 
  proof in \S\ref{app:conformal-proof}).
  (\textbf{a}) The population FPR curve $\mathrm{FPR}(\tau)$ is fixed by the extraction procedure (Def.~\ref{app:def:extraction-procedure}) and the non-member distribution, but computing it needs that whole distribution; 
  in practice, we have only $N$ draws from it, so it's unknown (dashed; shape illustrative only).
  Our controls set $\tau_{\alpha_N}$, and $\alpha^* = \mathrm{FPR}(\tau_{\alpha_N})$ is the curve's height there  
  (one point on that unknown curve).
  We do know $\alpha^*$'s distribution across control draws: $\alpha^* \sim \mathrm{Beta}(K, N{-}K{+}1)$, which here is $\mathrm{Beta}(5, 4996)$ \citep{vovk2012conditional} (\S\ref{app:sec:conformal:alpha-star:beta-law}).
  The red curve is that distribution (drawn along the threshold line); 
  its shaded part is where $\alpha^*$ falls for $95\%$ of draws (since $\alpha^*$ is unknown, the dashed circle is drawn at an arbitrary point inside the $95\%$ band). 
  The diamond marks the mean, which is exactly the certified level $\alpha_N$ (Eq.~\ref{eq:alpha-star-mean}).
  (\textbf{b}) All four FPR quantities on one axis, for this same $\tau_{\alpha_N}$: 
  the target $\alpha$ and the in-sample rate, which (a) does not show, alongside $\alpha_N$ and the distribution of $\alpha^*$ (the same as in (a), laid along the axis).
  The $95\%$ band for $\alpha^*$ runs from $0.03\%$ to $0.20\%$; 
  the other three values sit within $0.02$ percentage points of one another.
  $\alpha^*$ is the one that depends on the control draw (it varies across draws and is unknown for our specific draw), so it can't be marked. 
  The other three and the band are fixed once $\alpha$ and $N$ are chosen, before any control is drawn.
  This is why the band isn't a confidence interval: 
  the interval is fixed, $\alpha^*$ isn't.\looseness=-1}
  \label{fig:fpr-objects}
\end{figure}

There are several different FPR objects related to the conformal test.
Here, we walk through exactly what each one is, so that it's very clear what the conformal guarantee is about. 
Fig.~\ref{fig:fpr-objects} provides a visualization of what this section describes in prose. 
There are four key objects to distinguish:
the target FPR, the certified FPR, the in-sample FPR, and the population FPR. 
We describe them all, and then how they relate to the calibrated threshold. 
We'll use our results on Wikipedia as a running example.\looseness=-1

\paragraph{The target and certified FPR.} 
As noted in \S\ref{sec:test}, for the conformal test, one picks a \newterm{target FPR} $\alpha$ for calibration.
From this, the test derives a \newterm{conformal $p$-value} denoted $\widehat p$ (Eq.~\ref{eq:conformal-pvalue});
for a candidate sequence $\candseq$, we claim memorization when $\widehat p(\candseq) \leq \alpha$. 
Empirically, the $p$-value comes from a rank cutoff in the size-$N$ control sample.
As a result, it may not be possible to obtain exactly the FPR $\alpha$ in practice; 
it depends on what $N$ is. 
A finite control sample can only certify an FPR in steps of $1/(N+1)$, because the cutoff has to be one of the $N$ control scores.
In other words, the rank $K=\lfloor\alpha(N+1)\rfloor$ provides the cutoff, from which we derive the largest attainable \newterm{certified FPR} $\alpha_N = K/(N+1) \leq \alpha$.
For instance, consider a control sample from Wikipedia where $N=5000$;
for $\alpha=0.1\%$, $\alpha_N = 5/5001 \approx 0.09998\%$ (Fig.~\ref{fig:fpr-objects-axis}). 
This is also why $1/(N+1)$ is the smallest level at which the test can claim anything: 
since $\widehat p\geq1/(N+1)$, setting $\alpha<1/(N+1)$ gives $K=0$, and no memorization claim is ever made (the test can never reject).
The floor of $1/(N+1)$ is on how strong a claim the controls can support, not on the certified FPR;
a test that never makes a claim has no false positives (but it also doesn't have any use).\looseness=-1 

The conformal guarantee in Prop.~\ref{prop:conformal-validity} is ultimately about finding an $\alpha_N \leq \alpha$ (\S\ref{app:conformal-proof});
it's a guarantee about the procedure, and has to do with a count on the number of controls in the sample.
This means that $\alpha_N$ is pinned by construction---the rank at which you specify the cutoff for the threshold. 

\paragraph{In-sample FPR.}
The certified FPR is guaranteed with respect to randomness across control draws (and so the stated guarantee is with respect to random variables for the control sequence, $\ctrlrv_i$).  
So it's important to realize that the guaranteed FPR is not the same as the actual \newterm{in-sample FPR} that comes from a \emph{specific} drawn control pool (for which we use $\vw$ in Fig.~\ref{fig:conformal-illustration} to indicate a concrete control sequence). 
To see where this rate comes from, apply the test's rule to the actual drawn controls.
For $K\geq1$ the threshold is the $K$th-largest control score, and a sequence is flagged only if it strictly exceeds that score (Eq.~\ref{eq:conformal-threshold}).
Absent ties at the cutoff, the controls that clear it are exactly those ranked $1,\ldots,K-1$, giving an in-sample FPR of $(K-1)/N$.
(There are only $N$ drawn controls, which is why that's the in-sample denominator).
At $K=0$ there is no $K$th-largest score, so there's no threshold and nothing gets flagged. 

For example, for a control pool of $N=5000$ concrete sequences, we set a threshold from those controls and a chosen $\alpha$, and then count how many actual control sequences in the pool sit above the threshold---e.g., for $K=5$, the in-sample FPR is $4/5000 = 0.08\%$ (compare to $\alpha_N=K/(N+1)=5/5001$, which reserves a slot for the fresh candidate among $N+1$ exchangeable scores). 
This is also why the in-sample FPR is $0\%$ whenever $K=1$, even though the certified FPR there is $1/(N+1)$, not $0$. 
Another way to think of this is that, for a fixed $N$ and chosen $\alpha$, neither $\alpha_N$ nor the in-sample FPR carries any randomness: 
both are pinned by the rank $K$ before any control is drawn. 
Two runs of this experiment on entirely different control samples would report the same in-sample FPR, because $\tau_{\alpha_N}$ is defined as the $K$th-largest control score. 
What does change across draws is the population FPR $\alpha^*$.\looseness=-1

\paragraph{Population FPR.}
All three of the above quantities are different from the \newterm{population FPR} $\alpha^*$. 
To continue our Wikipedia example, this is the FPR over all Wikipedia entries.
This is the quantity we actually wish we could know;
it's $\alpha^* \coloneqq \mathrm{FPR}(\tau_{\alpha_N})$, the value at our calibrated threshold of the function $\mathrm{FPR}(\cdot)$ that maps each threshold to the fraction of the non-member population scoring above it.
That function is fixed and unknown, determined by the model, the score, and the population, with no sampling involved; it exists whether or not anyone ever runs an experiment.
$\alpha^*$ is one point on that curve. 
The curve doesn't change, but the threshold at which we read it changes across control sample draws, and so does $\alpha^*$.

To make this concrete:
you draw $N$ controls, and they get scores.
You sort those scores and take the $K$th largest one, which gives you $\tau_{\alpha_N}$.
There's no other input here, e.g., if $\alpha = 0.1\%$ and $N = 5000$, we have $\alpha_N=0.09998\%$ and $K = 5$, so $\tau_{\alpha_N}$ is the 5th-largest of the $5000$ control scores.
For instance, for Wikipedia and $50$-token suffixes, $\tau_{\alpha_N}=8.93 \times 10^{-6}$;
$\alpha^*$ is the FPR curve's height at $8.93 \times 10^{-6}$ (Fig.~\ref{fig:fpr-objects-curve}).
Now say you draw a different set of $N=5000$ controls from the same population: 
their 5th-largest probably won't be exactly $8.93 \times 10^{-6}$;
maybe it's $7.5 \times 10^{-6}$.
The curve hasn't changed, but you end up reading that curve at a different horizontal position $\tau_{\alpha_N}$, and so you get a different height $\alpha^*$.
And the reason the position moves at all on the curve is because the 5th-largest score of a random sample is itself random.
So once the controls are drawn, $\alpha^*$ is a fixed but unknown number; across control draws it's random, and it inherits that randomness entirely through $\tau_{\alpha_N}$.\looseness=-1

This isn't something the conformal test gives us.
Prop.~\ref{prop:conformal-validity} assumes only exchangeability, and it bounds the probability of falsely claiming a non-member is memorized;
it says nothing about the value of $\alpha^*$.
If we additionally assume the scores are independent and identically distributed---which our sampling design provides (\S\ref{sec:experiments:population})---as well as continuous, then the probability of that false claim is the average of $\alpha^*$ over control samples (\S\ref{app:sec:conformal:alpha-star}).
(Continuity is the assumption that doesn't come for free: 
under a top-$k$ policy many continuations may have probability exactly zero, so control scores can tie, and Lemma~\ref{lem:alpha-star-atom} recovers the result in that case.)
Under those assumptions we know its mean, which is fixed as soon as we choose the target $\alpha$ and the number of controls $N$:
\vspace{-.15cm}
\begin{align}
\label{eq:alpha-star-mean}
    \E[\alpha^*] = \alpha_N \leq \alpha.
\end{align}
This is what makes $\alpha_N$ the right quantity to subtract from the member generation rate $\hat G$ to compute the empirical estimate of calibrated memorization (Eq.~\ref{eq:emp-cmr}, $\hat M \coloneqq \hat G - \alpha_N$).
We subtract off the expected value of the population rate (which we don't see directly), so $\hat G - \alpha_N$ is unbiased for the procedure-level calibrated rate.

We also know how it's distributed~\citep{vovk2012conditional}:
the Beta law (\S\ref{app:sec:conformal:alpha-star:beta-law}). 
The Beta interval on the population FPR (e.g., $[0.032\%, 0.205\%]$) is fixed and known in advance.
It depends only on control sample size $N$ and rank $K$.
This means the interval is distribution-free, and so it's the same for both the $10$-token and $50$-token suffix experiments using the same $N$ and $K$, even though their thresholds differ by five orders of magnitude ($0.95$ versus $8.93\times10^{-6}$). 
But $\alpha^*$ varies, and falls inside it (e.g., for 95\% of control draws for the prior interval example). 
The specific \emph{quantity} is therefore conditional on the control draw we got, while its \emph{law} is a statement over control draws.
This is \emph{not} a confidence interval because the interval itself is \emph{not} computed from data and does \emph{not} vary across draws---the reverse of the usual situation, where the interval moves and the parameter stays the same. 
(In contrast, Clopper-Pearson intervals on the generation rate $\hat G$ are confidence intervals, computed from a member or non-member sample.) 

\FloatBarrier

\begin{center}
\footnotesize
\begin{tabular}{@{}llp{7.4cm}@{}}
\toprule
\textbf{object} & \textbf{example value} & \textbf{where the value comes from} \\
\midrule
target $\alpha$ & $0.1\%$ & we choose it freely \\
\midrule
\shortstack[l]{certified $\alpha_N$ \\\;} & \shortstack[l]{$5/5001 = 0.09998\%$ \\\;} & \shortstack[l]{a consequence of $N$ and $\alpha$; what Prop.~\ref{prop:conformal-validity} actually certifies; \\ e.g., $K=\lfloor 0.001\times 5001\rfloor = 5$, then $K/(N+1)$} \\
\midrule
\shortstack[l]{in-sample FPR\\\;} & \shortstack[l]{$4/5000 = 0.08\%$ \\\;} & \shortstack[l]{for $K\geq1$, the controls above $\tau$ are $K-1$ of them, \\ so the rate is $(K-1)/N$; no guarantee attached}\\
\midrule
\shortstack[l]{population $\alpha^*$ \\\;\\\;} &\shortstack[l]{unknown; $\E[\alpha^*] = \alpha_N$ \\\; \\\;} & \shortstack[l]{guarantee is about this, mean is $\alpha_N$ (e.g., $0.09998\%$);\\ Beta-distributed (law Beta$(5, 4996)$),\\ e.g., 95\% central interval $[0.032\%, 0.205\%]$} \\
\midrule
\shortstack[l]{threshold $\tau_\alpha = \tau_{\alpha_N}$ \\\;} & \shortstack[l]{$0.95$ (10-token) / \\ $8.9\times10^{-6}$ (50-token)} & \shortstack[l]{the $K$-th-largest control score\\ e.g., $K=5$} \\
\bottomrule
\end{tabular}
\end{center}

\paragraph{Understanding the calibrated threshold.}
Because the threshold is set via a rank, it's important to realize that $\tau_\alpha = \tau_{\alpha_N}$ (which is why we just use $\tau_{\alpha_N}$ in the main text). 
Consider an arbitrary $x \in (0, 1)$.
The subscript notation $\tau_x$ means ``run the recipe with $x$ as the input'': 
first $K = \lfloor x(N+1)\rfloor$, then take the $K$th-largest control score.
Supplying the target $\alpha = 0.1\%$ as $x$ with $N = 5000$ gives $K = \lfloor 5.001 \rfloor = 5$, so $\tau_\alpha = \score_{(5)}$, and $\alpha_N = K/(N+1) = 5/5001 \approx 0.09998\%$.
Now instead, supply the attained certified level $\alpha_N$ as $x$. 
This gives $K = \lfloor (5/5001)(5001) \rfloor = \lfloor 5 \rfloor = 5$, so $\tau_{\alpha_N} = \score_{(5)}$, as well.
In other words, both yield the same order statistic (literally the same number).
This is always the case, because $\floor{.}$ is the only step that discards information, and $\alpha_N$'s computation already drops that information.
Applying the floor again does nothing;
mapping a target FPR to its attained certified level is idempotent.
The same idempotence applies to the test itself: 
$\widehat p$ takes values on the $1/(N+1)$ grid, so $\widehat p(\candseq)\leq\alpha$ exactly when $\widehat p(\candseq)\leq\alpha_N$---testing at the target FPR and testing at the certified FPR are the same test.

The two thresholds are the same number, but the two subscripts don't record the same information.
The map $\alpha \mapsto \tau_\alpha$ is \emph{many-to-one}: 
every $\alpha \in [5/5001,\, 6/5001) = [0.09998\%,\, 0.11997\%)$ produces the identical threshold $\score_{(5)}$, so $\tau_\alpha$ names a threshold by an input that could have been any of the values in that continuum.
But the map $\alpha_N \mapsto \tau_{\alpha_N}$ is \emph{one-to-one} on the $1/(N+1)$ grid: 
each attained level corresponds to exactly one order statistic.
So, in summary, $\tau_\alpha$ records what was asked for, $\tau_{\alpha_N}$ records what was obtained, and $\alpha_N$ is the smallest target in the interval that yields this threshold.\looseness=-1

%% file: section/iclr/appendix/220-conformal-proof.tex
\subsection{Finite-sample validity of the conformal test}
\label{app:conformal-proof}

We prove Prop.~\ref{prop:conformal-validity}, including the treatment of ties and the finite-control resolution.
The proof follows directly from the canonical results on finite-sample validity of the conformal test from \citet{shafer2008tutorial}.\looseness=-1 

\addtocounter{theorem}{-1}
\begin{proposition}[Finite-sample validity of the conformal test]
Fix $\alpha\in(0,1)$ and the complete extraction procedure independently of the calibration controls and candidate scores.
Let $\candrv$ be a fresh candidate that satisfies the null---a non-member ($\memberind{\candrv}=0$) drawn by the same rule as the controls---and suppose its score and the $N$ control scores are exchangeable.
Then
\begin{equation*}
\Pr_0\!\left[\widehat p(\candrv)\leq\alpha\right]
\leq \alpha_N\leq\alpha,
\end{equation*}
where $\Pr_0$ is probability under the null, taken over both the random controls and the fresh candidate, so the guarantee is marginal and the left-hand side is the FPR. 
For $K\geq1$ (Eq.~\ref{eq:alpha-n}), rejection is exactly the strict threshold test in Eq.~\ref{eq:conformal-threshold}; 
for $K=0$, the deterministic test can't reject.
If the $N+1$ scores have no ties, the first inequality is an equality; ties can only make this test conservative. 
\end{proposition}

\paragraph{Setup.}
Fix the target FPR $\alpha$ and the complete extraction procedure (Def.~\ref{app:def:extraction-procedure}) before any control or candidate score is observed. 
In particular, this fixes the model $\params$, decoding policy $\dec$, prompt-selection rule $\promptrule$ (Eq.~\ref{app:eq:prompt-rule}), scoring function $\score$ (Eq.~\ref{app:eq:candidate-score}), and the form of the conformal $\verify$.
The one thing not fixed in advance is the threshold $\tau_{\alpha_N}$, which $\verify$ calibrates from the control scores.

Let
\[
Z_i=\score(\ctrlrv_i)\quad (i=1,\ldots,N),
\qquad
Z_{N+1}=\score(\candrv),
\]
where $\candrv$ is a fresh non-member candidate drawn from the matched-non-member null.
The assumption is that $(Z_1,\ldots,Z_{N+1})$ is exchangeable: 
their joint distribution is unchanged by any permutation of the $N+1$ indices. 
Exchangeability doesn't require a parametric model for their common distribution, or even independence.
Our experiments do draw controls i.i.d., which implies exchangeability; 
the Clopper--Pearson intervals in \S\ref{sec:experiments:population} and the Beta law in \S\ref{app:sec:conformal:alpha-star:beta-law} additionally require independence.

\begin{proof}
Let $U_1,\ldots,U_{N+1}$ be i.i.d.\ continuous random variables, independent of the scores $Z_1,\ldots,Z_{N+1}$.
Order the pairs $(Z_i,U_i)$ in decreasing order: 
larger scores come first, and larger $U_i$ breaks a score tie.

The pairs remain exchangeable and are almost surely distinct.
Both halves are immediate.
The $U_i$ are i.i.d.\ and independent of the scores, so permuting the $N+1$ indices permutes both coordinates together and leaves the joint distribution unchanged.
And the $U_i$ are also continuous, so $\Pr[U_i=U_j]=0$ for $i\neq j$ and no two pairs tie; 
the decreasing order is therefore a strict total order, even when scores tie.
Because the pairs are almost surely distinct, each of the $N+1$ positions in this order holds exactly one pair, so the candidate's descending rank $R$ is well defined.
And because the pairs are exchangeable, swapping index $N+1$ with any index $i$ leaves the joint distribution unchanged, so the candidate is equally likely to occupy any position.
The positions are exhaustive and mutually exclusive, so $R$ is uniform: 
\begin{equation}
\label{app:eq:r-uniform}
\Pr[R=r]=1/(N+1)
\end{equation}
for every $r\in\{1,\ldots,N+1\}$.

We write 
\[
R
=1+\sum_{i=1}^N\mathbf{1}\{Z_i>Z_{N+1}\}
  +\sum_{i=1}^N\mathbf{1}\{Z_i=Z_{N+1},\,U_i>U_{N+1}\}.
\]
That is, $R$ counts the candidate itself, plus the controls that strictly beat its score, plus the controls that tie its score but win the tie-break.

Define the randomly tie-broken rank value among the $N+1$ scores as 
\[
p^\star=R/(N+1).
\]
The deterministic conformal $p$-value (Eq.~\ref{eq:conformal-pvalue}) counts \emph{all} control ties against the candidate:
\[
\widehat p(\candrv)
=\frac{1+\sum_{i=1}^N\mathbf{1}\{Z_i\geq Z_{N+1}\}}{N+1}.
\]
So $R$ and the numerator of $\widehat p$ agree on everything except the ties: 
on the tied controls, $R$ counts only the tie-break winners; $\widehat p$ counts all of them. 
The tie-break winners are a subset of the tied controls, so the numerator of $\widehat p$ is at least $R$, i.e., 
\begin{equation}
\label{app:eq:hatp-geq-p-star}
\widehat p(\candrv) \geq p^\star.
\end{equation}

Now let $K=\lfloor\alpha(N+1)\rfloor$ and recall $\alpha_N = K/(N+1)$ (Eq.~\ref{eq:alpha-n}).
$\widehat p(\candrv)\leq\alpha$ holds exactly when $(N+1)\widehat p(\candrv)\leq\alpha(N+1)$.
The left-hand side is an integer ($1+\sum_{i=1}^N\mathbf{1}\{Z_i\geq Z_{N+1}\}$), so this holds exactly when $(N+1)\widehat p(\candrv)\leq\lfloor\alpha(N+1)\rfloor=K$. 
And $(N+1)\widehat p(\candrv)\geq R$ by Eq.~\ref{app:eq:hatp-geq-p-star}, so $(N+1)\widehat p(\candrv)\leq K$ means $R\leq K$.

Together,
\[
\{\widehat p(\candrv)\leq\alpha\}
=\{(N+1)\widehat p(\candrv)\leq K\}
\subseteq\{R\leq K\}.
\]
The inclusion and uniformity of $R$ (Eq.~\ref{app:eq:r-uniform}) therefore give
\[
\Pr_0[\widehat p(\candrv)\leq\alpha]
\leq\Pr[R\leq K]
=\frac{K}{N+1}
=\alpha_N
\leq\alpha.
\]
If the scores are almost surely distinct, the tie sum vanishes and
$\widehat p=p^\star$.
The event inclusion is then equality, so the false-positive probability is exactly $\alpha_N$.
When ties occur, counting all of them against the candidate makes $\widehat p$ weakly larger than its tie-broken counterpart (Eq.~\ref{app:eq:hatp-geq-p-star}) and can only remove rejections.
Therefore, the deterministic test is conservative.

Rejection has so far been expressed as a condition on the candidate's rank among the $N+1$ scores.
It remains to express it as a condition on a cutoff: 
a comparison of the candidate's score to the $K$th-largest control score (Eq.~\ref{eq:conformal-threshold}).

For $K\geq1$, sort the control scores in decreasing order, $Z_{(1)}\geq\cdots\geq Z_{(N)}$, and write $C=\bigl|\{i:Z_i\geq Z_{N+1}\}\bigr|$ for the number of controls scoring at least as high as the candidate.
Recall from above that $\widehat p(\candrv)\leq\alpha$ holds exactly when
\[
(N+1)\widehat p(\candrv) = 1+C \leq \lfloor\alpha(N+1)\rfloor=K. 
\]
So,
\begin{align*}
\widehat p(\candrv)\leq\alpha
&\iff 1+C\leq K\\
&\iff C\leq K-1\\
&\iff Z_{N+1}>Z_{(K)} =\tau_{\alpha_N}.
\end{align*}
The last step follows from counting positions in the sorted order:
\begin{itemize}
    \item One direction assumes $Z_{N+1}>Z_{(K)}$ and concludes $C\leq K-1$: 
    this follows because every control counted in $C$ is then strictly above $Z_{(K)}$, so it sits in one of positions $1,\ldots,K-1$.
    \item The other direction assumes $C\leq K-1$ and concludes $Z_{N+1}>Z_{(K)}$.
    This follows by the contrapositive:
    if $Z_{N+1}\leq Z_{(K)}$, then $Z_{(1)},\ldots,Z_{(K)}$ are all at least as large as the candidate score $Z_{N+1}$, so $C\geq K$ and $C\leq K-1$ fails.
    The hypothesis $Z_{N+1}\leq Z_{(K)}$ includes $Z_{N+1}=Z_{(K)}$, so the equality case is one where the test doesn't reject.
    This is why the last line is strict: 
    a candidate whose score ties the threshold fails.
\end{itemize}

Finally, if $K=0$, the numerator of $\widehat p$ is at least one, so rejection is impossible.
Equivalently, the smallest attainable conformal value is $1/(N+1)$.
The effective level $\alpha_N$ therefore changes in increments of $1/(N+1)$:
the tightest nonzero level occurs at $K=1$, when the candidate must strictly exceed the largest control score.
For example, for $N=5{,}000$ and $\alpha=0.1\%$, 
$K=\lfloor5.001\rfloor=5$ and
$\alpha_N=5/5001\approx0.09998\%$.
\end{proof}

The probability $\Pr_0[\widehat p(\candrv)\leq\alpha]$ bounded in Prop.~\ref{prop:conformal-validity} averages over both the $N$ random calibration controls and the fresh non-member candidate.
The guarantee is therefore marginal, not conditional on the controls:
conditioning on a particular realized control sample generally destroys the rank symmetry used in the proof.
The population FPR $\alpha^*$ varies across control sample draws; its distribution is given in \S\ref{app:sec:conformal:alpha-star:beta-law}.

Likewise, applying the test to many candidates gives this marginal guarantee to each candidate, but doesn't by itself bound the probability of one or more false claims or the false-discovery rate (\S\ref{app:sec:conformal:multiplicity}, \citet{bates2021testing}).
This marginal, per-candidate guarantee is what the test is designed to provide, so neither is a shortcoming of the construction.

%% file: section/iclr/appendix/240-matching.tex
\subsection{Practical limits of matching}\label{app:sec:matching}

The conformal guarantee rests on two conditions (\S\ref{sec:test}):
the controls must be exchangeable with the candidate under the matched-non-member null, and they must actually be non-members.
Exchangeability requires the controls to be selected by the same rule that would select the candidate if it were a non-member and to have the same source, format, length, and other score-relevant features.
However carefully controls are matched, some distribution shift can remain, so in practice one must make a best effort at exchangeability.
The calibrated claim is always relative to the resulting null:
the certified FPR $\alpha_N$ is a statement about the control distribution, so if a candidate comes from too different a distribution, it isn't the FPR for candidates like it.

Genuine non-members are also difficult to identify because modern training corpora are enormous and typically are of uncertain composition~\citep{lee2023explainers}.
Post-cutoff sources are useful, but they don't solve the problem automatically:
Wikipedia entries can reuse pre-cutoff text, and a post-cutoff book can quote public-domain or earlier material that was almost certainly in training.
Those sequences are actually members, rather than valid controls.

Both experiments therefore inspect the high-scoring control tail and verify reuse.
For illustrative purposes, in the main text, we do \emph{not} remove identified members from the controls in the population experiments (\S\ref{sec:experiments:population});
this is so that we can make clear that such contamination of the null raises the threshold, and so depresses memorization claims.
We detail this further in \S\ref{app:population}, where we do decontaminate for the same (and additional) experimental settings, which shows how removing the verified contaminants changes the calibrated rate reported in the main text.
In the census experiments, we do decontaminate:
we remove verified contaminants before setting each book's threshold (\S\ref{sec:experiments:census}).\looseness=-1

To sum up, distribution shift and null contamination cause issues in different directions.
Distribution shift between controls and candidates can mean a looser null than intended:
if the controls score lower than matched non-members would, the threshold calibrated from them is too low, more non-members clear it, and the actual FPR (if we'd had a better-matched null) could exceed the certified level $\alpha_N$.
In contrast, contamination only makes the test more conservative:
a contaminated control that scores high raises the threshold (\S\ref{app:population}).
Calibrated claims should therefore be qualified accordingly (\S\ref{sec:takeaways});
there should be clear reporting of how the controls were selected and matched, and their decontamination status, rather than treating a post-cutoff date as sufficient evidence of non-membership.
This matters for recent work claiming that extraction is unreliable evidence of memorization because models can verbatim complete text they weren't explicitly trained on~\citep{liu2025forged,morris2025howmuch}.
Neither \citet{liu2025forged} nor \citet{morris2025howmuch} removes near-duplicates of training data from the sequences they treat as non-members.
Without proper decontamination, the sequences treated as non-members can include near-duplicates of training data, which are members (Eq.~\ref{eq:membership-indicator}, \citet{lee2022dedup}), so a verbatim completion of such a sequence actually reflects memorization, rather than being a counterexample to it (\S\ref{sec:takeaways}).

%% file: section/iclr/appendix/250-conformal-beta-dist.tex
\subsection{The population false-positive rate of the threshold}
\label{app:sec:conformal:alpha-star}

Throughout this appendix we assume more than Prop.~\ref{prop:conformal-validity} does.
Its guarantee needs only exchangeability of the candidate's score and the $N$ control scores (\S\ref{app:conformal-proof});
the results here also require that those scores be independent and identically distributed, and---for Prop.~\ref{prop:alpha-star-law}---continuous (a requirement Lemma~\ref{lem:alpha-star-atom} relaxes).
Recall from \S\ref{app:sec:conformal:notation} that the certified FPR $\alpha_N$ is fixed by the target FPR $\alpha$ and $N$ alone.
However, for different drawn $N$-sized control samples, that $\alpha_N$ is associated with different thresholds $\tau_{\alpha_N}$.
This is because, while $\alpha_N$ is set by \emph{rank} within the controls, $\tau_{\alpha_N}$ is set by the \emph{specific scores} in the control sample (and of course, the same rank in different samples can be associated with different values).

Every threshold value is associated with a true population FPR:
the gray curve in Fig.~\ref{fig:fpr-objects-curve}.
We write $\alpha^*$ for the true FPR for a specific threshold---the value of that curve at $\tau_{\alpha_N}$.
Its randomness comes entirely from the control sample:
the curve doesn't move, but a different sample gives a different $\tau_{\alpha_N}$, and so a different point on it.
So we don't know $\alpha^*$, even though we do know $\alpha_N$.
We do know its mean.
Under the assumptions above, Prop.~\ref{prop:conformal-validity}'s guarantee---that a fresh non-member is falsely claimed memorized with probability at most $\alpha_N$, over both the control draw and the candidate---becomes a statement about $\alpha^*$: $\alpha_N$ is the average of $\alpha^*$ over control samples, i.e., $\E[\alpha^*]=\alpha_N$ (Eq.~\ref{eq:alpha-star-mean}).
But that constrains only the mean, and says nothing about how far any one draw's $\alpha^*$ sits from it.
The rest of this appendix supplies the spread: $\alpha^*$ follows a Beta distribution, and the $95\%$ band in Fig.~\ref{fig:fpr-objects} is where that distribution puts $\alpha^*$ for $95\%$ of samples.\looseness=-1

Here, we show this precisely.
We translate results from \citet{vovk2012conditional} into our notation, and make every step explicit.
That is, we state and prove the Beta distribution of $\alpha^*$ (Prop.~\ref{prop:alpha-star-law}), and show how the $95\%$ band in Fig.~\ref{fig:fpr-objects} follows from it.
We then extend that result to the score used in an extraction procedure (Eq.~\ref{app:eq:candidate-score}).
The original result assumes that ties between (arbitrary) scores have probability zero, but that's not always the case for actual extraction procedures.
For instance, under a top-$k$ decoding policy, many continuations may have probability exactly zero, in which case many controls have a tied score of $0$.
Ties don't change anything about Prop.~\ref{prop:conformal-validity}:
its bound holds whether or not scores tie, and ties can only make the test conservative (\S\ref{app:conformal-proof}).
What they change is the law of $\alpha^*$, and so the interval around it.
(\citet{vovk2012conditional}'s argument already covers tied scores, but yields only a one-sided bound on that law, with equality when the score is continuous.)
Lemma~\ref{lem:alpha-star-atom} shows that, as long as the threshold falls among the nonzero scores, the Beta law still holds exactly on a range of $\alpha^*$ values that, at the level we illustrate ($K=5$, $N=5000$), contains the whole $95\%$ band---so the intervals we report are exact.
We can also read the same law as a guarantee conditional on the drawn control sample (\S\ref{app:sec:conformal:alpha-star:conditional}), which the marginal guarantee alone doesn't provide.

\subsubsection{The Beta law}
\label{app:sec:conformal:alpha-star:beta-law}
\paragraph{Setup.}
Notation follows the proof of Prop.~\ref{prop:conformal-validity} in \S\ref{app:conformal-proof}:
$\ctrlrv_1,\ldots,\ctrlrv_N$ are the $N$ matched non-member controls and $\candrv$ is the fresh non-member candidate;
$Z_i=\score(\ctrlrv_i)$ is the $i$th control's score, and $Z_{(K)}$ is the $K$th-largest of those $N$ scores, which is the calibrated threshold $\tau_{\alpha_N}$.
But there's one deliberate exception.
That proof writes $Z_{N+1}$ for the fresh non-member candidate's score, because its argument is about the candidate's rank among the $N+1$ exchangeable scores;
the index on the score is useful for calling attention to this.
Here, we don't need a rank argument, so we instead write the fresh non-member candidate's score as $Y$.
This keeps it visually distinct from the controls, but $Z_{N+1}$ and $Y$ are the same object.

\begin{proposition}[Distribution of the calibrated threshold's population FPR]
\label{prop:alpha-star-law}
Fix $\alpha\in(0,1)$ and the complete extraction procedure independently of the calibration controls and candidate scores.
Let $Z_i=\score(\ctrlrv_i)$, $i=1,\ldots,N$, be the control scores and let $Y=\score(\candrv)$ be the score of a fresh non-member candidate.
Suppose the $N+1$ scores $Z_1,\ldots,Z_N,Y$ are mutually independent and identically distributed, sharing the cumulative distribution function (CDF) $F_0(z)=\Pr[Y\leq z]$---the fraction of the non-member population whose score is at or below $z$---and suppose $F_0$ is continuous.

Let $K=\lfloor\alpha(N+1)\rfloor\geq1$ and let $\tau_{\alpha_N}=Z_{(K)}$ be the $K$th-largest control score.
Then the population FPR of the calibrated threshold---the fraction of the whole non-member population scoring strictly above $\tau_{\alpha_N}$ (Eq.~\ref{eq:conformal-threshold})---is
\[
\alpha^* \,\coloneqq\, \mathrm{FPR}(\tau_{\alpha_N}) \,=\, 1-F_0(\tau_{\alpha_N}),
\]
and it satisfies
\[
\alpha^*\sim\mathrm{Beta}(K,\,N-K+1),
\]
so in particular
\[
\E[\alpha^*]=\frac{K}{N+1}=\alpha_N .
\]
\end{proposition}

\begin{proof}
\leavevmode\\
\textbf{Step 1: $\alpha^*$ as a function of the controls.}\\
By definition of $\mathrm{FPR}(\cdot)$ at a fixed threshold $\tau$,
\begin{equation}
\label{app:eq:fpr-id}
\mathrm{FPR}(\tau)=\Pr[Y>\tau]=1-\Pr[Y\leq \tau]=1-F_0(\tau).
\end{equation}
This holds for a threshold fixed in advance, but $\tau_{\alpha_N}=Z_{(K)}$ is random;
it depends on which controls we drew.
Conditioning on the controls resolves this, for two reasons.
First, $\tau_{\alpha_N}=Z_{(K)}$ is a function of controls $(Z_1,\ldots,Z_N)$ alone, so once we condition on the controls it's a fixed number.
Second, $Y$ is independent of $(Z_1,\ldots,Z_N)$, so conditioning on them doesn't change its distribution:
for every $z$,\looseness=-1
\[
\Pr[Y\leq z\mid Z_1,\ldots,Z_N]=\Pr[Y\leq z]=F_0(z).
\]
Conditional on the controls, we may therefore apply the identity in Eq.~\ref{app:eq:fpr-id} at the now-fixed threshold $\tau_{\alpha_N}=Z_{(K)}$:\looseness=-1
\[
\Pr[Y>\tau_{\alpha_N}\mid Z_1,\ldots,Z_N]=1-F_0(\tau_{\alpha_N})=1-F_0(Z_{(K)})=\alpha^*
\]
which is precisely how $\alpha^*$ was defined (Prop.~\ref{prop:alpha-star-law}).
So $\alpha^*=1-F_0(Z_{(K)})$ is a function of the controls alone.
But we still need to find its distribution when the controls are redrawn.

\textbf{Step 2: $\alpha^*$ is the $K$th-smallest of $N$ independent uniforms.}\\
Set $V_i=1-F_0(Z_i)$ for $i=1,\ldots,N$.
Because $F_0$ is continuous, $F_0(Z_i)$ is uniform on $[0,1]$ by the probability integral transform, and therefore so is $V_i=1-F_0(Z_i)$.
The $V_i$ are functions of the independent $Z_1,\ldots,Z_N$, so they are independent.

Let $Z_{(1)}\geq\cdots\geq Z_{(N)}$ be the controls in decreasing order, as in \S\ref{app:conformal-proof}.
Because $1-F_0$ is nonincreasing, $z\geq z'$ implies $1-F_0(z)\leq 1-F_0(z')$.
So
\[
1-F_0(Z_{(1)})\;\leq\;1-F_0(Z_{(2)})\;\leq\;\cdots\;\leq\;1-F_0(Z_{(N)})
\]
is an arrangement of $V_1,\ldots,V_N$ in nondecreasing order.

Write $V_{[1]}\leq\cdots\leq V_{[N]}$ for the $V_i$ in increasing order.
Applying $1-F_0$ reverses the control score ordering:
the largest control score $Z_{(1)}$ gives the smallest value $1-F_0(Z_{(1)})$, the second-largest gives the second-smallest, and so on.
So the $K$th-largest control score $Z_{(K)}$ gives the $K$th-smallest $V$, $V_{[K]}$.
That is,
\[
\alpha^*=1-F_0(Z_{(K)})=V_{[K]}.
\]

\textbf{Step 3: the distribution function of $\alpha^*$.}\\
Fix $u\in[0,1]$ and let $C_u=\sum_{i=1}^{N}\mathbf{1}\{V_i\leq u\}$ count the $V_i$ that are at most $u$.

First, because the $V_{[1]}\leq\cdots\leq V_{[N]}$ are sorted, the $V_i$ that are at most $u$ are exactly the first $C_u$ of them:
$V_{[1]},\ldots,V_{[C_u]}$.
So $V_{[K]}$ is among them (i.e., $V_{[K]}\leq u$) exactly when $C_u\geq K$.

Second, $C_u\sim\mathrm{Binomial}(N,u)$.
This is because each $V_i$ is uniform on $[0,1]$, so $\Pr[V_i\leq u]=u$.
And the $V_i$ are independent (Step~2), so the indicators $\mathbf{1}\{V_i\leq u\}$ are independent as well.
A sum of $N$ independent indicators, each equal to $1$ with probability $u$, is by definition $\mathrm{Binomial}(N,u)$.

Putting these together with $\alpha^*=V_{[K]}$ (Step~2),
\[
H(u) \coloneqq \Pr[\alpha^*\leq u]=\Pr[C_u\geq K]=\sum_{j=K}^{N}\binom{N}{j}u^{j}(1-u)^{N-j},
\]
the binomial probability of at least $K$ successes.

\textbf{Step 4: $H$ is the distribution function of $\mathrm{Beta}(K,N-K+1)$.}\\
For every $u\in[0,1]$, the binomial tail $\sum_{j=K}^{N}\binom{N}{j}u^{j}(1-u)^{N-j}$ equals the $\mathrm{Beta}(K,\,N-K+1)$ distribution function at $u$.

This is the standard binomial--Beta identity from Lemma~3 in \citet{vovk2012conditional}.
So $H$ is the Beta distribution function, and therefore
\[
\alpha^*\sim\mathrm{Beta}(K,\,N-K+1).
\]

\textbf{Step 5: the mean is $\alpha_N$.}\\
A $\mathrm{Beta}(a,b)$ variable has mean $a/(a+b)$.
With $a=K$ and $b=N-K+1$ the two parameters sum to $N+1$, so
\[
\E[\alpha^*]=\frac{K}{K+(N-K+1)}=\frac{K}{N+1}=\alpha_N.
\]
\end{proof}
The result in Prop.~\ref{prop:alpha-star-law} isn't new:
it's a classical fact about rank-based thresholds, stated by \citet[Proposition~2b and Lemma~3]{vovk2012conditional}.
We derive it in our own notation because the paper's variables map onto ours with a sign change that is easy to get wrong.
The one step we take from \citet{vovk2012conditional} rather than derive is the binomial--Beta identity (Step~4).
(In that paper, a conformity score is \emph{high} when an example conforms, so his score is the negative of our extraction score and his $k$th-\emph{smallest} calibration score is our $K$th-\emph{largest} control score;
his $\varepsilon$, $n$, and $k$ are our $\alpha$, $N$, and $K$, with $k=\lfloor\varepsilon(n+1)\rfloor=K$;
his ``error''---the true label falling outside the prediction set---is our false positive for memorization;
and the symbol $\alpha^*$ in his proof denotes a quantile of the score distribution, not the population FPR we use $\alpha^*$ for here.)

\subsubsection{Our score isn't always continuous}
\label{app:sec:conformal:alpha-star:atoms}
Prop.~\ref{prop:alpha-star-law} assumes a continuous $F_0$, but an extraction $\score$ doesn't always satisfy this.
For instance, under the top-$40$ decoding policy, a natural suffix whose target tokens are not all in the top $40$ has probability exactly zero.
(This is the case for the results in Fig.~\ref{fig:olmo-calibrated-rate:gen}:
$89.6\%$ of the $50$-token control scores are exactly $0$;
$41.1\%$ at $10$ tokens).
A value like this---one that the score takes with strictly positive probability---is an atom of the score distribution.
An atom at $z$ is a jump in $F_0$ at $z$, of height $\Pr[Y=z]$;
a score distribution is continuous precisely when it has no atoms anywhere.

The definition $\alpha^*=1-F_0(\tau_{\alpha_N})$ (Prop.~\ref{prop:alpha-star-law}) doesn't need a continuity assumption;
only its Beta distribution does.
So the question is, how much of that distribution survives when the score has an atom?
The answer depends on the \emph{position} of the atom, not whether an atom exists.
The calibrated threshold is one of the highest control scores, while the set of tied zero scores sits at the very bottom of the range.
As long as enough controls score above zero, the threshold lands clear of that block, and the argument behind Prop.~\ref{prop:alpha-star-law} goes through.

In terms of the proof of Prop.~\ref{prop:alpha-star-law}, continuity entered in exactly one place:
Step~2, where the probability integral transform makes each $V_i=1-F_0(Z_i)$ uniform on $[0,1]$.
The atom is where $F_0$ fails to be continuous, and all it costs is the uniformity of $V_i$.
$V_i=1-F_0(Z_i)$ is just the FPR at control $i$'s score, and FPR falls as that score rises: the higher the threshold, the fewer non-members lie above it (Step~2).
So the atom at the bottom of the score range lands at the largest FPR any threshold can produce:
every control scoring $0$ gets the same $V_i=\alpha^\dagger=\mathrm{FPR}(0)$, the fraction of the non-member population scoring above $0$.
Below $\alpha^\dagger$---a region only controls scoring above $0$ can reach---$V_i$ is still uniform:
$\Pr[V_i\leq v]=v$ for every $v<\alpha^\dagger$.
Fig.~\ref{fig:olmo-calibrated-rate:mem} annotates $\alpha^\dagger$ as the maximum $\alpha_N$: $10.4\%$ for $50$-token suffixes and $58.9\%$ for $10$-token suffixes, in each case the fraction of controls whose score is nonzero.
And $\alpha^*=V_{[K]}$ is the $K$th-\emph{smallest} of the $V_i$ (Step~2), so whenever at least $K$ controls score above the atom, $\alpha^*$ falls below $\alpha^\dagger$.

The following lemma makes this precise:

\begin{lemma}[The Beta law when atoms sit below the calibrated threshold]
\label{lem:alpha-star-atom}
We use the same setting of Prop.~\ref{prop:alpha-star-law}:
the $N+1$ scores $Z_1,\ldots,Z_N,Y$ are mutually independent and identically distributed, sharing the CDF $F_0$;
let $\alpha\in(0,1)$, $K=\lfloor\alpha(N+1)\rfloor\geq1$, $\tau_{\alpha_N}=Z_{(K)}$, and $\alpha^*=\mathrm{FPR}(\tau_{\alpha_N})=1-F_0(Z_{(K)})$, but do \emph{not} assume that $F_0$ is continuous.
Suppose instead that all atoms sit at or below some score $\tau^\dagger$, so $\Pr[Y=z]=0$ for every $z>\tau^\dagger$.
Write
\begin{align}
\label{app:eq:alpha-dagger}
\alpha^\dagger=\mathrm{FPR}(\tau^\dagger)=\Pr[Y>\tau^\dagger]
\end{align}
for the share of the non-member population scoring above $\tau^\dagger$, and suppose $0<\alpha^\dagger<1$, i.e., some non-members do score above $\tau^\dagger$, and some do not.
Then for every $v\in[0,\alpha^\dagger)$,
\[
\Pr[\alpha^*\leq v]=\sum_{j=K}^{N}\binom{N}{j}v^{j}(1-v)^{N-j},
\]
which is the distribution function of $\mathrm{Beta}(K,\,N-K+1)$ evaluated at $v$---the same distribution Prop.~\ref{prop:alpha-star-law} gives under continuity.
That accounts for every $v$ below $\alpha^\dagger$---a range $\alpha^*$ can only reach when the threshold is above $\tau^\dagger$;
the rest of the probability sits at or above $\alpha^\dagger$, and is exactly
$\Pr[\alpha^*\geq \alpha^\dagger]=\Pr[\mathrm{Binomial}(N,\alpha^\dagger)\leq K-1]$, i.e., the probability that fewer than $K$ of the $N$ controls score above $\tau^\dagger$.
\end{lemma}

\begin{proof}
\leavevmode\\
\textbf{Step 1: $\alpha^*$ is still the $K$th-smallest of the $V_i$.}\\
Step~2 of the proof of Prop.~\ref{prop:alpha-star-law} does two separate things:
it identifies $\alpha^*$ as the $K$th-smallest of the values $V_i=1-F_0(Z_i)$, and it shows that each $V_i$ is uniform.
Only the second uses continuity.
Here we check that the first still holds.
(Step~2 will recover what we need with respect to continuity:
each $V_i$ is uniform below $\alpha^\dagger$, even though it's not uniform on all of $[0,1]$.)

Set $V_i=1-F_0(Z_i)=\mathrm{FPR}(Z_i)$, and write $V_{[1]}\leq\cdots\leq V_{[N]}$ for these values in increasing order.
Because $1-F_0$ is nonincreasing, ordering the controls as $Z_{(1)}\geq\cdots\geq Z_{(N)}$ makes $1-F_0(Z_{(1)})\leq\cdots\leq1-F_0(Z_{(N)})$ a nondecreasing permutation of $V_1,\ldots,V_N$.
Its $K$th entry is therefore $V_{[K]}$, so $\alpha^*=1-F_0(Z_{(K)})=V_{[K]}$.

Nothing above assumes the scores are distinct.
Without continuity two controls can score exactly the same value (i.e., there can be an atom), which is why the ordering $Z_{(1)}\geq\cdots\geq Z_{(N)}$ may contain equalities rather than being a strict ranking.
Even so, because that ordering is written with $\geq$ and $\leq$, the $K$th entry of the sorted list is still the $K$th smallest.

\textbf{Step 2: the $V_i$ are uniform below $\alpha^\dagger$.}\\
Continuity fails only where the atoms are, and by hypothesis they all sit at or below $\tau^\dagger$.
So the conditional distribution of score $Z_i$, given that it exceeds $\tau^\dagger$, is continuous;
the probability integral transform applies to that distribution, and multiplying by $\alpha^\dagger$ turns the uniform it produces into the statement we want about $V_i$.

Let $F_0^\dagger$ be the distribution function of $Y$ conditional on $Y>\tau^\dagger$:
\begin{align}
\label{app:eq:f0-dagger}
F_0^\dagger(z)=\Pr[Y\leq z\mid Y>\tau^\dagger]=
\begin{cases}
0 & z\leq \tau^\dagger,\\[2pt]
\dfrac{F_0(z)-F_0(\tau^\dagger)}{\alpha^\dagger} & z>\tau^\dagger.
\end{cases}
\end{align}
The numerator is the population mass in $(\tau^\dagger,z]$;
dividing by $\alpha^\dagger$, the mass above $\tau^\dagger$, rescales it so that $F_0^\dagger$ reaches $1$.
Each $Z_i$ shares $F_0$ with $Y$, so each shares $F_0^\dagger$ too:
$F_0^\dagger$ is also the distribution function of $Z_i$ conditional on $Z_i>\tau^\dagger$.

$F_0^\dagger$ is continuous everywhere.
Below $\tau^\dagger$, conditioning on $Y>\tau^\dagger$ gives that whole region probability $0$, so $F_0^\dagger$ is flat at $0$ there and is continuous.
That region is where the atoms are: the conditioning discards exactly the part of the score range where $F_0$ fails to be continuous.
By hypothesis there are no atoms above $\tau^\dagger$, so $F_0$ is continuous above $\tau^\dagger$.
For $z>\tau^\dagger$, $F_0^\dagger(z)$ is $F_0(z)$ shifted and rescaled, so $F_0^\dagger$ is continuous above $\tau^\dagger$ too.
At $\tau^\dagger$ itself it is continuous because $F_0$ is right-continuous.

Therefore, the probability integral transform applies, exactly as it did to $F_0$ in Step~2 of the proof of Prop.~\ref{prop:alpha-star-law}.
The transform says that if a score has a continuous distribution function $F$, then $F$ applied to that score is uniform on $[0,1]$.
Here the score is $Z_i$ given $Z_i>\tau^\dagger$, and its distribution function is $F_0^\dagger$.
So, conditionally on $Z_i>\tau^\dagger$, $F_0^\dagger(Z_i)$ is uniform on $[0,1]$.

The uniformity we just proved is attached to $F_0^\dagger(Z_i)$, but we need it for $V_i=\mathrm{FPR}(Z_i)$;
we next write $V_i$ as a function of $F_0^\dagger(Z_i)$, so that the uniformity carries over.
Fix $z>\tau^\dagger$.
Scoring above $z$ already implies scoring above $\tau^\dagger$, so the event $\{Y>z\}$ sits inside $\{Y>\tau^\dagger\}$, and their intersection is just $\{Y>z\}$.
Therefore
\begin{align}
\label{app:eq:fpr-chain}
\mathrm{FPR}(z)=\Pr[Y>z]=\Pr[Y>\tau^\dagger]\cdot\Pr[Y>z\mid Y>\tau^\dagger]=\alpha^\dagger\bigl(1-F_0^\dagger(z)\bigr),
\end{align}
where the first equality is Eq.~\ref{app:eq:fpr-id} at $z$, and the last substitutes Eq.~\ref{app:eq:alpha-dagger} for the first factor and the complement of Eq.~\ref{app:eq:f0-dagger} for the second.

Eq.~\ref{app:eq:fpr-chain} holds for every $z>\tau^\dagger$, so we can put $z=Z_i$ whenever $Z_i>\tau^\dagger$:
\[
V_i=\mathrm{FPR}(Z_i)=\alpha^\dagger\bigl(1-F_0^\dagger(Z_i)\bigr).
\]
Since $F_0^\dagger(Z_i)$ is uniform on $[0,1]$ conditionally on $Z_i>\tau^\dagger$, so is $1-F_0^\dagger(Z_i)$.
Multiplying a uniform on $[0,1]$ by $\alpha^\dagger$ gives a uniform on $[0,\alpha^\dagger]$.
So, conditionally on $Z_i>\tau^\dagger$, $V_i$ is uniform on $[0,\alpha^\dagger]$: for every $v<\alpha^\dagger$, $\Pr[V_i\leq v\mid Z_i>\tau^\dagger]=v/\alpha^\dagger$.

A control with $Z_i\leq\tau^\dagger$ has $V_i=\mathrm{FPR}(Z_i)\geq\mathrm{FPR}(\tau^\dagger)=\alpha^\dagger$, because $\mathrm{FPR}=1-F_0$ is nonincreasing (Step~1).
So for every $v<\alpha^\dagger$, such a control has $V_i\geq\alpha^\dagger>v$, and therefore never has $V_i\leq v$: $\Pr[V_i\leq v\mid Z_i\leq\tau^\dagger]=0$.
(This conditioning is well defined because $\Pr[Z_i\leq\tau^\dagger]=1-\alpha^\dagger>0$.)
The two cases above---$Z_i>\tau^\dagger$ and $Z_i\leq\tau^\dagger$---are disjoint and exhaustive, so the law of total probability recombines them:
for every $v\in[0,\alpha^\dagger)$,
\begin{align*}
\Pr[V_i\leq v]
&=\Pr[Z_i>\tau^\dagger]\cdot\Pr[V_i\leq v\mid Z_i>\tau^\dagger]
+\Pr[Z_i\leq\tau^\dagger]\cdot\Pr[V_i\leq v\mid Z_i\leq\tau^\dagger]\\
&=\alpha^\dagger\cdot\frac{v}{\alpha^\dagger}+(1-\alpha^\dagger)\cdot 0=v .
\end{align*}
This is uniformity below $\alpha^\dagger$:
the distribution function of $V_i$ is the identity on $[0,\alpha^\dagger)$, which is the range both cases above cover.

\textbf{Step 3: below $\alpha^\dagger$, the rest of the proof of Prop.~\ref{prop:alpha-star-law} applies unchanged.}\\
Steps~3 and~4 of the proof of Prop.~\ref{prop:alpha-star-law} need exactly three things from our setting:
\begin{itemize}
    \item that the $V_i$ are independent, which holds here because they are functions of the independent $Z_1,\ldots,Z_N$
    \item that $\Pr[V_i\leq v]=v$, which Step~2 of this proof supplies for every $v<\alpha^\dagger$
    \item the identity $\alpha^*=V_{[K]}$, which is Step~1 of this proof
\end{itemize}
So for every $v\in[0,\alpha^\dagger)$, those Steps~3 and~4 of the proof of Prop.~\ref{prop:alpha-star-law} run verbatim, with $C_v=\sum_{i=1}^{N}\mathbf{1}\{V_i\leq v\}\sim\mathrm{Binomial}(N,v)$:
\[
\Pr[\alpha^*\leq v]=\Pr[V_{[K]}\leq v]=\Pr[C_v\geq K]=\sum_{j=K}^{N}\binom{N}{j}v^{j}(1-v)^{N-j}=H(v),
\]
the distribution function of $\mathrm{Beta}(K,\,N-K+1)$, where $H$ denotes this polynomial in $v$.

\textbf{Step 4: the mass at and above $\alpha^\dagger$.}\\
Step~3 gives $\Pr[\alpha^*\leq v]=H(v)$ only for $v<\alpha^\dagger$, so we take the limit of both sides as $v$ increases to $\alpha^\dagger$ from below (written $v\uparrow\alpha^\dagger$).

Take the left side first.
Increasing $v$ only adds outcomes, so $\{\alpha^*\leq v_1\}\subseteq\{\alpha^*\leq v_2\}$ whenever $v_1<v_2$.
Over all $v<\alpha^\dagger$, these events cover exactly $\{\alpha^*<\alpha^\dagger\}$: strictly less than, because no such $v$ reaches $\alpha^\dagger$ itself.
The probability of a growing family of events is the limit of their probabilities, so $\Pr[\alpha^*\leq v]$ tends to $\Pr[\alpha^*<\alpha^\dagger]$.

Now the right side.
$H$ is a polynomial, and so it's continuous.
Therefore $H(v)$ tends to $H(\alpha^\dagger)$.
Equating the two limits gives $\Pr[\alpha^*<\alpha^\dagger]=H(\alpha^\dagger)$, and therefore
\begin{align}
\label{app:eq:mass-at-above-alpha-dagger}
\Pr[\alpha^*\geq \alpha^\dagger]=1-H(\alpha^\dagger)&=\sum_{j=0}^{K-1}\binom{N}{j}(\alpha^\dagger)^{\,j}(1-\alpha^\dagger)^{N-j}\nonumber\\
&=\Pr[\mathrm{Binomial}(N,\alpha^\dagger)\leq K-1].
\end{align}
The sum $\sum_{i=1}^{N}\mathbf{1}\{Z_i>\tau^\dagger\}$ counts the controls scoring above $\tau^\dagger$.
Each control does so with probability $\alpha^\dagger$ (Eq.~\ref{app:eq:alpha-dagger}), and the $Z_i$ are independent, so that count is $\mathrm{Binomial}(N,\alpha^\dagger)$.
Therefore, Eq.~\ref{app:eq:mass-at-above-alpha-dagger} says that $\Pr[\alpha^*\geq\alpha^\dagger]$ is the probability that fewer than $K$ of the $N$ controls score above $\tau^\dagger$.
\end{proof}

\begin{remark}[An atom exactly at $\tau^\dagger$]
\label{rem:atom-at-tau-dagger}
The lemma allows atoms ``at or below'' $\tau^\dagger$.
Under the top-$40$ policy the atom is the block of zero scores, so $\tau^\dagger=0$ and the atom sits at $\tau^\dagger$ rather than strictly below it.
This is exactly the boundary case the lemma's ``at or below'' allows.

It's therefore worth highlighting that nothing in the proof depends on the atom being strictly below $\tau^\dagger$.
Four places could have been sensitive to this case, and each resolves the same way: through a strict inequality.
\begin{enumerate}
    \item $\alpha^\dagger$ is what $F_0^\dagger$ is normalized by, so it has to be exactly the probability of the event we condition on.
    The lemma defines $\alpha^\dagger=\Pr[Y>\tau^\dagger]$ with the same strict inequality as that conditioning, so a control scoring exactly $\tau^\dagger$ counts in neither.

    \item Step~2 needs $F_0^\dagger$ to be continuous at $\tau^\dagger$, and $F_0^\dagger$ is built from $F_0$, which is \emph{not}: $F_0$ rises at $\tau^\dagger$ by the atom's mass $\Pr[Y=\tau^\dagger]$.
    The strict inequality in the conditioning $Y>\tau^\dagger$ is what keeps that mass out.
    In the formula it appears as the subtraction of $F_0(\tau^\dagger)$: a distribution function is $\Pr[Y\leq z]$, so $F_0(\tau^\dagger)$ already counts the atom's mass, and subtracting it removes it.

    \item Step~2 needs controls at or below $\tau^\dagger$ to contribute nothing to $\Pr[V_i\leq v]$ when $v<\alpha^\dagger$.
    A control landing exactly on the atom has $V_i=\mathrm{FPR}(\tau^\dagger)=\alpha^\dagger$, and the lemma's range $[0,\alpha^\dagger)$ stops strictly short of that value, so $V_i\leq v$ never holds for such a control.
    Its contribution is therefore zero: $\Pr[V_i\leq v\mid Z_i\leq\tau^\dagger]=0$.

    \item Step~4 needs the count $\sum_{i=1}^{N}\mathbf{1}\{Z_i>\tau^\dagger\}$ to be $\mathrm{Binomial}(N,\alpha^\dagger)$, so a single indicator must succeed with probability exactly $\alpha^\dagger$.
    It does: the $Z_i$ share the distribution of $Y$, and the indicator tests the same strict inequality that defines $\alpha^\dagger$ (Eq.~\ref{app:eq:alpha-dagger}).
    The strict inequality matters here, since the atom puts positive probability on $Z_i=\tau^\dagger$.
    Had the indicator tested $Z_i\geq\tau^\dagger$, it would have succeeded with probability $\alpha^\dagger$ plus the atom's mass.
\end{enumerate}
The atom does affect $\alpha^*$ in one way: it gives $\alpha^*$ positive probability of equaling $\alpha^\dagger$ exactly.
But the Beta law is claimed only on $[0,\alpha^\dagger)$, and everything from $\alpha^\dagger$ upward is reported as the single number $\Pr[\alpha^*\geq\alpha^\dagger]$, which includes $\Pr[\alpha^*=\alpha^\dagger]$.
That is why the lemma splits its two claims at a strict inequality.
\end{remark}

\subsubsection{Conditional guarantees}
\label{app:sec:conformal:alpha-star:conditional}
Recall that Prop.~\ref{prop:conformal-validity}'s guarantee is \emph{marginal}: 
it averages over control draws.
Under our additional assumption that the control scores and the candidate's score are independent and identically distributed (stated at the start of this appendix), this means that the average of the population FPR $\alpha^*$ over control draws is the certified level $\alpha_N$, i.e., $\E[\alpha^*]=\alpha_N$.
Recall that $\alpha^*$ changes with the control draw: it's the population FPR of the calibrated threshold, and that threshold is set from the specific control sample drawn, for the chosen target $\alpha$.
An average of $0.1\%$ is compatible with $\alpha^*$ being $0.05\%$ for half of the control samples and $0.15\%$ for the other half.
So, for the specific control sample we draw for our experiments, the marginal guarantee doesn't say how far the specific (unknown) $\alpha^*$ may sit from $\alpha_N$.

A \emph{conditional} guarantee is a statement about that specific control sample's $\alpha^*$~\citep{vovk2012conditional}.
Since $\alpha^*$ can't be observed, such a guarantee takes the form of a bound that holds for most control samples: 
$\alpha^*\leq v$ for all but a fraction $\delta$ of possible control draws.
This statement requires the distribution of $\alpha^*$ across control draws, not just its mean, which is what Prop.~\ref{prop:alpha-star-law} and Lemma~\ref{lem:alpha-star-atom} supply (and what exchangeability alone, as in Prop.~\ref{prop:conformal-validity}, can't).

\begin{corollary}[A guarantee conditional on the drawn control sample]
\label{cor:alpha-star-conditional}
Under the assumptions of Prop.~\ref{prop:alpha-star-law} for every $v\in(0,1)$, and under those of Lemma~\ref{lem:alpha-star-atom} for every $v\in[0,\alpha^\dagger)$,
\begin{equation}
\label{app:eq:alpha-star-cdf}
\Pr[\alpha^*\leq v]=\sum_{j=K}^{N}\binom{N}{j}v^{j}(1-v)^{N-j}.
\end{equation}
That is, over the control draws, the calibrated threshold set from the drawn control sample has population FPR $\alpha^*$ at most $v$, with exactly the above probability.
Equivalently, for any $\delta\in(0,1)$, let $v_{1-\delta}$ be the $(1-\delta)$-quantile of $\mathrm{Beta}(K,\,N-K+1)$ (with $v_{1-\delta}<\alpha^\dagger$ under the lemma's assumptions).
Then, with probability at least $1-\delta$ over control draws,
\begin{equation}
\label{app:eq:alpha-star-quantile}
\alpha^*\leq v_{1-\delta}.
\end{equation}
\end{corollary}
\begin{proof}
Both statements are the distribution function of $\alpha^*$, evaluated at $v$ and at $v_{1-\delta}$ respectively.
Prop.~\ref{prop:alpha-star-law} gives that distribution function for every $v$ when $F_0$ is continuous; Lemma~\ref{lem:alpha-star-atom} gives it for every $v<\alpha^\dagger$ when $F_0$ has atoms at or below $\tau^\dagger$.
\end{proof}

To make this concrete, we can walk through the corollary with some of the specific numbers in our Wikipedia experiments. 
With $N=5000$ controls and target FPR $\alpha=0.1\%$, $K=5$ and the certified FPR is $\alpha_N=5/5001\approx0.1\%$.
The certified level $\alpha_N$ is the mean of $\alpha^*$, but it is not a bound on it: 
evaluating Eq.~\ref{app:eq:alpha-star-cdf} at $v=\alpha_N$, $\alpha^*$ exceeds $\alpha_N$ for $44\%$ of control samples.
So the statement ``the specific calibrated threshold's FPR is at most $\alpha_N$'' holds for only $56\%$ of control samples; 
it's not a statement one can make with high confidence.
Using Eq.~\ref{app:eq:alpha-star-quantile}, what one can say with $90\%$ confidence is that $\alpha^*\leq v_{0.9}=0.16\%$, and with $95\%$ that $\alpha^*\leq v_{0.95}=0.18\%$.
(The two-sided $95\%$ interval $[0.032\%,0.205\%]$ reported in \S\ref{app:sec:conformal:notation} comes from the same distribution: its endpoints are $v_{0.025}$ and $v_{0.975}$.)

Next, define the relative gap between the $(1-\delta)$-quantile and the certified level,
\begin{equation}
\label{app:eq:alpha-star-gap}
\rho_{1-\delta}\coloneqq\frac{v_{1-\delta}-\alpha_N}{\alpha_N},
\end{equation}
which depends only on $N$, $K$, and $\delta$.
For a fixed $\delta$, $\rho_{1-\delta}$ is proportional to $1/\sqrt{K}$.\footnote{The standard deviation of $\mathrm{Beta}(K,\,N-K+1)$ is approximately $\sqrt{K}/(N+1)$ while its mean is $K/(N+1)$.
    In the example above ($K=5$, $\delta=0.1$), $\rho_{0.9}=0.60$.
}
To continue the example, consider also the most conservative operating point, where the calibrated threshold is the top control score ($K=1$, $\alpha_N=1/5001\approx0.02\%$).
There, Eq.~\ref{app:eq:alpha-star-cdf} gives $\Pr[\alpha^*\leq0.02\%]=63\%$: 
the calibrated threshold's population FPR is at most its own certified level for only $63\%$ of control samples.
It's at most $0.05\%$ for $92\%$ of samples, and at most $0.1\%$ for $99.3\%$.
A smaller $K$ therefore lowers both $\alpha_N$ and $v_{1-\delta}$, but it raises the relative gap between them: at $K=1$, $\rho_{0.9}=1.30$.
Because Lemma~\ref{lem:alpha-star-atom} makes the corollary exact for every $v<\alpha^\dagger$, this reading applies unchanged to the top-$40$ results of the main text, where $\alpha^\dagger\approx10.4\%$.

In our figures, we focus on the $K=5$ operating point.
The calibrated rate we report subtracts the certified level, $\hat M=\hat G-\alpha_N$ (Eq.~\ref{eq:emp-cmr}), rather than the unknown $\alpha^*$ (\S\ref{app:sec:population:intervals}).
We can't subtract $\alpha^*$ itself, since it isn't observable, and subtracting $\alpha_N$ keeps $\hat M$ unbiased (\S\ref{app:sec:conformal:notation}).
A conservative alternative would be to subtract a high quantile of its distribution instead, understating the rate by design.
Doing so would change little here. 
Replacing $\alpha_N\approx0.1\%$ with $v_{0.95}=0.18\%$ lowers $\hat M$ by $0.08$ percentage points, roughly a tenth of the width of the interval already plotted around the member curve at that point (Fig.~\ref{fig:olmo-calibrated-rate:mem}).

It's also worth noting that Prop.~\ref{prop:alpha-star-law} assumes every control is a non-member, which our running example shows isn't always the case in practice.
In that experiment, a detailed decontamination process revealed that twelve of the $5{,}000$ are actually members (\S\ref{app:sec:population:contamination}).
Their presence raises the calibrated threshold;
removing them raises the calibrated rate from $0.44\%$ to $3.06\%$ at the tightest certified FPR.
Since $\alpha^*$ is the fraction of the non-member population above the calibrated threshold, a higher threshold means a smaller $\alpha^*$ than a non-contaminated control pool would have produced.
Bounds like $\alpha^*\leq v_{0.95}=0.18\%$ are therefore conservative here, holding for more than the stated $95\%$ of control draws.

\subsubsection{The point of all this in practice}
\label{app:sec:conformal:alpha-star:practice}
We allow an extraction procedure to choose its decoding policy $\dec$, and some policies put an atom in the score distribution.
Under top-$40$ decoding the score is exactly $0$ for every suffix the policy cannot generate, and strictly positive otherwise, so $\tau^\dagger=0$.
The atom has one consequence: $\alpha^\dagger=\mathrm{FPR}(0)$, the share of the non-member population the policy can generate at all, is a ceiling on the attainable certified FPR, since no threshold puts more than $\alpha^\dagger$ of that population above it.
This is the maximum attainable $\alpha_N$ annotated in Fig.~\ref{fig:olmo-calibrated-rate:mem} for OLMo~2 32B on Wikipedia: $10.4\%$ for $50$-token suffixes and $58.9\%$ for $10$-token suffixes.

Importantly, though, this regime (the highest attainable FPR) is actually the \emph{least} interesting for extraction:
we care about the \emph{low} FPR regime, which is far away from all the zero-probability controls (the atom).
For our relatively small sample size of $N=5000$, we calibrate at $\alpha_N\approx0.1\%$.
(For larger sample sizes, we'd ideally calibrate to an even lower $\alpha_N$.)
And the lemma says the Beta law is exact for every $\alpha^*$ below $\alpha^\dagger\approx10.4\%$, which is everywhere our reported intervals live (\S\ref{app:sec:conformal:notation}).
So those intervals are exact, not approximations that ignore the atom.

In practice, we don't have to worry about atoms at nonzero scores.
For this to happen, continuations for controls would have to share an identical product of token probabilities, which is extremely unlikely in practice.
(Nevertheless, we can check this; in our experiments, the nonzero control scores are all distinct.)

The one case the lemma rules out is $\alpha^\dagger=0$:
the extraction procedure can't generate any non-member at all.
There the Beta law has nothing to say, since the interval $[0,\alpha^\dagger)$ it's claimed on is empty.
The test itself still works---every control scores $0$, so the threshold is $0$, and any candidate the procedure can generate is claimed memorized, with a realized false-positive rate of $0$.
But $0$ is now the only attainable FPR:
every $\alpha_N$ returns the same threshold and the same rejection region, so the sweep of operating points our figures plot collapses to a point (and so with it the Beta band).

Greedy decoding is the same story, but with two attainable FPRs instead of one:
scores take only the values $0$ and $1$, so every threshold sits on an atom and the only attainable FPRs are $\Pr[Y=1]$ and $0$ (\S\ref{sec:experiments:population}).
Prop.~\ref{prop:conformal-validity} still holds there---the marginal guarantee never needed continuity---but $\alpha^*$ isn't Beta-distributed and the band shouldn't be drawn.
Greedy plots carry error bars (Fig.~\ref{fig:olmo-greedy-bar-10-vs-50}), but those are ordinary Clopper--Pearson confidence intervals on the measured generation rates, computed from the sample---unlike the Beta band on $\alpha^*$, which is fixed before any control is drawn (\S\ref{app:sec:conformal:notation}).

%% file: section/iclr/appendix/260-conformal-multiplicity.tex
\subsection{Multiple testing across candidates}
\label{app:sec:conformal:multiplicity}

The conformal test makes a claim about one candidate at a time: 
for each candidate it either claims memorization or makes no claim at all.
Prop.~\ref{prop:conformal-validity} bounds the probability of claiming memorization for a non-member, taken over both the random controls and the fresh candidate (\S\ref{app:conformal-proof}).
One control sample can therefore serve to make calibrated claims for many candidates;
nothing in the guarantee requires those candidates to be independent of one another, so each candidate's claim stands on its own.\looseness=-1

Reusing that sample across candidates does, however, make the resulting claims dependent on one another.
Every candidate is tested against the same threshold $\tau_{\alpha_N}$, the $K$th-largest score in the one control sample we drew (Eq.~\ref{eq:conformal-threshold}).
Its value depends on which controls we happened to draw.
So under a different draw, the same candidates would be tested against a different threshold; 
raising it can only remove claims, and lowering it can only add them.
Given a continuous $\score$, take two non-member candidates drawn independently of each other and of the controls, and see whether the test claims memorization for each of them; 
the two outcomes are positively correlated, at exactly $1/(N+2)$~\citep{bates2021testing}.
The same $1/(N+2)$ correlation holds for the conformal $p$-values (Eq.~\ref{eq:conformal-pvalue}) and for any finite-valued function of them, so the dependence can't be transformed away.\footnote{Our $\score$ isn't always continuous.
    For instance, under top-$k$ decoding many sequences share a score of exactly $0$ (\S\ref{app:sec:conformal:alpha-star:atoms}).
    \citet{bates2021testing} prove the exact value $1/(N+2)$ for a continuous $\score$ and extend the positive dependence to scores with atoms for a randomized $p$-value (their Appendix~A.3; see the footnote in the BH paragraph below), but don't extend this exact correlation to that setting. 
    We don't pursue this here, either.
}
Each individual claim still carries the guarantee of Prop.~\ref{prop:conformal-validity}.
But it doesn't bound the error rates that arise when many candidates are tested together---e.g., all $5000$ Wikipedia candidates at once.

\paragraph{Marginal error rates across candidates.}
Suppose the conformal test is applied to candidates $\candrv_1,\ldots,\candrv_J$, where the candidates are random draws of a specified length from data of interest, e.g., all Wikipedia entries (the random counterparts of $\candseq_1,\ldots,\candseq_J$ in Fig.~\ref{fig:conformal-illustration}).
Let $\sS_0=\{j:\memberind{\candrv_j}=0\}$ index the non-members among them (Eq.~\ref{eq:membership-indicator}).
As in \citet{bates2021testing}, we treat $\sS_0$ as fixed rather than random: 
what varies across draws is the candidates' scores, not which of them are non-members.
Each $\candrv_j$ with $j\in\sS_0$ is then a fresh non-member drawn from the matched control population of interest (exactly the $\candrv$ of Prop.~\ref{prop:conformal-validity}).
Let $\sM=\{j:\score(\candrv_j)>\tau_{\alpha_N}\}$ index the candidates claimed to be memorized (Eq.~\ref{eq:conformal-threshold}).
A false positive is a non-member claimed to be memorized, so $|\sM\cap\sS_0|$ counts the false positives among the $J$ candidates.

Generally speaking, there are two error rates for false positives across the $J$ candidates that are of interest: 
the \newterm{familywise error rate (FWER)} and the \newterm{false discovery rate (FDR)}~\citep{benjamini1995controlling}.
The familywise error rate is the probability of at least one false positive among the $J$ candidates.
The false discovery rate is the expected share of false positives among all the candidates claimed to be memorized.
More formally,
\begin{equation}
\label{app:eq:fwer-fdr}
\mathrm{FWER}\,\coloneqq\,\Pr\bigl[\,|\sM\cap\sS_0|\geq1\,\bigr],
\qquad
\mathrm{FDR}\,\coloneqq\,\E\!\left[\frac{|\sM\cap\sS_0|}{\max\{1,|\sM|\}}\right].
\end{equation}
Both of these rates, like the FPR in Prop.~\ref{prop:conformal-validity}, are marginal:
they average over the draw of the $N$ controls and the candidates.
Therefore, they are statements about the conformal test, rather than the specific control sample and set of candidates that an experiment happens to use. 

A difference from the FPR is that Prop.~\ref{prop:conformal-validity} concerns a claim for a single candidate (whether or not \emph{that} candidate is memorized), whereas these two rates concern all $J$ candidates' claims together.
All three rates depend on the threshold $\tau_{\alpha_N}$, which the controls set (Eq.~\ref{eq:conformal-threshold}).
But the FPR is a probability about a single hypothetical candidate: 
a fresh non-member, drawn the same way the controls are.
It doesn't depend on which candidates we test, or on how many.
FWER and FDR, by contrast, are defined over the $J$ candidates we actually test, through $\sS_0$ and $\sM$ (Eq.~\ref{app:eq:fwer-fdr}).
So if none of those candidates is a non-member, $|\sM\cap\sS_0|=0$ regardless of $\sM$, and both rates are $0$ by construction.
That is the case in our experiments, where every candidate is a member (\S\ref{sec:experiments:population}, \S\ref{app:population}).\footnote{This is a different point from the one made in the main text, that the certified FPR (based on the control generation rate) gives the share of the member generation rate that may reflect false positives (\S\ref{sec:experiments:population}).
    In that setting, all candidates are members, so those specific candidates can't literally be false positives with respect to membership.
    But this is beside the point if we step through the inferential claim (Box~\ref{box:extraction-claim:main}). 
    We use $\score$ as the signal: 
    a generation probability that clears the calibrated threshold is too high to be explained by general predictability, which is what supports the conclusion that the candidate is a member that the model memorized.
    In this sense, the conformal test is a test for whether a candidate \emph{behaves like a member}.
    (And indeed, member candidates that don't surpass the calibrated threshold are clearly still members;
    they just don't have sufficiently high signal to be deemed memorized.
    This is partly why ``membership inference''~\citep{shokri2017membership} can be a bit of a confusing term;
    it's really about whether a sequence ``behaves enough like a member'' to be reflective of being memorized by the model.) 
    A certified FPR $\alpha_N$ then says how often the same calibrated threshold would incorrectly deem a non-member to be memorized;
    at that level, some members that clear the threshold are indistinguishable from non-members with respect to their $\score$ (up to that certified FPR), and for them we may be over-claiming memorization. 
    The calibrated rate corrects for this (Eq.~\ref{eq:emp-cmr}).
    FWER and FDR, by contrast, aren't being interpreted with respect to this over-claiming: 
    unlike what the FPR measures, they count non-members in the candidate set that clear the threshold (and in our candidate set, there are none).
}       

Saying something about either of these rates means changing the rule in the conformal test that turns $p$-values into claims.
Recall that our conformal test decides whether to make a claim for each candidate on its own: 
for a specific candidate $\candseq$, the test claims memorization when that candidate's $p$-value is at most $\alpha_N$ (Eq.~\ref{eq:conformal-pvalue}), with the cutoff fixed by the controls alone.
A \newterm{multiple-testing procedure} instead uses the whole family of $J$ $p$-values to decide which candidates to make a claim about.
Both rates depend on the non-member candidates being tested (those indexed by $\sS_0$), which (when testing candidates with unknown membership) we don't know.
Not knowing $\sS_0$ also means that a false positive can't be identified when it occurs.
So we ask for a level the rate never exceeds, regardless of which of the $J$ candidates are non-members;
a procedure that achieves this bounds the rate at that level.
We next discuss procedures that do this for FWER and FDR, respectively.

\paragraph{Bonferroni and FWER.}
Even though Prop.~\ref{prop:conformal-validity} doesn't \emph{determine} the FWER, it does \emph{say something} about the FWER, coming from the union bound.
The union bound says that the probability that any of several events occurs is at most the sum of those events' probabilities.
Here, the events are ``candidate $j$ is a false positive,'' one for each $j\in\sS_0$;
by Prop.~\ref{prop:conformal-validity} each of those events has probability at most $\alpha_N$.
Therefore, 
\begin{align}
\label{app:eq:fwer-union}
\mathrm{FWER}&=\Pr\bigl[\,|\sM\cap\sS_0|\geq1\,\bigr]\nonumber\\
& \leq\;\sum_{j\in\sS_0}\Pr\bigl[\,j\in\sM\,\bigr]
\;\leq\;J_0\,\alpha_N
\;\leq\;J\,\alpha_N,
\qquad \text{where } J_0=|\sS_0|.
\end{align}
The tighter of the last two bounds, $J_0\alpha_N$, is valid, but its value is unknown because $\sS_0$ (and so $J_0$) is unknown.
The bound we can actually compute is $J\alpha_N$, since we do know $J$ (we chose it). 
To guarantee $\mathrm{FWER}\leq\delta$ for a target $\delta$ chosen in advance (a level for the FWER, playing the analogous role that $\alpha$ does for the FPR), one sets $\alpha_N\leq\delta/J$. 
This is the Bonferroni correction.\footnote{The same bound covers the FDR, since $\mathrm{FDR}\leq\mathrm{FWER}$ always:
    the ratio in Eq.~\ref{app:eq:fwer-fdr} is $0$ when there is no false positive and at most $1$ when there is, so its expectation is at most $\Pr[\,|\sM\cap\sS_0|\geq1\,]$.}

For our memorization setting, the Bonferroni correction isn't practically useful.
The reason is that $\alpha_N$ can't be made arbitrarily small:
the smallest FPR that $N$ controls can certify is $1/(N+1)$ (Eq.~\ref{eq:alpha-n}).
So Bonferroni's requirement of $\alpha_N\leq\delta/J$ can only be met when $\delta/J$ is at least $1/(N+1)$, i.e., $1/(N+1)\leq\delta/J$.
Put one way, $J \leq \delta(N+1)$: 
with $N$ controls, Bonferroni can hold the FWER at $\delta$ over at most $\delta(N+1)$ candidates.
Put another, $N\geq J/\delta-1$: 
holding the FWER at $\delta$ over $J$ candidates takes at least $J/\delta-1$ controls.
With our $N=5000$ controls, even a loose $\delta=5\%$ means we could hold the FWER at $5\%$ over at most $250$ candidates ($0.05\times5001$);
at $\delta=0.1\%$, we could hold the FWER over at most $5$ candidates.
In our experiments, holding the FWER at $0.1\%$ over all $5{,}000$ of our Wikipedia candidates would take about five million controls ($5000/0.001-1$).

\paragraph{Benjamini--Hochberg and FDR.}
For the FDR, a bound at a useful level is available without a correction of this kind, as shown by \citet{bates2021testing}.\footnote{\citet{bates2021testing} study \newterm{outlier detection}, so their vocabulary differs from ours:
    their \newterm{calibration set} is our control sample, their \newterm{inliers} are our non-members, and their \newterm{outliers} are our members.
    Their marginal conformal $p$-value (their Eq.~3) is Eq.~\ref{eq:conformal-pvalue} up to the orientation of the score:
    they count calibration scores at or \emph{below} the test point's;
    we count control scores at or \emph{above} the candidate's.
}
The Benjamini--Hochberg (BH) procedure~\citep{benjamini1995controlling} sorts the $J$ conformal $p$-values as $p_{(1)}\leq\cdots\leq p_{(J)}$, finds the largest $r$ with $p_{(r)}\leq r\delta/J$ for a target level $\delta$, and (in the context of our conformal test) would claim memorization for the $r$ candidates with the smallest $p$-values.
It bounds the FDR at $\delta$ when the $p$-values are independent, and also when they're positively dependent in a specific sense: 
\newterm{positive regression dependence on a subset (PRDS)}~\citep{benjamini2001control}.
\citet{bates2021testing} prove that marginal conformal $p$-values computed from one shared calibration set are PRDS, provided the score is continuous\footnote{For a score with atoms, as ours has under top-$k$ decoding (\S\ref{app:sec:conformal:alpha-star:atoms}), they prove the same for a randomized $p$-value that breaks ties uniformly at random (their Proposition~4, Appendix~A.3).
    This randomized $p$-value differs from Eq.~\ref{eq:conformal-pvalue}, where ties count against the candidate.
} and (in our context) the non-member candidates (their inliers) are independent of one another and of the controls (their Theorem~2).
So the BH procedure can be run on our $p$-values unchanged, and bounds the FDR at $\pi_0\delta\leq\delta$, where $\pi_0=J_0/J$ is the fraction of candidates that are non-members (their Corollary~1).

We don't study either bound in this work.
As noted above, Bonferroni isn't practically useful here: holding the FWER over all of our candidates, even at a relatively modest level, would take orders of magnitude more control sequences.
BH is something we could study, but it would only apply to extraction procedures with continuous scores, and the results we highlight in the main paper---greedy and top-$k$ decoding---don't use a continuous $\score$ (both have atoms, see \S\ref{app:sec:conformal:alpha-star:atoms}).
Regardless, our main object of study is per-candidate claims about memorization, which is exactly what Prop.~\ref{prop:conformal-validity} is about (and from which we can derive an empirical estimate of calibrated memorization, $\hat M$; see Eq.~\ref{eq:emp-cmr}).\looseness=-1

In future work, it would be perfectly reasonable for others to use the conformal test to make candidate-set-level claims about error rates---for instance, studying a model where it's unknown whether the candidates were in its training data, and where the goal is a list of sequences claimed to be memorized.
For BH, one could use the same $p$-values (Eq.~\ref{eq:conformal-pvalue}), and would need a continuous $\score$ and the i.i.d.\ assumption we already make in practice in our experiments (see \S\ref{sec:experiments:population}).

%% file: section/iclr/appendix/300-population.tex
\section{Population experiments}\label{app:population}

This appendix gives additional details on the population experiments of \S\ref{sec:experiments:population}, which run the conformal test (\S\ref{sec:test}, \S\ref{app:sec:conformal}).
We first state exactly which confidence interval each plot uses, and why they differ across plots (\S\ref{app:sec:population:intervals}).
We then describe contamination of the Wikipedia control sample (\S\ref{app:sec:population:contamination}):
some post-cutoff controls contain pre-cutoff (member) text.
The main paper deliberately leaves these contaminants in the control pool, to make the point that contamination pushes the threshold up and so only makes the test more conservative (\S\ref{app:sec:matching}).
Here, we show what the results would look like with decontamination, which shows how much the calibrated memorization rate rises.

We then extend the experiments in the main paper in three ways.
First, suffix length: the main paper uses $10$- and $50$-token suffixes; here we sweep every length from $1$ to $50$ tokens (OLMo~2 32B; \S\ref{app:sec:population:length}).
Second, decoding policy: the main paper's greedy and top-$k$ ($k=40$) policies both use scores with atoms (greedy is binary, and under top-$k$ a sequence can have generation probability exactly $0$), so we also run probabilistic extraction under the base distribution at temperature $T=1$, where the score is continuous (\S\ref{app:sec:population:policy}).
Third, model size: under that policy we add OLMo~2 7B and 13B alongside 32B, with a paired comparison of calibrated memorization across sizes (also \S\ref{app:sec:population:policy}).

\subsection{Plotted intervals}\label{app:sec:population:intervals}

In the plots in our population experiments (here and in the main paper), all shown intervals are $95\%$ and pointwise:
each is a statement about one plotted point, not about a curve as a whole.
Where we sweep a threshold (e.g., Fig.~\ref{fig:olmo-calibrated-rate}; the sweep figures below), the shaded band is the pointwise interval at each threshold, drawn continuously.
It isn't a simultaneous band over the whole sweep.

There are four kinds of intervals, and which one is appropriate depends on what is random in the specific plot.

\begin{itemize}[leftmargin=.5cm]
    \item \textbf{Clopper--Pearson on a single rate.}
    For greedy decoding (Fig.~\ref{fig:olmo-greedy-bar-10-vs-50}, Fig.~\ref{app:fig:greedy-length-raw}), each sampled sequence is either reproduced verbatim or not, so the member and control counts are binomial and both bars carry exact Clopper--Pearson intervals.
    For probabilistic extraction, the member rate above a threshold $\tau$ is also a binomial proportion, so the member curve in Fig.~\ref{fig:olmo-calibrated-rate:gen} and in the generation panels of Fig.~\ref{fig:olmo-calibrated-rate-t1} carries a pointwise Clopper--Pearson band;
    the certified level $\alpha_N$ is a designed constant and carries no interval (though \S\ref{app:sec:conformal:alpha-star:beta-law} shows that $\alpha_N$ is the mean of the Beta distribution of the population FPR $\alpha^*$, which gives an interval for $\alpha^*$).

    \item \textbf{Clopper--Pearson shifted by a constant.}
    The calibrated rate $\hat M=\hat G-\alpha_N$ (Eq.~\ref{eq:emp-cmr}) subtracts a constant from the generation rate $\hat G$, so its interval is the Clopper--Pearson interval on $\hat G$ translated down by $\alpha_N$, with the same width.
    This is the band in Fig.~\ref{fig:olmo-calibrated-rate:mem}, Fig.~\ref{app:fig:population-contamination}, Fig.~\ref{app:fig:prob-generation-length}, Fig.~\ref{app:fig:prob-cmr-length}, the absolute rates in Fig.~\ref{app:fig:population-scale:absolute}, and the calibrated-rate panels of Fig.~\ref{fig:olmo-calibrated-rate-t1}.
    The interval is exact because the threshold $\tau_{\alpha_N}$ is set by the controls, which are independent of the members:
    conditional on that threshold, the number of members scoring above it is binomial.
    So the interval covers one source of randomness, the member sample, holding the threshold fixed at the value the drawn controls set.
    It doesn't cover the other source, the control draw itself: a different control sample would set a different threshold, with a different population FPR $\alpha^*$.
    We keep that second type of uncertainty out of the band on purpose: 
    the plotted quantity is defined with the certified constant $\alpha_N$, not the unknown $\alpha^*$;
    an interval that also carried the control-draw uncertainty would be an interval for $\hat G-\alpha^*$, which is a different quantity.
    \S\ref{app:sec:conformal:alpha-star:conditional} discusses that second source of uncertainty.

    \item \textbf{Newcombe's independent-proportions interval.}
    In two figures, the quantity subtracted from the member rate depends on the control draw, so it carries sampling variance of its own.
    For greedy decoding, the score is binary, so there's no target FPR to choose: the false-positive floor is set by how many controls the model reproduces, and a different control draw would give a different floor (Fig.~\ref{app:fig:greedy-length-cmr}).
    For the budget plots at $n=10^5$ (Fig.~\ref{app:fig:olmo-budget-cmr}), the subtracted term is the measured non-member rate at that budget, again an empirical proportion over the control sample.
    For a difference of two estimated proportions from independent samples, Newcombe's independent-proportions interval is a useful choice: 
    a Wilson score interval for each rate, combined into one interval for the difference~\citep{newcombe1998interval}.

    \item \textbf{Newcombe's paired-proportions interval.}
    For the same candidates scored by two models (Fig.~\ref{app:fig:population-scale:diffs}), the two rates aren't independent, because every candidate contributes to both.
    We use Newcombe's paired-proportion interval, which builds a Wilson interval for each model's rate and combines them for the difference with a correction for the correlation between the two rates~\citep{newcombe1998improved}.
    That correlation is estimated from how often the two models agree on whether a given candidate is claimed memorized.
    \S\ref{app:sec:population:policy} explains why this is useful in further detail.
\end{itemize}

\subsection{Score distributions and contamination}\label{app:sec:population:contamination}

Probabilistic extraction scores each suffix by a continuous quantity, so the conformal threshold can be swept across operating points.
At $10$ tokens, many non-member continuations are highly predictable.
Holding the target FPR to $0.1\%$ therefore forces the threshold close to one:
the test deems a candidate memorized only when the model generates it with near certainty.
At $50$ tokens, non-members are almost never generated verbatim, so the threshold giving the same target FPR is orders of magnitude smaller.

\begin{figure}[h!]
  \centering
  \includegraphics[width=0.65\linewidth]{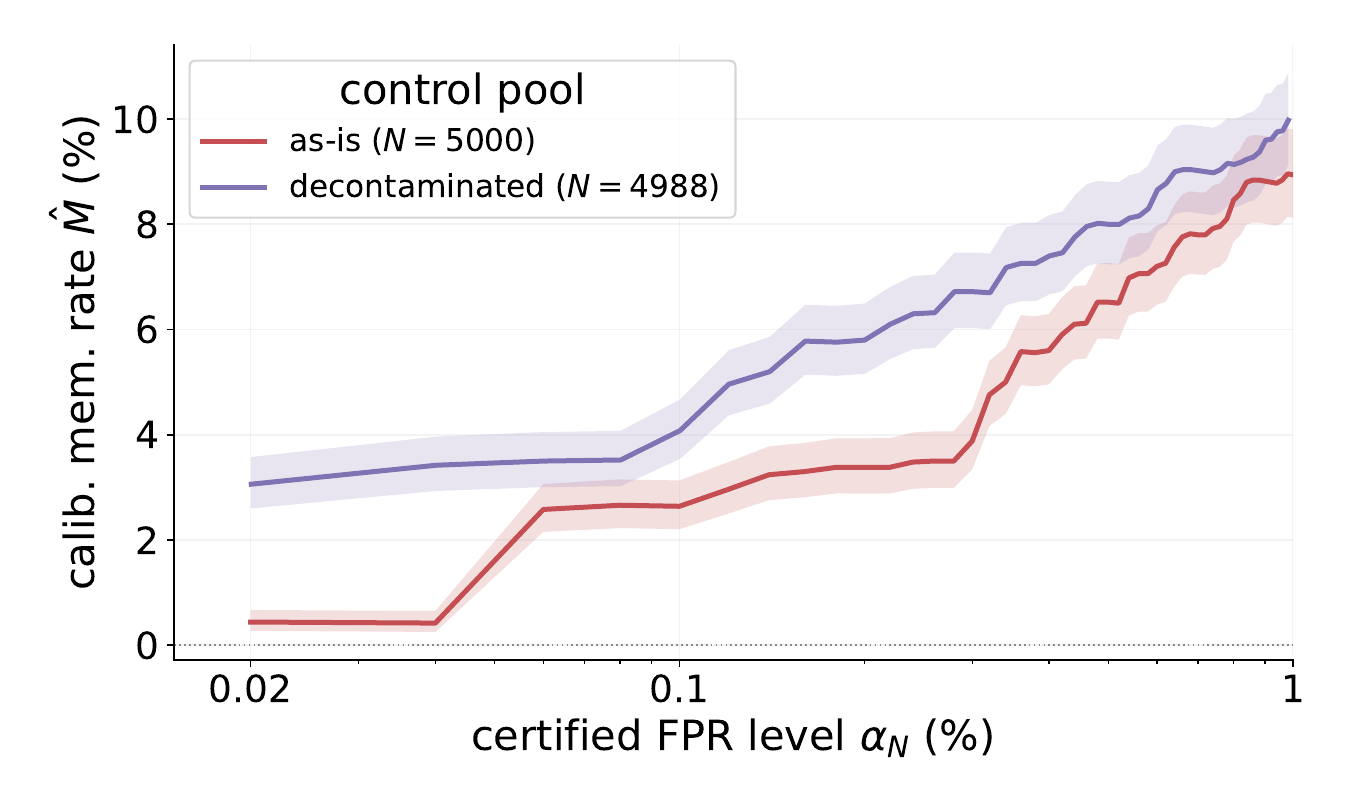}
  \caption{\textbf{Effect of decontaminating the matched Wikipedia controls.}
  We zoom in on the $50$-token, sub-$1\%$ FPR regime, comparing the as-is control pool ($N=5000$) against the decontaminated pool ($N=4988$).
  Decontamination drops $12$ sequences: the $2$ the model generates with probability above $10^{-3}$, and $10$ more found through an exhaustive, agent-enabled manual check.
  Contaminated controls sit at the top of the control distribution, so they raise the extreme-tail threshold and suppress the calibrated rate.
  Removing contaminated controls raises the calibrated rate from $0.44\%$ to $3.06\%$ at the tightest certified level ($\alpha_N\approx0.02\%$).
  Bands are $95\%$ Clopper--Pearson intervals on the member rate, shifted down by the constant $\alpha_N$ (\S\ref{app:sec:population:intervals}).}
  \label{app:fig:population-contamination}
\end{figure}

It turns out that some of the controls contain pre-cutoff (member) text that survived our post-cutoff filter:
the $2$ highest-scoring controls, and $10$ more we found through an exhaustive, agent-enabled manual check. 
Ten of these twelve sit at the top of the control distribution, where they raise the extreme-tail threshold; the remaining two score exactly $0$, so they don't move it.
Fig.~\ref{app:fig:population-contamination} shows how removing these $12$ verified contaminants changes the estimated calibrated rate.
The budget results in \S\ref{app:extractability:population}, and the $T=1$ experiments, including the model-scaling comparison (\S\ref{app:sec:population:policy}), calibrate on this decontaminated pool, which gives $\alpha_N = 4/4989 \approx 0.0802\%$.
The results in the main paper and the suffix-length sweep in \S\ref{app:sec:population:length} use the full pool ($N=5000$), for which $\alpha_N = 5/5001 \approx 0.09998\%$.
(The main results report the control construction and decontamination status explicitly rather than treating a post-cutoff date as sufficient evidence of non-membership.)

\subsection{Varying suffix length}\label{app:sec:population:length}

The main paper compares $10$- and $50$-token suffixes; 
here we sweep every suffix length from $1$ to $50$ tokens for OLMo~2 32B, scoring the same $5{,}000$ members and $5{,}000$ matched controls, with the same $50$-token prefixes, at every suffix length.
Throughout the sweep we use the full control pool ($N=5000$) rather than the decontaminated one (\S\ref{app:sec:population:contamination}).
Contamination isn't easily defined at short suffixes: 
at one or two tokens, nearly every control continuation likely appears somewhere in pre-cutoff text, so removing ``contaminated'' controls would remove the whole pool.

\begin{figure}[h!]
  \centering
  \begin{subfigure}[t]{0.49\linewidth}
    \centering
    \includegraphics[width=\linewidth]{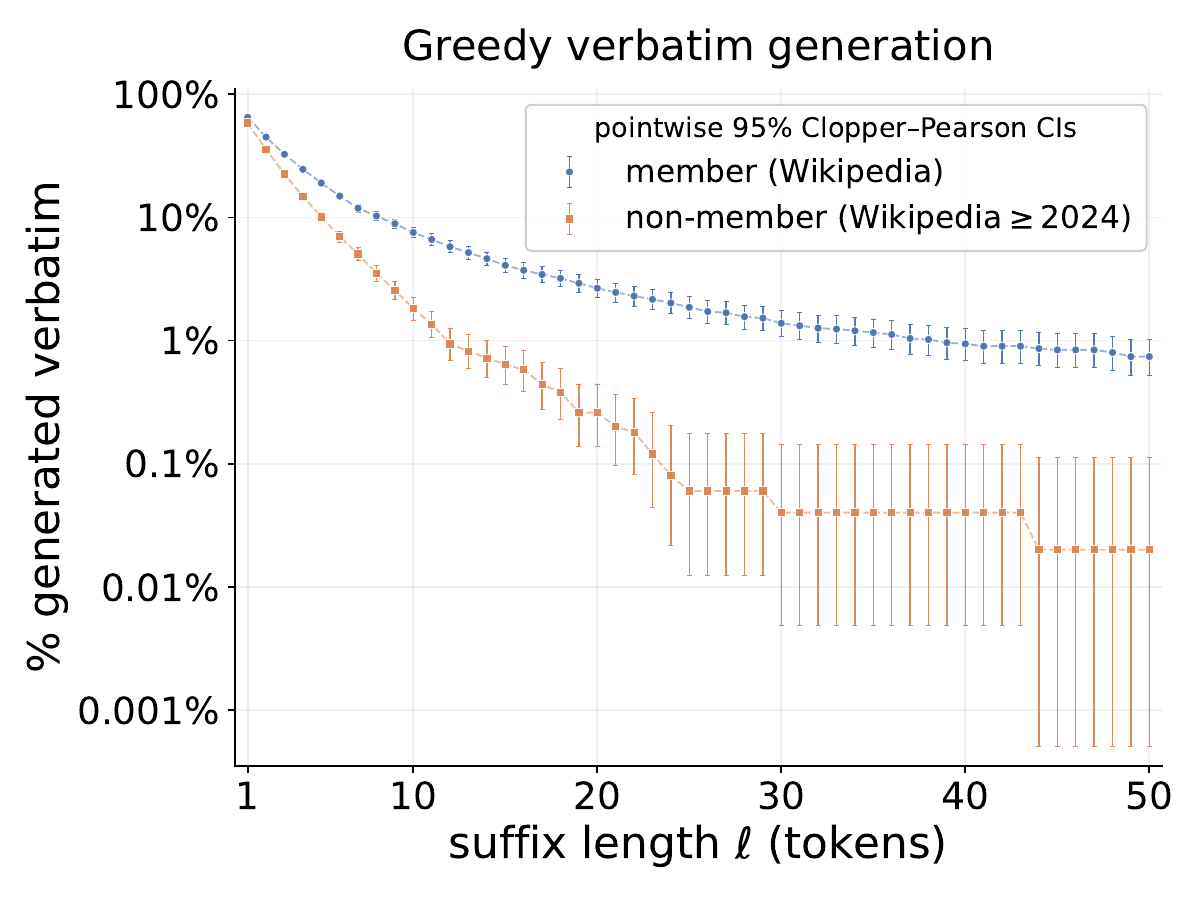}
    \caption{Member and non-member generation rates, with pointwise $95\%$ Clopper--Pearson intervals.}
    \label{app:fig:greedy-length-raw}
  \end{subfigure}
  \hfill
  \begin{subfigure}[t]{0.49\linewidth}
    \centering
    \includegraphics[width=\linewidth]{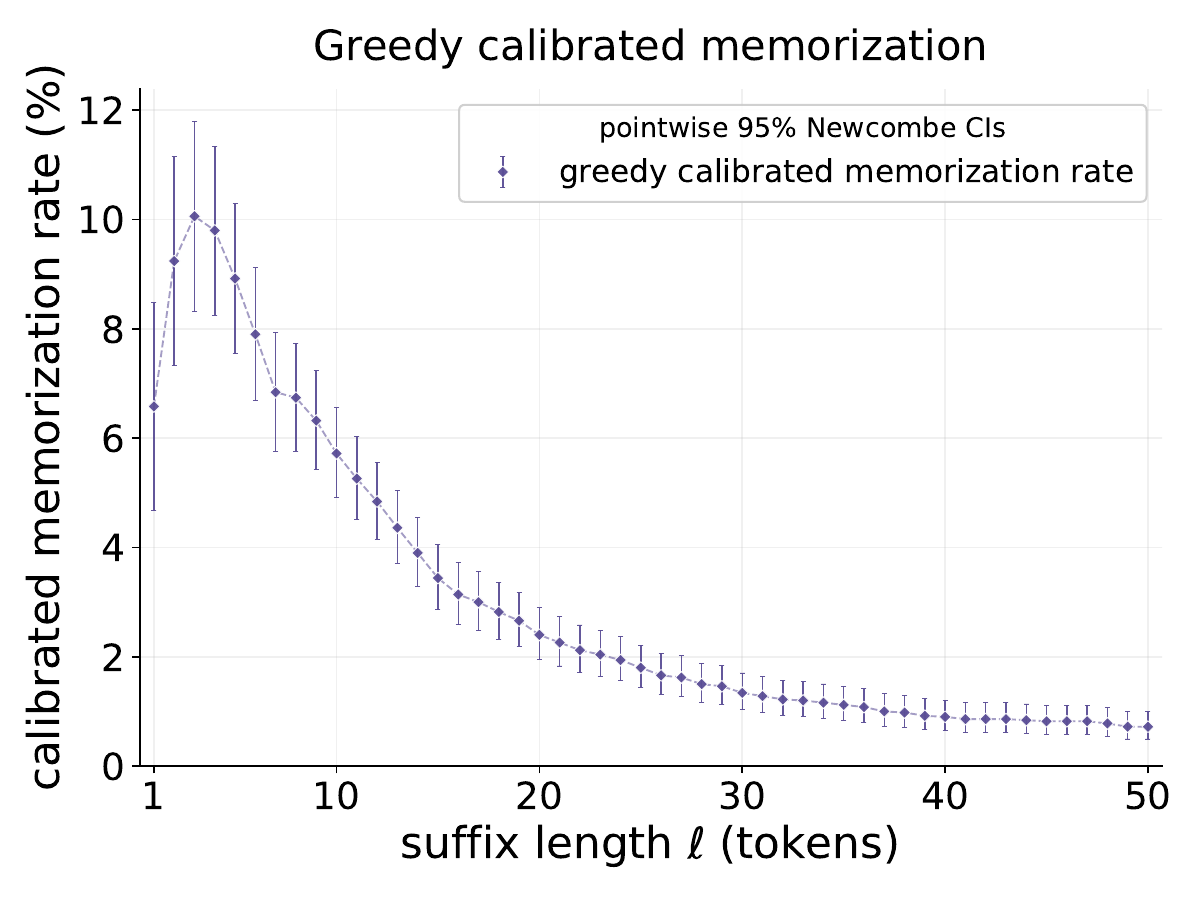}
    \caption{Member rate minus the in-sample non-member rate, with pointwise $95\%$ Newcombe independent-proportions intervals.}
    \label{app:fig:greedy-length-cmr}
  \end{subfigure}
  \caption{\textbf{Greedy discoverable extraction across suffix lengths (OLMo~2 32B).}
  Both rates fall with length, and the non-member rate falls faster: at $\ell=1$ it is $58\%$ against $65\%$ for members, and at $\ell=50$ it is $0.02\%$ against $0.74\%$.
  The difference in (\textbf{b}) therefore rises first, as the gap opens over the first few tokens, peaking near $\ell=3$ at ${\approx}10$ percentage points, and then falls as both rates shrink, to $0.72$ points at $\ell=50$.}
  \label{app:fig:greedy-length}
\end{figure}

Greedy decoding has no threshold to calibrate, so Fig.~\ref{app:fig:greedy-length} shows the two raw rates and their difference.
Both fall with length;
the non-member rate falls faster, and the difference peaks near $\ell=3$ before declining.
Fig.~\ref{fig:olmo-greedy-bar-10-vs-50} in the main paper reports two non-member quantities: the control bars give the in-sample rate, and the dashed line gives the certified FPR $\alpha_N=(m+1)/(N+1)$, where $m$ is the number of controls the model reproduces.
Fig.~\ref{app:fig:greedy-length-cmr} subtracts the in-sample rate, which differs from $\alpha_N$ by less than $1/(N+1)$.
Subtracting an estimated rate is why Fig.~\ref{app:fig:greedy-length-cmr} carries a Newcombe interval rather than a shifted one (\S\ref{app:sec:population:intervals}).
At the peak at $\ell=3$, the certified FPR is about $22\%$, so the difference there is between two enormous rates;
each point on the curve is its own operating point, at its own FPR, so heights aren't comparable across lengths.

\begin{figure}[h!]
  \centering
  \includegraphics[width=\linewidth]{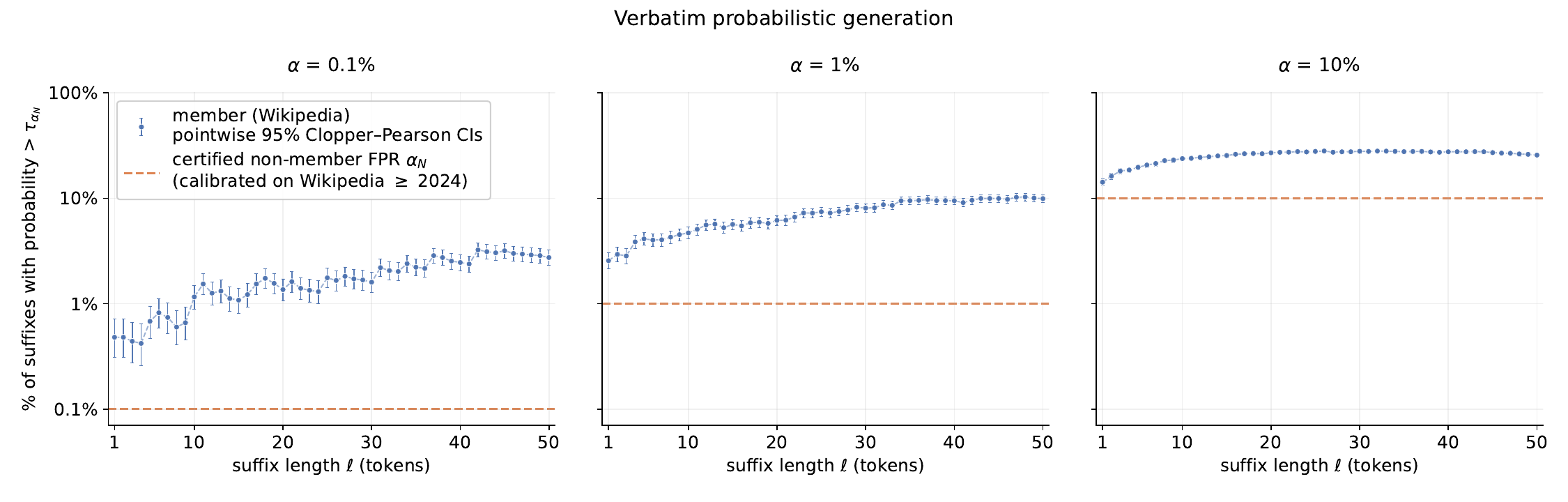}
  \caption{\textbf{Probabilistic member generation across suffix lengths at a fixed target FPR (OLMo~2 32B, full control pool).}
  For each target FPR ($\alpha=0.1\%$, $1\%$, $10\%$), the member rate is the share of members scoring above the threshold calibrated for that length; the dashed line is the certified FPR $\alpha_N$.
  Whiskers are pointwise $95\%$ Clopper--Pearson intervals.}
  \label{app:fig:prob-generation-length}
\end{figure}

\begin{figure}[h!]
  \centering
  \includegraphics[width=\linewidth]{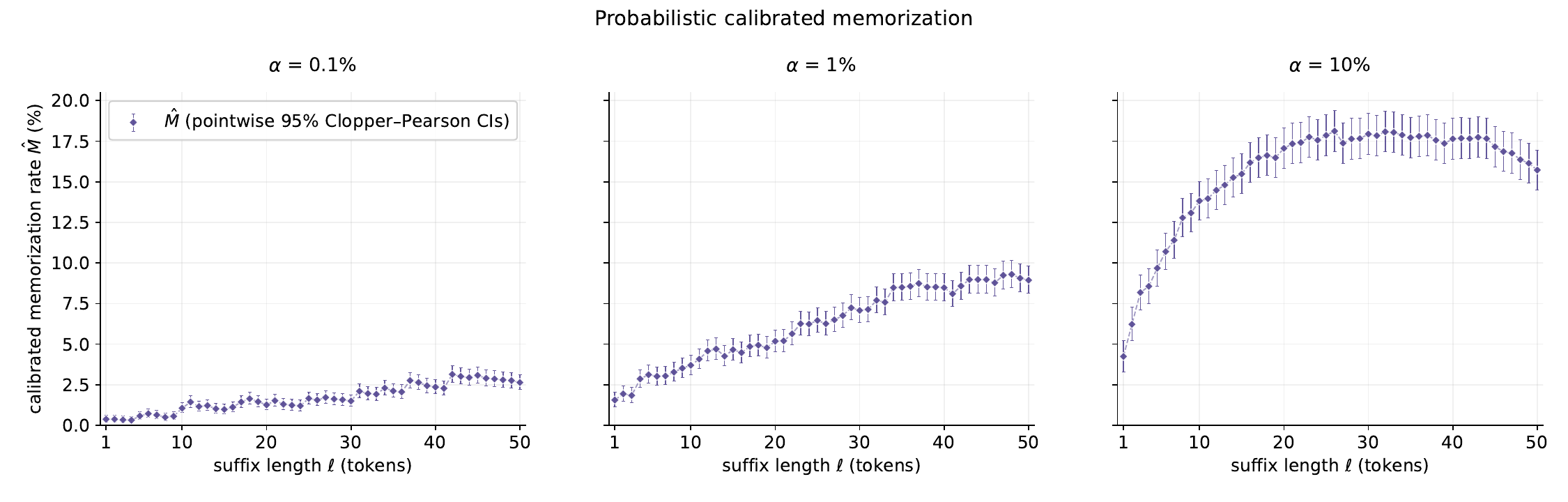}
  \caption{\textbf{Probabilistic calibrated memorization across suffix lengths (OLMo~2 32B, full control pool).}
  For each target FPR ($\alpha=0.1\%$, $1\%$, $10\%$), the calibrated rate $\hat M=\hat G-\alpha_N$ (Eq.~\ref{eq:emp-cmr}) is the member rate above the threshold calibrated for that length, minus the certified FPR $\alpha_N$.
  Whiskers are pointwise $95\%$ Clopper--Pearson intervals on $\hat G$, shifted down by $\alpha_N$.}
  \label{app:fig:prob-cmr-length}
\end{figure}

Probabilistic extraction behaves differently, because its score is continuous: at each length we can set the threshold to calibrate to a target FPR.
Fig.~\ref{app:fig:prob-generation-length} plots the member generation rate above the length-specific calibrated threshold at three targets ($\alpha=0.1\%$, $1\%$, $10\%$).
The member rate rises with length: 
at $\alpha=0.1\%$ it climbs from about $0.5\%$ at $\ell=1$ to about $2.7\%$ at $\ell=50$, and the curves for larger target FPRs ($1\%$, $10\%$) sit higher throughout.
This is the opposite behavior from greedy: 
as $\ell$ grows, non-member scores concentrate near zero, so the threshold that certifies a given FPR falls, and more members clear it.
Fig.~\ref{app:fig:prob-cmr-length} shows the corresponding calibrated rates $\hat M=\hat G-\alpha_N$; 
since $\alpha_N$ is a constant at each target FPR, these are the member curves shifted down by $\alpha_N$.
At $\alpha=0.1\%$ the calibrated rate rises to about $2.64$ percentage points by $\ell=50$ (the $2.64\%$ of \S\ref{sec:experiments:population}); 
at $\alpha=1\%$ to about $9\%$; at $\alpha=10\%$ it rises to about $18\%$ by $\ell\approx25$ and then levels off.
So a longer suffix certifies more memorization at a matched FPR even though its raw generation rate is lower, because at a common threshold the non-member rate collapses faster than the member rate.

\FloatBarrier

\subsection{Testing a different decoding policy}\label{app:sec:population:policy}
The main paper's results use decoding policies that produce scores with atoms.
Greedy's score is binary; 
under top-$k$ a verbatim candidate might not be reachable from expanding the top-$k$ tokens at every step, and in this case has generation probability exactly $0$.

Atoms have two consequences for calibration and probabilistic extraction.
First, the certified level has a ceiling: 
the largest attainable $\alpha_N$ is the share of controls with a non-zero score, which at $50$ tokens under top-$40$ is $10.4\%$ for 32B (the ``max $\alpha_N$'' annotation in Fig.~\ref{fig:olmo-calibrated-rate:mem}), $8.0\%$ for 13B, and $6.4\%$ for 7B.
Second, the Beta law for the population FPR (\S\ref{app:sec:conformal:alpha-star:beta-law}) assumes a continuous score;
\S\ref{app:sec:conformal:alpha-star:atoms} extends it to every level below the ceiling $\alpha^\dagger$.
(This works for top-$k$ because the atom is the block of zero scores, which sits at the very bottom of the score range, while the calibrated threshold is one of the highest control scores.
As long as at least $K$ controls score above zero, the threshold lands clear of the block.)

To include a decoding policy with no atoms, we repeat the experiment under the base policy, sampling from the full softmax at temperature $T=1$.
We do this for OLMo~2 7B, 13B, and 32B, at $10$ and $50$ tokens on the same $5{,}000$ members, calibrating on the decontaminated control pool ($N=4988$), so $\alpha_N=4/4989\approx0.080\%$ at the $0.1\%$ target FPR.
Under this policy, every sequence has positive generation probability, so there's no certified FPR ceiling, the score is continuous, and the Beta law applies directly at every certified level.

\begin{figure}[t]
  \centering
  \vspace{-.2cm}
    \begin{subfigure}[t]{0.66\textwidth}\centering
      \includegraphics[width=\linewidth]{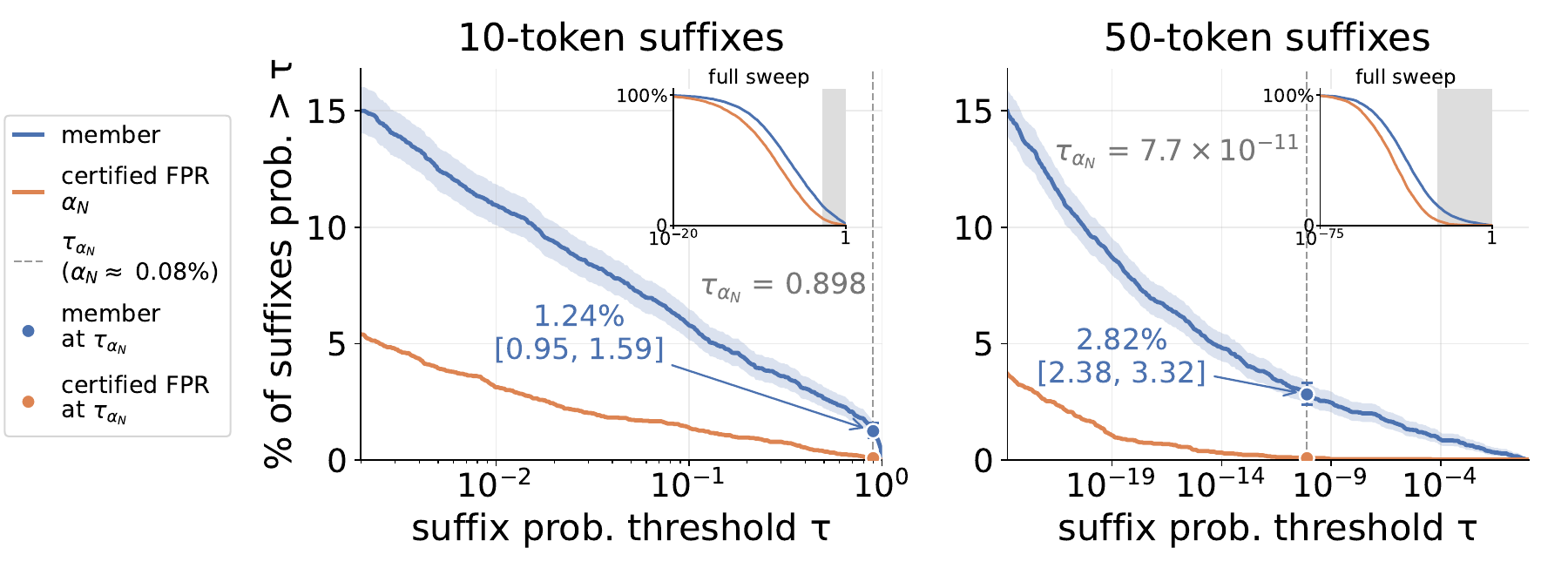}
      \vspace{-.25cm}
      \caption{7B: verbatim generation (same prefixes, $10$- and $50$-token suffixes)}
      \label{fig:olmo-calibrated-rate:gen-7b-t1}
    \end{subfigure}\hfill
    \begin{subfigure}[t]{0.33\textwidth}\centering
      \includegraphics[width=\linewidth]{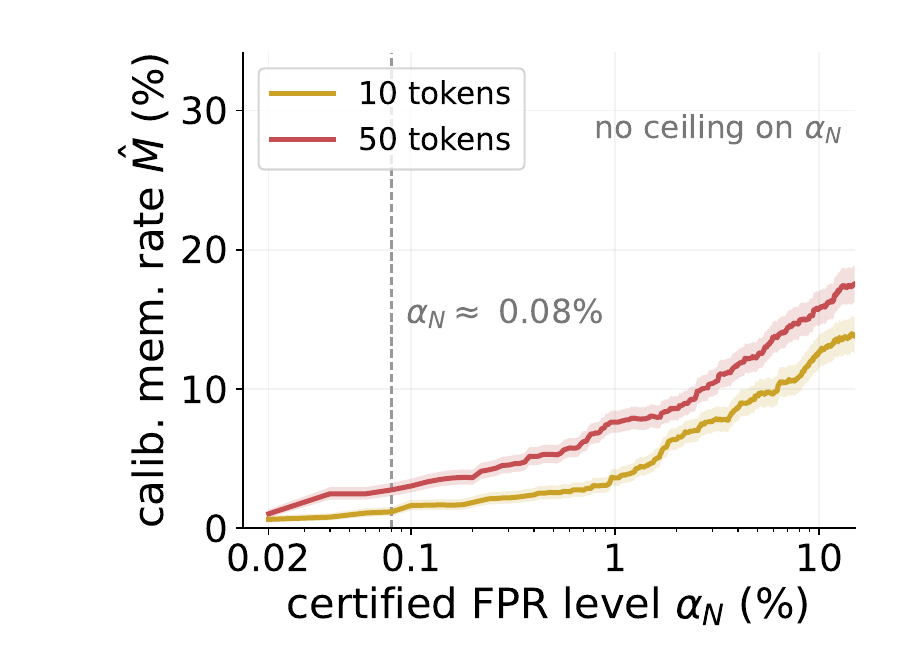}
      \vspace{-.25cm}
      \caption{7B: calibrated memorization}\label{fig:olmo-calibrated-rate:mem-7b-t1}
    \end{subfigure}\\
     \begin{subfigure}[t]{0.66\textwidth}\centering
      \includegraphics[width=\linewidth]{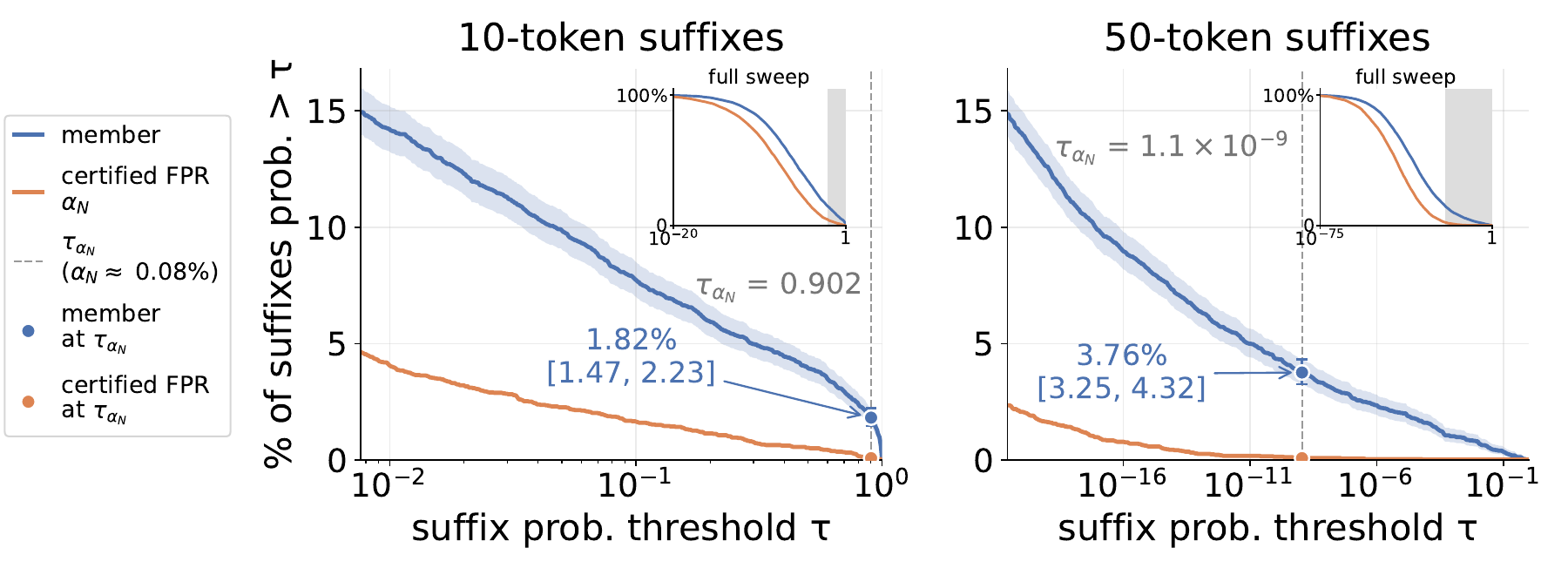}
      \vspace{-.25cm}
      \caption{13B: verbatim generation (same prefixes, $10$- and $50$-token suffixes)}
      \label{fig:olmo-calibrated-rate:gen-13b-t1}
    \end{subfigure}\hfill
    \begin{subfigure}[t]{0.33\textwidth}\centering
      \includegraphics[width=\linewidth]{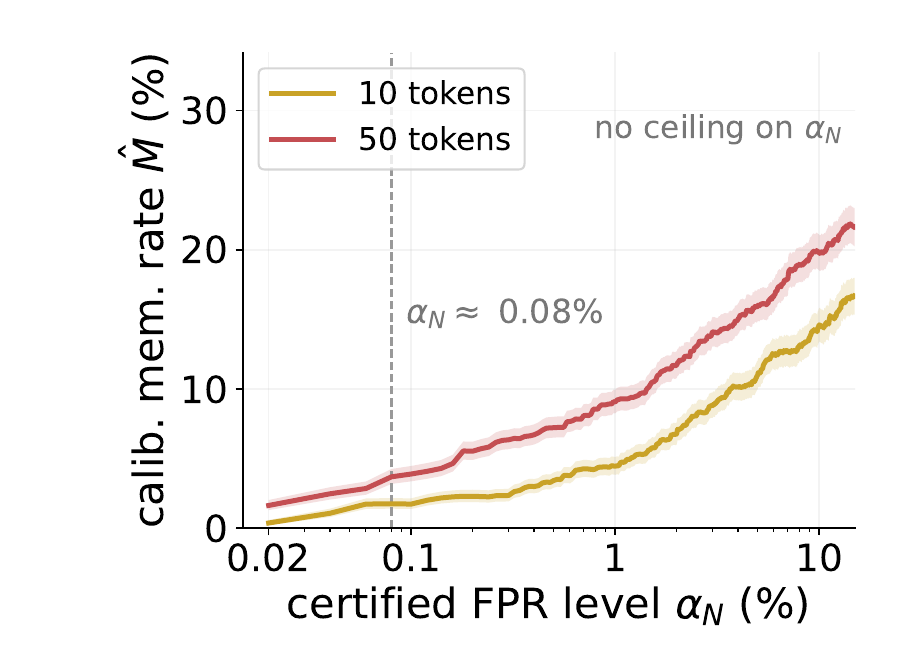}
      \vspace{-.25cm}
      \caption{13B: calibrated memorization}\label{fig:olmo-calibrated-rate:mem-13b-t1}
    \end{subfigure}\\
     \begin{subfigure}[t]{0.66\textwidth}\centering
      \includegraphics[width=\linewidth]{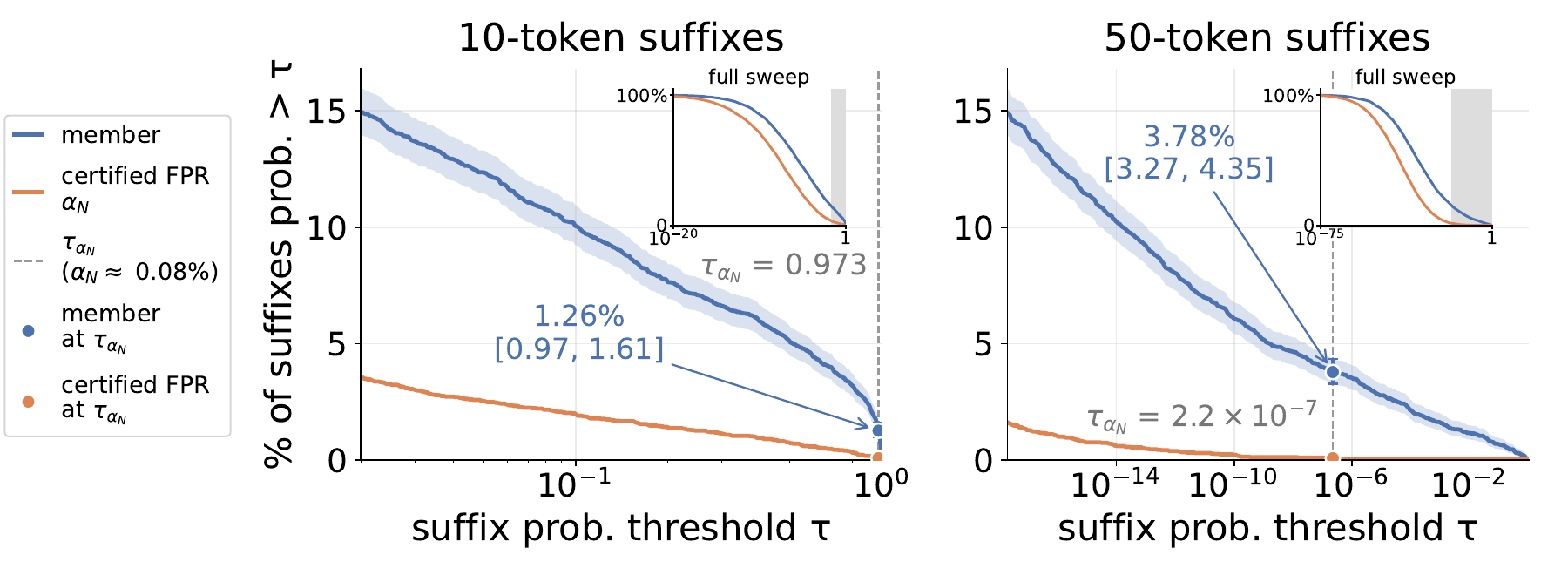}
      \vspace{-.25cm}
      \caption{32B: verbatim generation (same prefixes, $10$- and $50$-token suffixes)}
      \label{fig:olmo-calibrated-rate:gen-32b-t1}
    \end{subfigure}\hfill
    \begin{subfigure}[t]{0.33\textwidth}\centering
      \includegraphics[width=\linewidth]{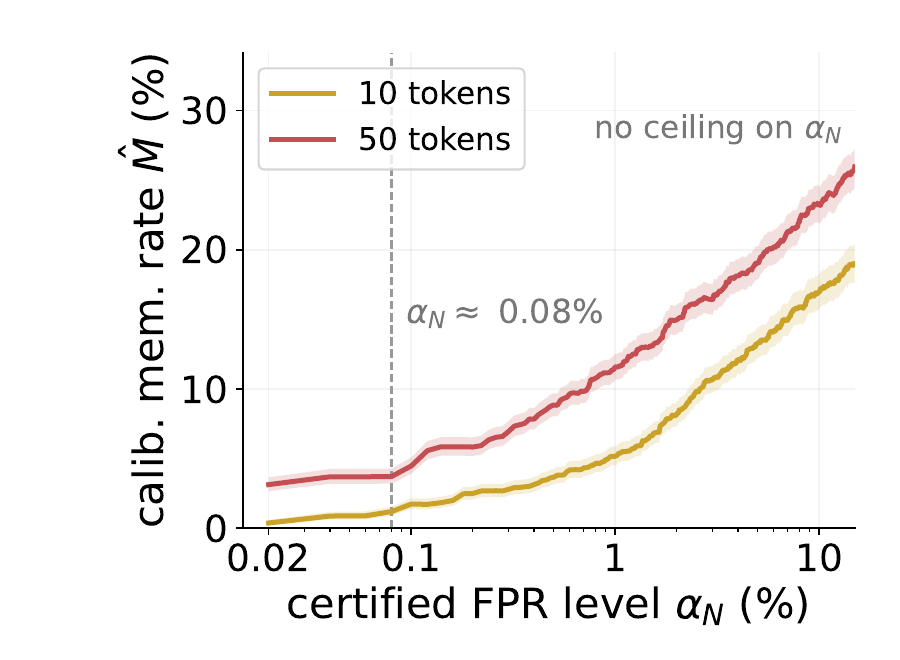}
      \vspace{-.25cm}
      \caption{32B: calibrated memorization}\label{fig:olmo-calibrated-rate:mem-32b-t1}
    \end{subfigure}
	  \caption{\textbf{Generation and calibrated memorization rate for OLMo~2 base models on Wikipedia ($T=1$).}
      We repeat the experiment shown in Fig.~\ref{fig:olmo-calibrated-rate} for OLMo~2 7B, 13B, and 32B and, instead of top-$k$ decoding ($k=40$, $T=1$), we use the base policy ($T=1$).
      With this decoding policy, no candidate or control sequences have $0\%$ generation probability, so there is no ceiling on $\alpha_N$.
      The top row is 7B, middle row is 13B, and the bottom row is 32B.
	  (All six panels calibrate on the decontaminated control pool, $N=4988$; see Fig.~\ref{app:fig:population-contamination}.
    This is why $\alpha_N=4/4989\approx0.0802\%$ here rather than the ${\approx}0.1\%$ of Fig.~\ref{fig:olmo-calibrated-rate}.)
      (\textbf{a}), (\textbf{c}), (\textbf{e}) Complementary CDFs of scores: 
      a point $(\tau, y)$ means $y\%$ of sequences score above $\tau$ (i.e., $\channel(\candseq \mid \promptseq) > \tau$). 
      Each panel is cropped to highlight the $\alpha_N\approx0.08\%$ operating point, beginning where the member rate falls to $15\%$.
      We visualize the three models this way because the plotted sweep spans roughly twenty-five orders of magnitude at $10$ tokens and a hundred at $50$, which leaves the threshold region unreadably compressed.
      The inset shows that full sweep, with the cropped window shaded.
      (\textbf{b}), (\textbf{d}), (\textbf{f}) 
	  Calibrated memorization (Eq.~\ref{eq:emp-cmr}) for $10$- and $50$-token suffixes across certified FPR levels $\alpha_N$. 
	  The $50$-token rate is the larger of the two across the sweep. 
    Bands are $95\%$ Clopper--Pearson intervals on the member rate, shifted by the constant $\alpha_N$ in (\textbf{b}), (\textbf{d}), (\textbf{f}).\looseness=-1
}
  \label{fig:olmo-calibrated-rate-t1}
  \vspace{-.3cm}
\end{figure}

Fig.~\ref{fig:olmo-calibrated-rate-t1} shows the results.
The conclusions of \S\ref{sec:experiments:population} hold under this policy for every model size: 
the $50$-token-suffix calibrated rate exceeds the $10$-token-suffix rate across the whole sweep.
At $\alpha_N\approx0.08\%$, the $50$-token calibrated rates are $2.74\%$, $3.68\%$, and $3.70\%$ for 7B, 13B, and 32B, compared to $1.16\%$, $1.74\%$, and $1.18\%$ at $10$ tokens.

\begin{figure}[t]
  \centering
  \begin{subfigure}[t]{0.49\linewidth}
    \centering
    \includegraphics[width=\linewidth]{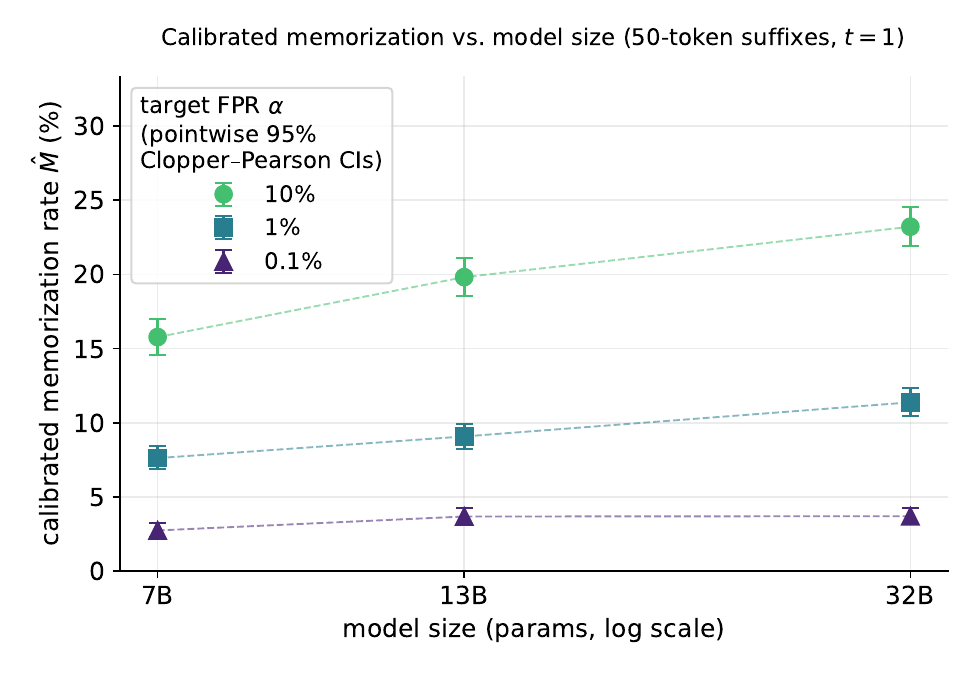}
    \caption{Absolute calibrated rates}
    \label{app:fig:population-scale:absolute}
  \end{subfigure}
  \hfill
  \begin{subfigure}[t]{0.49\linewidth}
    \centering
    \includegraphics[width=\linewidth]{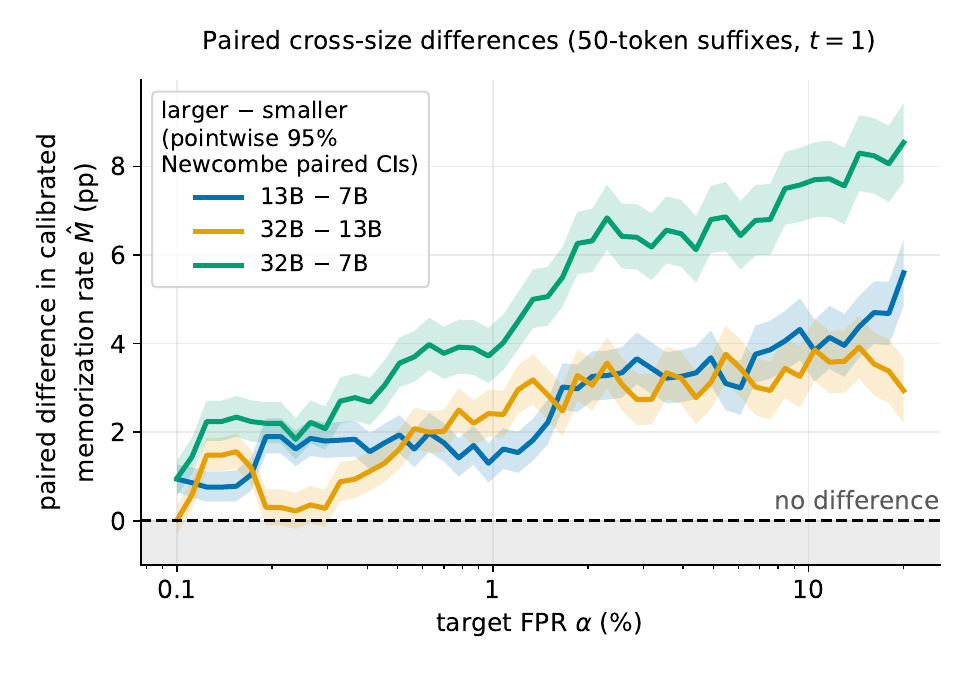}
    \caption{Paired cross-size differences}
    \label{app:fig:population-scale:diffs}
  \end{subfigure}
  \caption{\textbf{Calibrated verbatim memorization versus OLMo~2 model size ($T=1$).}
  The same $50$-token Wikipedia candidates and decontaminated controls ($N=4988$) are scored for the 7B, 13B, and 32B base models under the base policy ($T=1$).
  Left: absolute calibrated rates at three target FPRs, with pointwise $95\%$ Clopper--Pearson intervals shifted down by $\alpha_N$.
  Right: paired differences across the sweep of target FPRs, with pointwise Newcombe paired-proportion intervals~\citep{newcombe1998improved}; an interval above zero supports a larger calibrated rate for the larger model.}
  \label{app:fig:population-scale}
  \vspace{-.2cm}
\end{figure}

\paragraph{Model scaling.}
We also compare model sizes under the $T=1$ policy (Fig.~\ref{app:fig:population-scale}).
The same $J=5{,}000$ candidates are scored for 7B, 13B, and 32B.
Write $\hat M$ and $\hat M'$ for the calibrated memorization rates of two model sizes, with scores $\score, \score'$ and thresholds $\tau_{\alpha_N}, \tau'_{\alpha_N}$.
Both sizes calibrate on the same $N$ controls at the same target FPR, so they share the same $\alpha_N$.
Therefore,
\begin{align}
\hat M-\hat M'&=(\hat G-\alpha_N)-(\hat G'-\alpha_N)\nonumber\\
&=\hat G-\hat G'\nonumber\\
&=\frac{1}{J}\sum_{j=1}^{J}\Big(\mathbf{1}\{\score(\candseq_j)>\tau_{\alpha_N}\}-\mathbf{1}\{\score'(\candseq_j)>\tau'_{\alpha_N}\}\Big).
\label{app:eq:paired-diff}
\end{align}
In the last row, each summand is $+1$, $-1$, or $0$; 
it's $0$ exactly when the two sizes agree about candidate $j$.
So the difference is determined by the candidates the two sizes disagree on with respect to claiming memorization, and the band in Fig.~\ref{app:fig:population-scale:diffs} is an interval for that paired difference, $\hat M-\hat M'$ (Eq.~\ref{app:eq:paired-diff}).
The pairing therefore distinguishes a scale effect in memorization from the fact that larger models tend to be better predictors.

The calibrated rate rises with model size at every target FPR: 
at $\alpha=0.1\%$ it is $2.74\%$, $3.68\%$, and $3.70\%$ for 7B, 13B, and 32B; at $1\%$, $7.62\%$, $9.08\%$, and $11.38\%$; at $10\%$, $15.78\%$, $19.82\%$, and $23.22\%$ (Fig.~\ref{app:fig:population-scale:absolute}).
Fig.~\ref{app:fig:population-scale:diffs} plots the paired difference $\hat M-\hat M'$ (Eq.~\ref{app:eq:paired-diff}) for the three size pairs, sweeping the target FPR from $0.1\%$ to $20\%$.
In each pair, $\hat M$ is the calibrated rate of the larger size and $\hat M'$ that of the smaller one.
A positive value at a given level therefore means the larger size claims memorization for more of the shared candidates than the smaller one does, and the value is how many percentage points more.
We show pointwise $95\%$ Newcombe paired-proportion intervals (\S\ref{app:sec:population:intervals}).
At a given target FPR, if the interval lies entirely above zero, every difference it covers is positive, so the larger size's calibrated rate exceeds the smaller one's by at least the interval's lower endpoint.
If it includes zero, then the results don't resolve whether one of the two sizes has a higher calibrated memorization rate.

The paired differences are positive across that whole range for all three pairs.
Two of the three pairs have intervals entirely above zero at every one of those target FPRs: 13B minus 7B ($+0.94$ points $[0.64, 1.27]$ at $0.1\%$, $+4.04$ $[3.34, 4.74]$ at $10\%$) and 32B minus 7B ($+0.96$ $[0.57, 1.37]$ and $+7.44$ $[6.60, 8.28]$).
The third, 32B minus 13B, has an interval above zero at every target FPR from $0.33\%$ onwards ($+3.40$ $[2.71, 4.10]$ at $10\%$), but not at the tightest ones: 
at $0.1\%$ it is $+0.02$ $[-0.31, 0.35]$, with 13B and 32B at the same $3.7\%$.

\subsection{Margin}\label{app:sec:population:margin}

The calibrated threshold makes a binary claim, but the strength of evidence above it varies.
We can quantify this using a notion of \newterm{margin}:
how many orders of magnitude the candidate's score is above a threshold.
For a candidate $\candseq$ and a threshold $\tau$, define its margin in orders of magnitude by
\vspace{-.1cm}
\begin{equation}
\Delta(\candseq;\tau)
=
\log_{10}\frac{\score(\candseq)}{\tau}.
\label{app:eq:margin}
\end{equation}
Positive $\Delta$ means $\candseq$ clears $\tau$, and larger values place it farther above $\tau$.
Setting $\tau=\tau_{\alpha_N}$ gives the margin above the calibrated threshold. 

Two candidates in the same experiment are measured against the same threshold, so the one with the larger margin is the one with the higher score.
A large margin means the candidate's score sits orders of magnitude above what the matched non-members reached, which we can interpret as stronger evidence that the model memorized it.
A margin just above zero is the weakest evidence the test can claim on: 
the candidate clears the threshold, but it scores about as well as the control that set the threshold (and would be among the first to no longer be claimed memorized if a tighter $\alpha_N$ was used to set a different threshold). 

We can look directly at the scores to see which claimed candidates fall in the range where the controls that set the threshold also land (Fig.~\ref{fig:olmo-calibrated-rate:gen}).
Those are the claims that risk being false positives:
the share of the member generation rate that the certified FPR accounts for (\S\ref{sec:experiments:population}).
Margin is just another lens on this, but one that is normalized: dividing by the threshold sets it to zero, so a margin of $2$ means two orders of magnitude above the threshold whether that threshold is $8.9\times10^{-6}$ at $50$ tokens or $0.95$ at $10$ (\S\ref{sec:experiments:population}).
(\S\ref{sec:experiments:census} does this across books, comparing coverage five orders of magnitude above each book's own threshold.)

Greedy decoding has no margin to report, since its score is $0$ or $1$.
Every candidate that clears the threshold has the same score (\S\ref{app:sec:conformal:alpha-star:practice}).

%% file: section/iclr/appendix/400-census.tex
\clearpage
\section{Census experiments}\label{app:census}

We run census experiments on the four matched pairs of books shown in Tab.~\ref{app:tab:census-books}.
For each author, we include a book known to be in Llama 3.1's training data from the Books3 corpus and a book published after Llama 3.1's training cutoff. 
For Dan Brown and Suzanne Collins, the respective pairs of books are from the same series. 
We score every window with the sliding-window verbatim probabilistic extraction procedure of \citet{cooper2025books}, under top-$k$ decoding with $k=40$.
We use this policy because it's the one that prior work on books uses.
Because every window of each book is scored, the coverage we report is the exact value for that book rather than an estimate, so none of our census results carries an interval (contrast the plotted intervals in \S\ref{app:sec:population:intervals}).

\begin{table}[h]
\centering
\footnotesize
\caption{\textbf{Matched book pairs.}
Llama~3.1's pretraining cutoff is December~2023.
Each Books3 title is paired with a same-author book published after the cutoff.}
\label{app:tab:census-books}
\begin{tabular}{@{}p{0.16\linewidth}p{0.37\linewidth}p{0.37\linewidth}@{}}
\toprule
\textbf{Author} & \textbf{Member (In-training from Books3)} & \textbf{Non-member (Post-cutoff control)} \\
\midrule
Dan Brown &
\emph{The Da Vinci Code}~\citep{The_Da_Vinci_Code} &
\emph{The Secret of Secrets}~\citep{The_Secret_of_Secrets} \\
Suzanne Collins &
\emph{The Hunger Games}~\citep{The_Hunger_Games} &
\emph{Sunrise on the Reaping}~\citep{Sunrise_on_the_Reaping} \\
Ta-Nehisi Coates &
\emph{We Were Eight Years in Power}~\citep{We_Were_Eight_Years_in_Power} &
\emph{The Message}~\citep{The_Message}\\
Joseph Finder &
\emph{Killer Instinct}~\citep{Killer_Instinct} &
\emph{The Oligarch's Daughter}~\citep{The_Oligarchs_Daughter} \\
\bottomrule
\end{tabular}
\end{table}

\subsection{Calibrating thresholds and decontamination}\label{app:census:calibration}

For a book pair, the calibrated threshold is the largest score among decontaminated windows of the post-cutoff book, rounded up to the next power of ten.
That is, denote a control book's set of sliding-window sequences to be $\mathsf{control}$.
Then 
\begin{equation}
\label{app:eq:book-floor}
\tau^{(\mathrm{book})}
=
10^{\left\lceil
\log_{10}\max_{w'\in\mathsf{control}}\score(w')
\right\rceil}.
\end{equation}
Rounding upward is conservative. 
It avoids attaching meaning to an unreliable floating-point mantissa at probabilities near $10^{-20}$ and below.

As part of calibrating the threshold, we need to decontaminate the matched control books.
This is because, even when a book post-dates training, individual passages may quote training data.
For decontamination, we inspect the high-scoring control tail, decode the underlying text, and remove only passages verified as being reused from earlier sources; 
the in-training book is never culled.
We developed a rigorous, deliberately rote process for culling, in which we used Claude to perform verified searches on the internet to identify provenance of the high-scoring windows.
The process was multi-turn and strictly sequential: rather than assessing all candidate passages at once, each was worked through in its own step.
Every single identified passage was manually verified by a person before culling. 

For \emph{Sunrise on the Reaping}, for example, the removed passages include repeated lines from Edgar Allan Poe's ``The Raven'' and quotations from \emph{The Hunger Games}.
We use each control book's highest-scoring own-prose passage to set the threshold (Eq.~\ref{app:eq:book-floor}). 

\begin{figure}[p]
\centering
\begin{subfigure}{\linewidth}
  \centering
  \makebox[0.49\linewidth]{\small\bf before decontamination}\hfill
  \makebox[0.49\linewidth]{\small\bf after decontamination}\\[1pt]
  \includegraphics[width=0.49\linewidth]{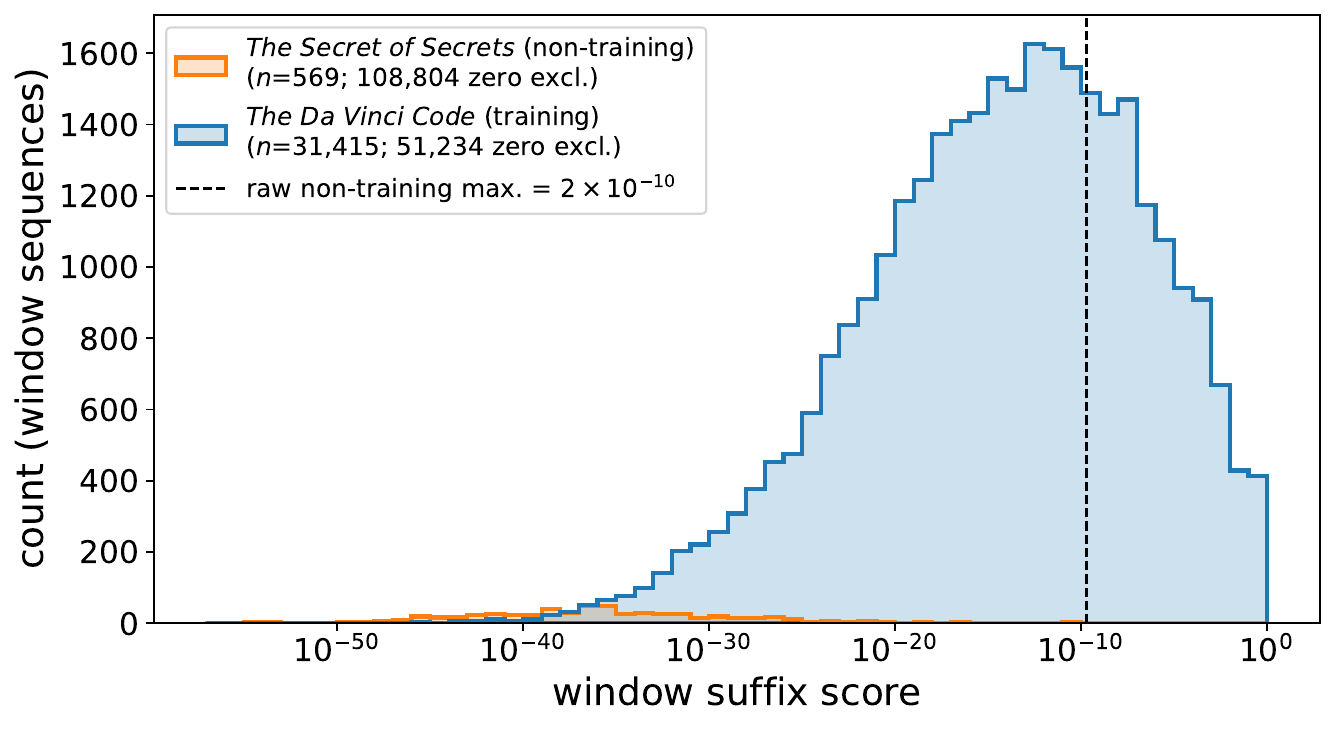}\hfill
  \includegraphics[width=0.49\linewidth]{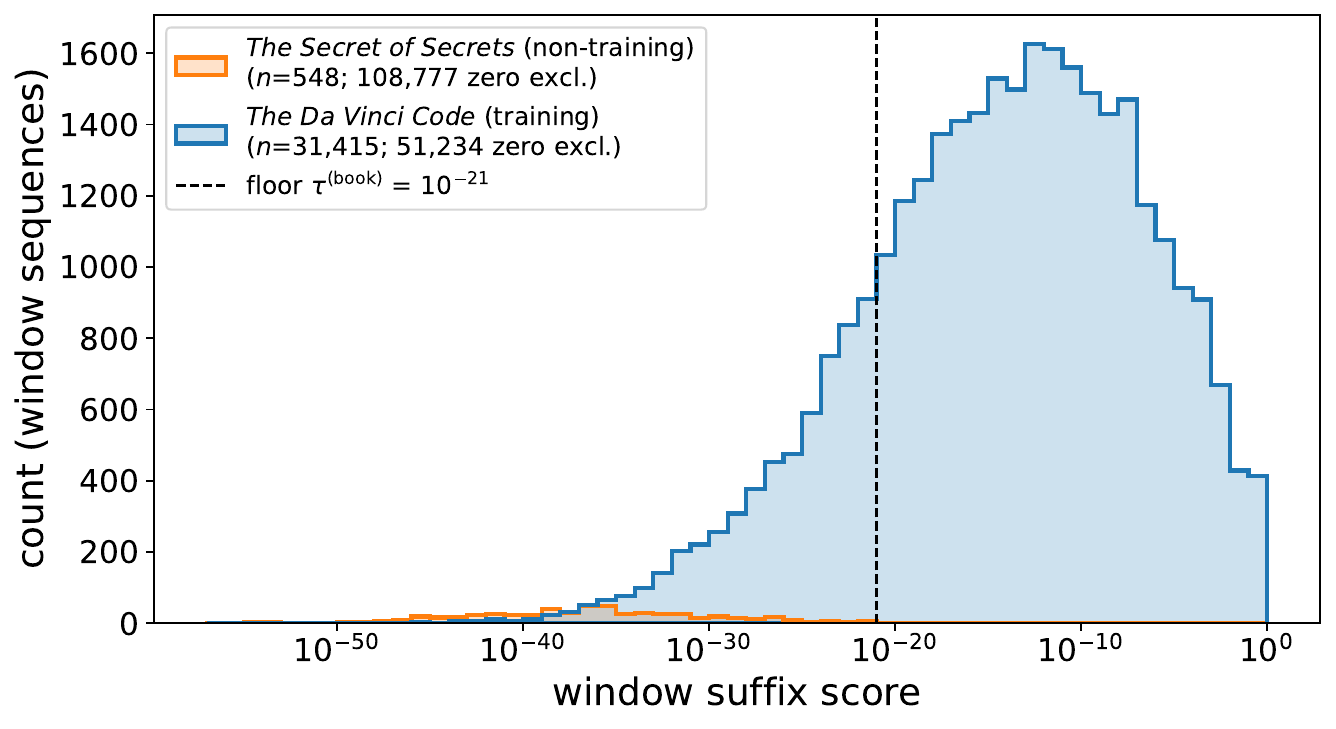}\\[2pt]
  \includegraphics[width=0.49\linewidth]{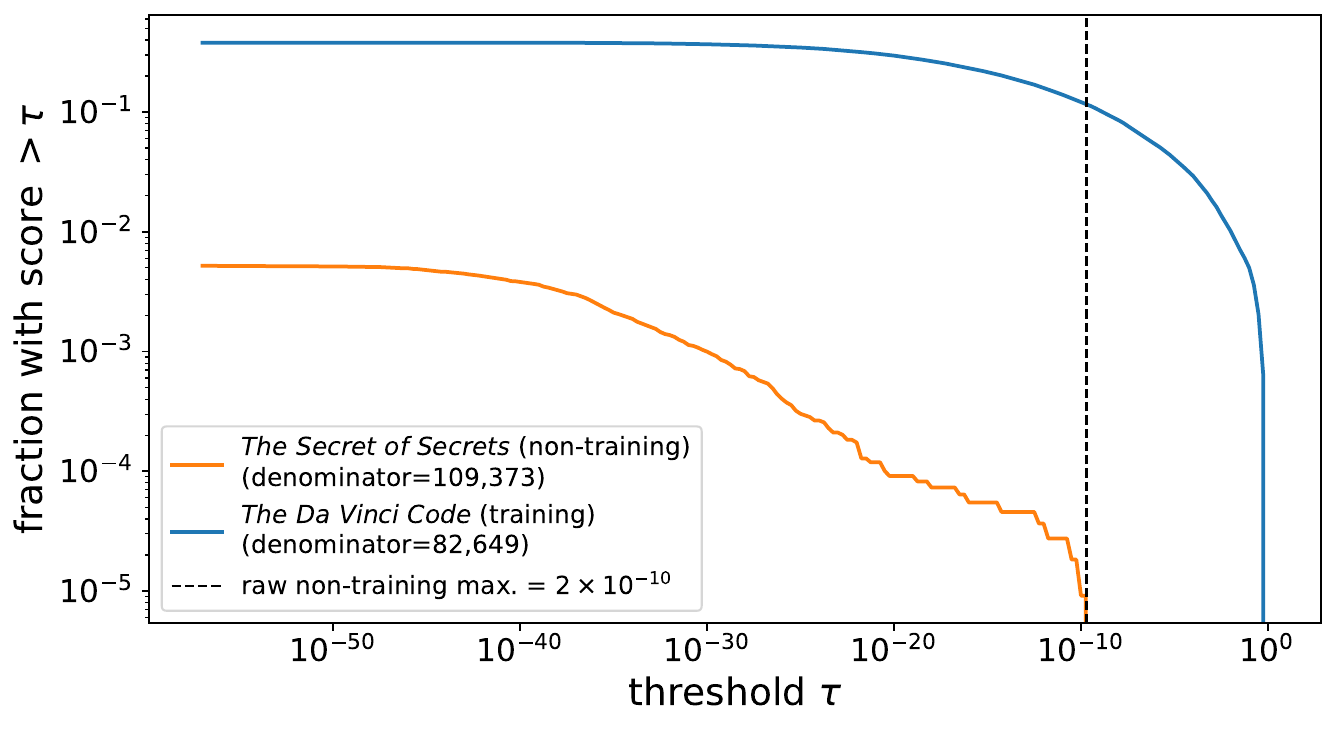}\hfill
  \includegraphics[width=0.49\linewidth]{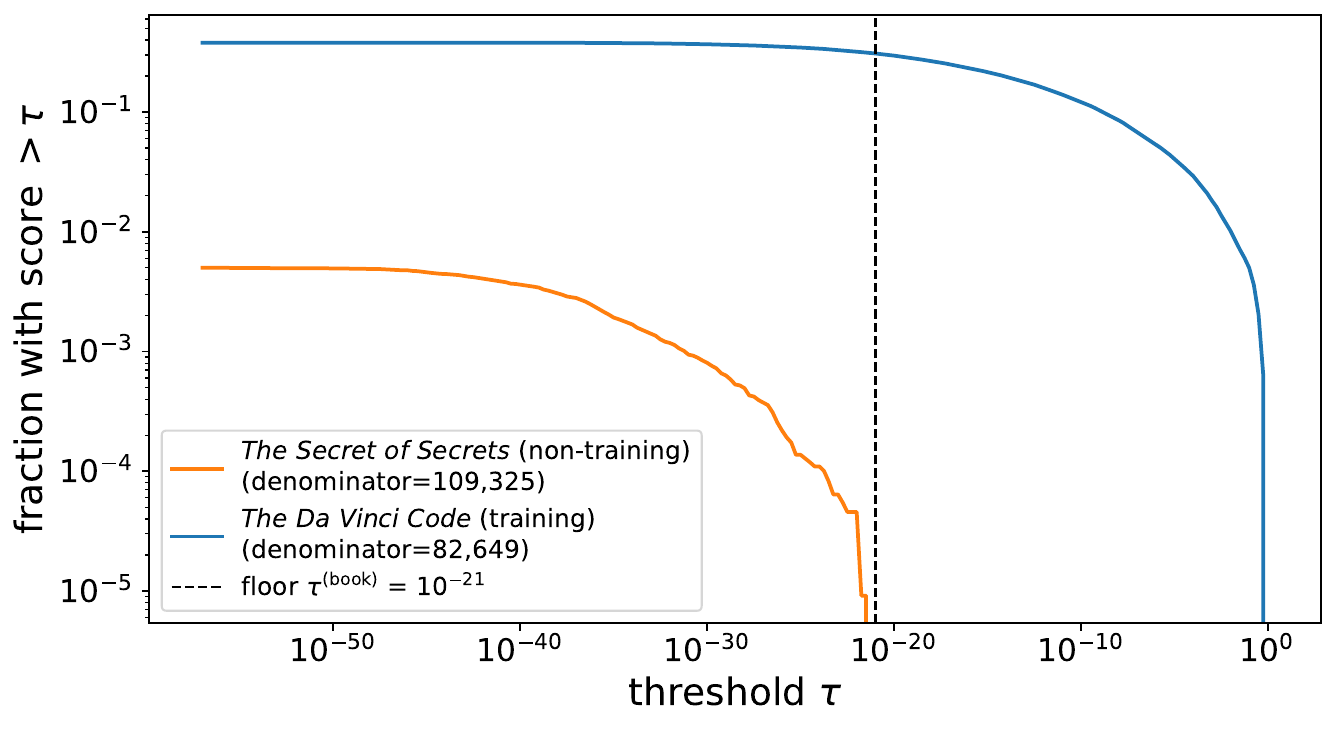}
  \caption{Dan Brown: \emph{The Secret of Secrets} (non-member) and \emph{The Da Vinci Code} (member).}
\end{subfigure}

\medskip

\begin{subfigure}{\linewidth}
  \centering
  \makebox[0.49\linewidth]{\small\bf before decontamination}\hfill
  \makebox[0.49\linewidth]{\small\bf after decontamination}\\[1pt]
  \includegraphics[width=0.49\linewidth]{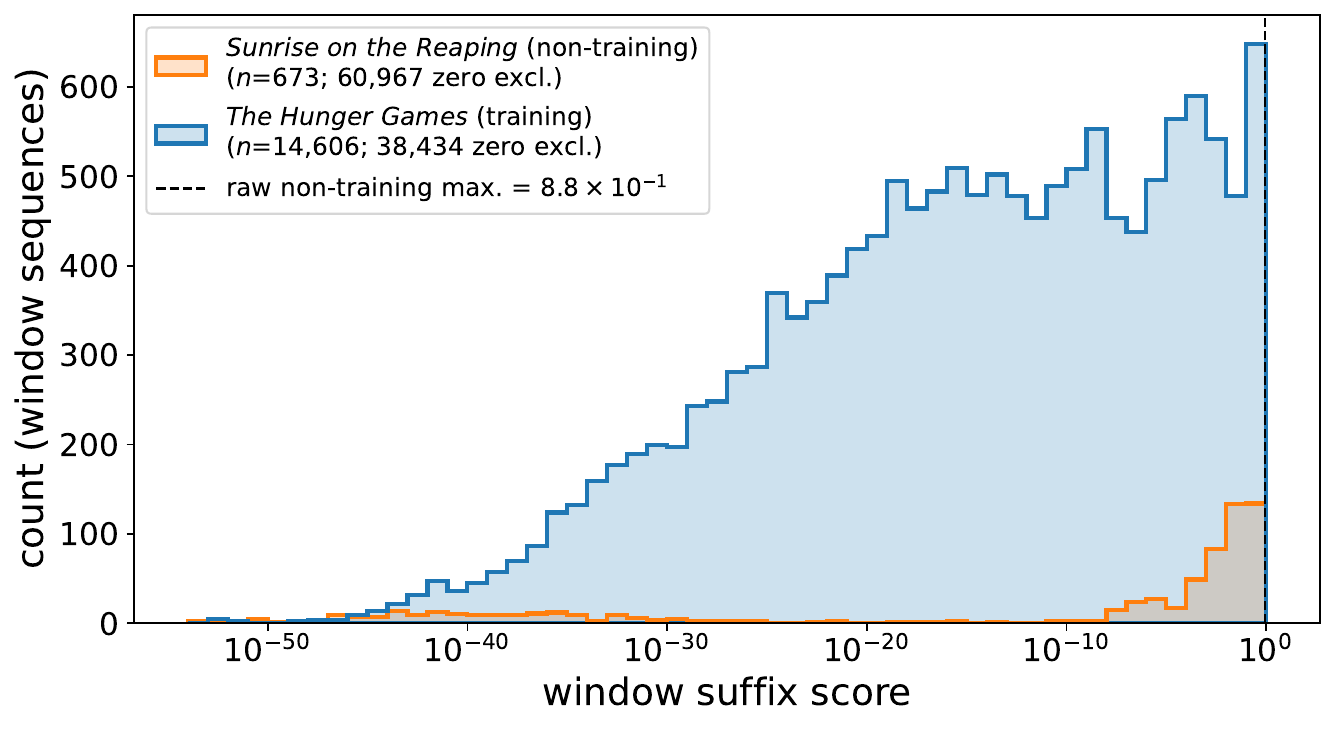}\hfill
  \includegraphics[width=0.49\linewidth]{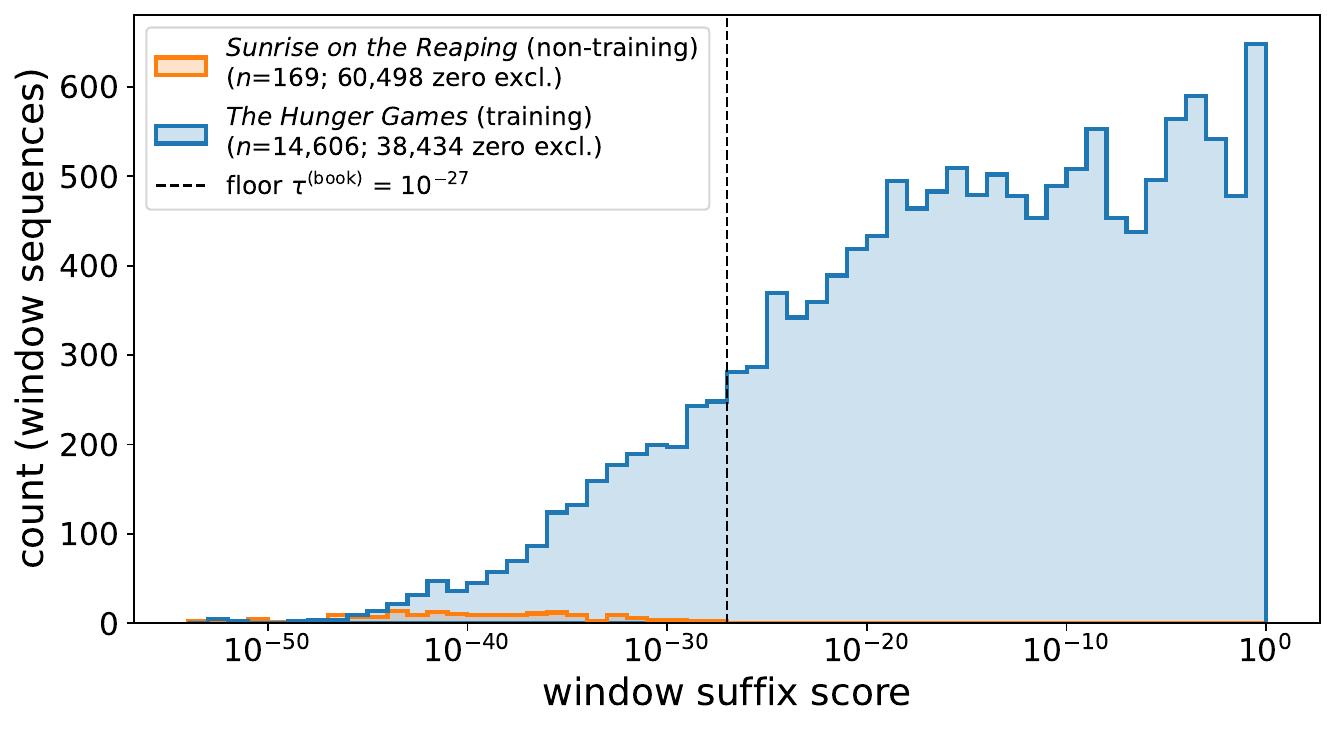}\\[2pt]
  \includegraphics[width=0.49\linewidth]{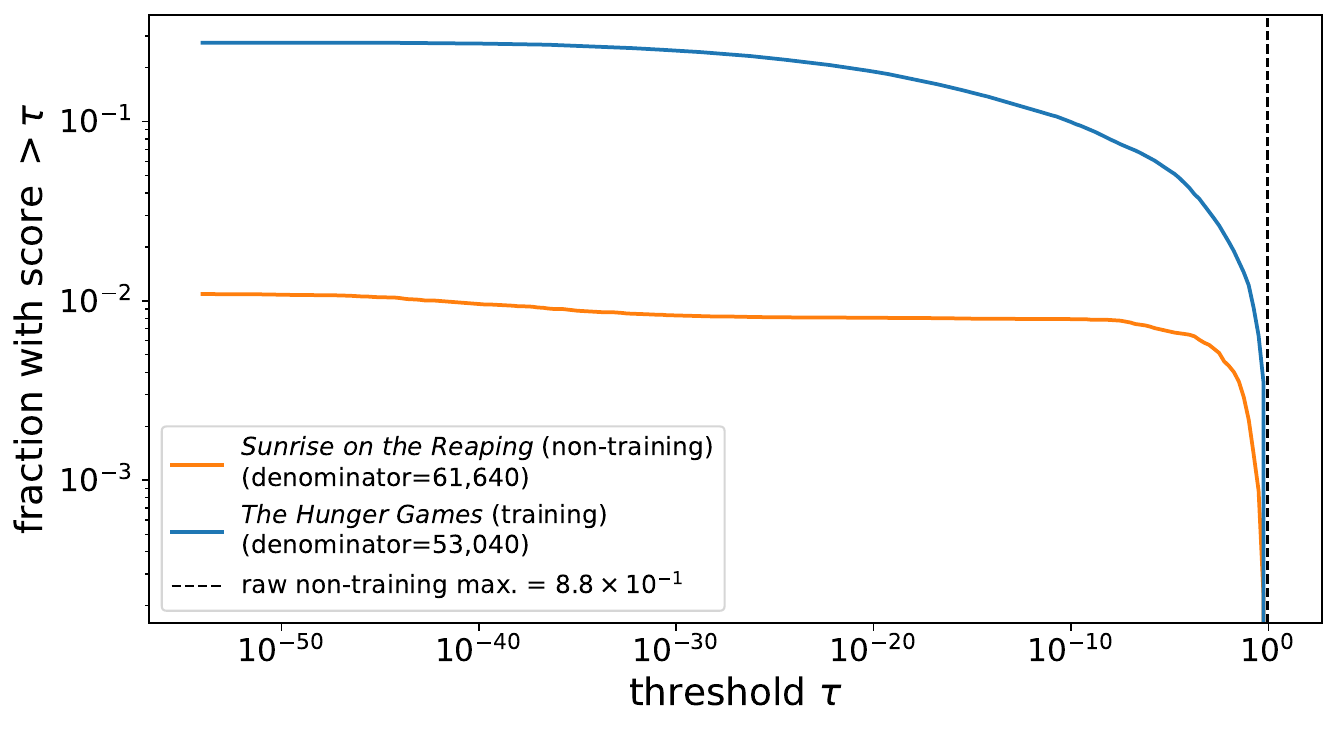}\hfill
  \includegraphics[width=0.49\linewidth]{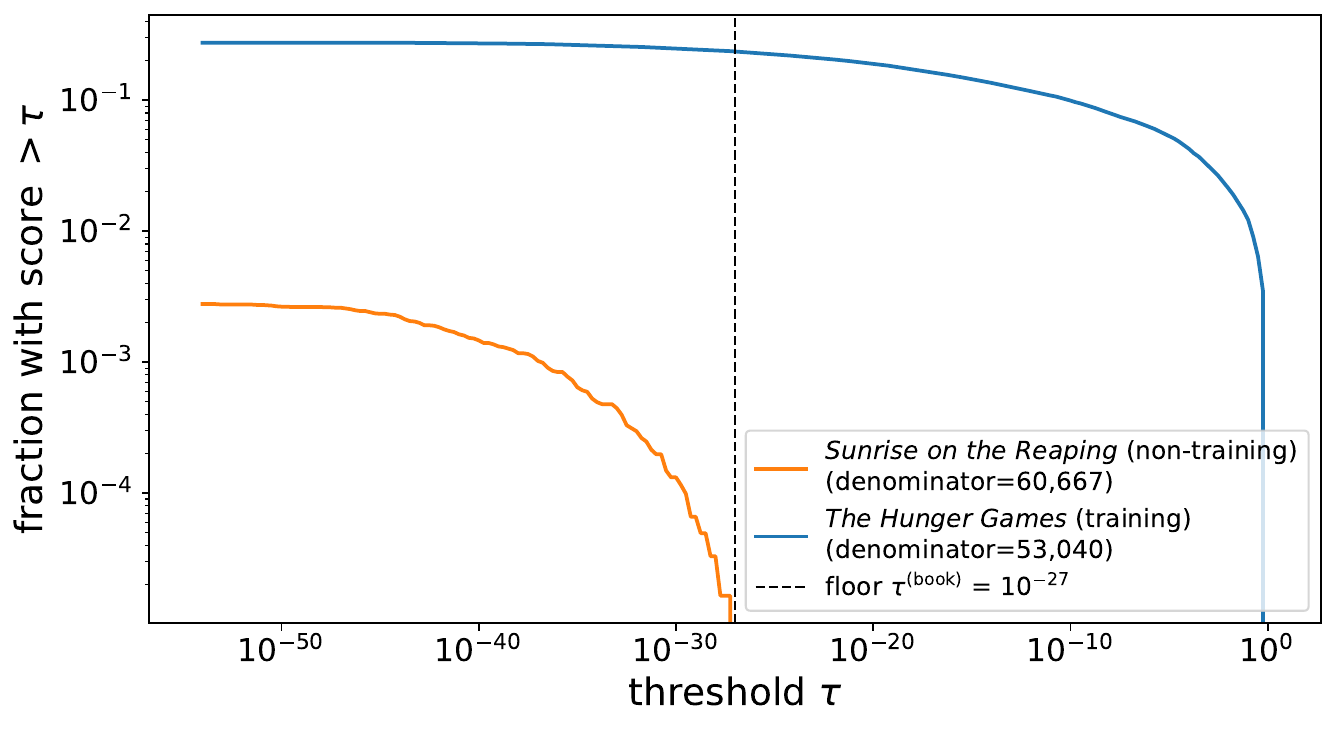}
  \caption{Suzanne Collins: \emph{Sunrise on the Reaping} (non-member) and \emph{The Hunger Games} (member).}
\end{subfigure}
\caption{\textbf{Matched-control decontamination for the Brown and Collins pairs.}
The top row within each pair is the score histogram and the bottom row the complementary CDF, before and after verified reused passages are removed.}
\label{app:fig:decontam-brown-collins}
\end{figure}

\begin{figure}[p]
\centering
\begin{subfigure}{\linewidth}
  \centering
  \makebox[0.49\linewidth]{\small\bf before decontamination}\hfill
  \makebox[0.49\linewidth]{\small\bf after decontamination}\\[1pt]
  \includegraphics[width=0.49\linewidth]{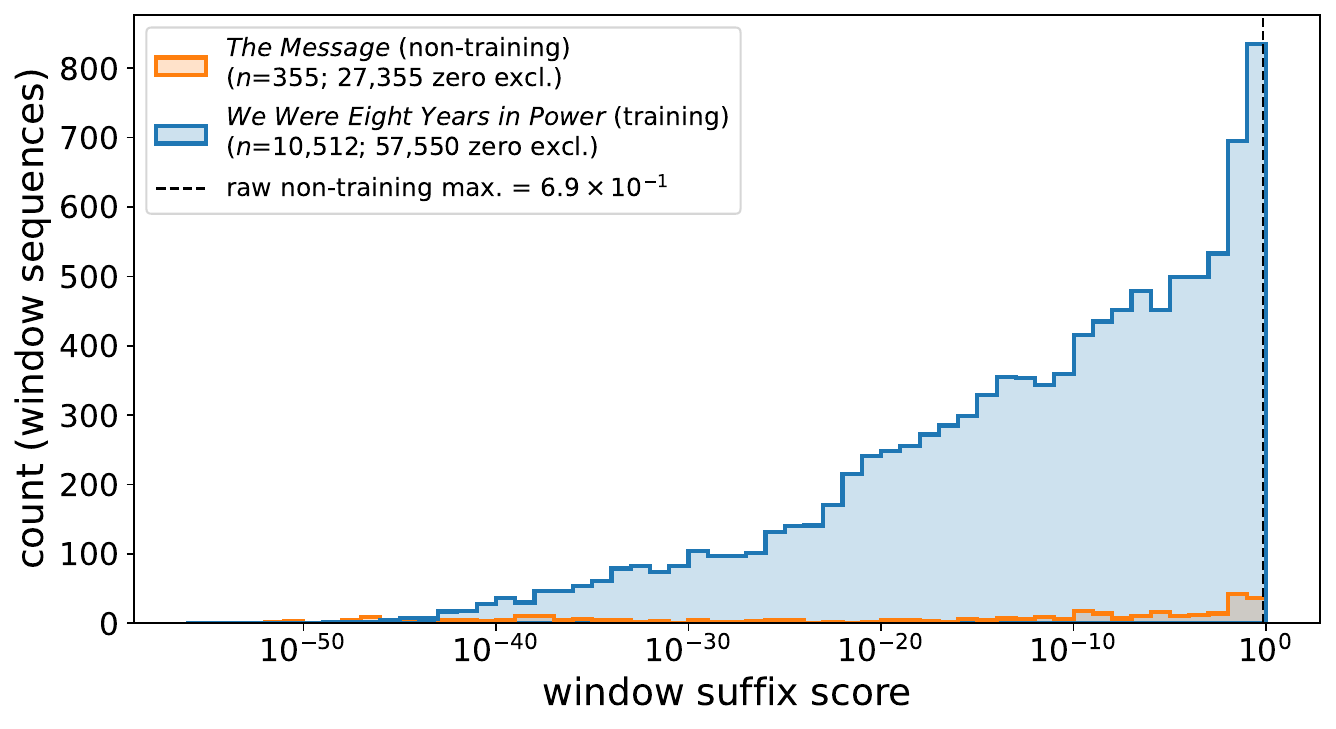}\hfill
  \includegraphics[width=0.49\linewidth]{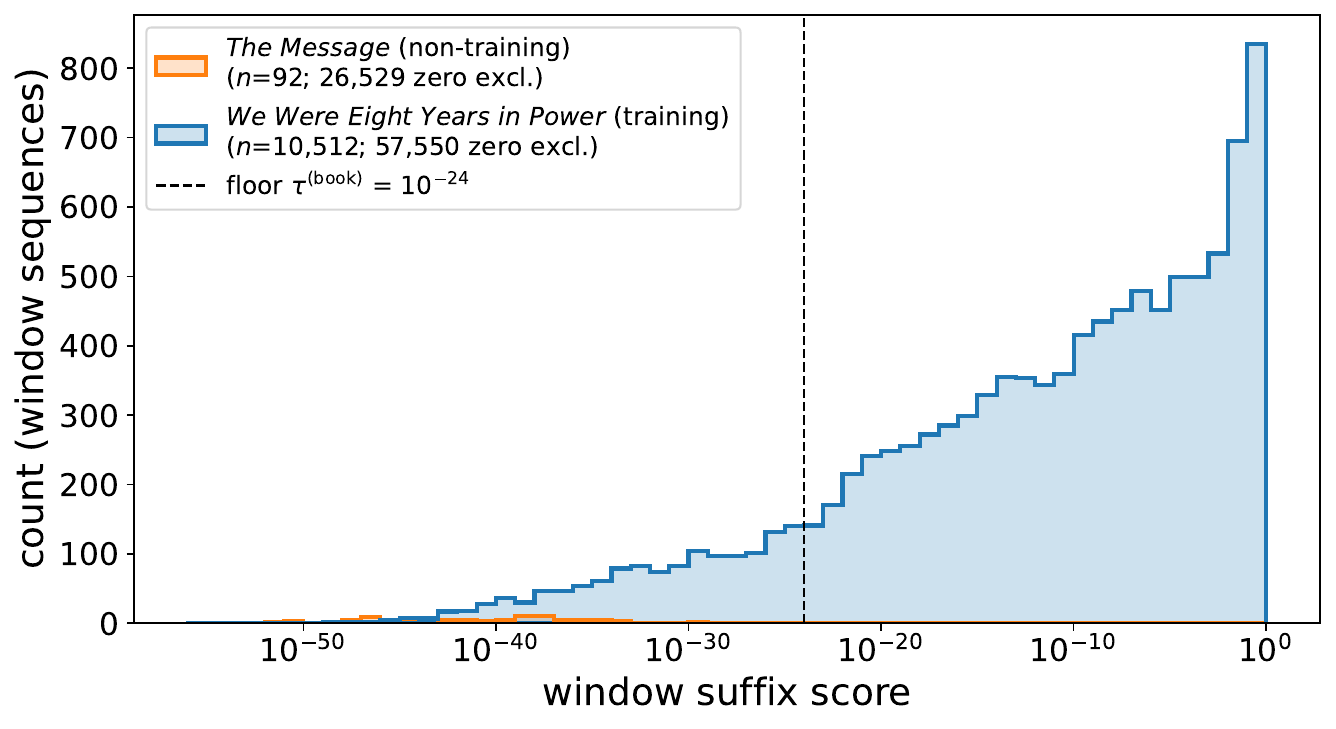}\\[2pt]
  \includegraphics[width=0.49\linewidth]{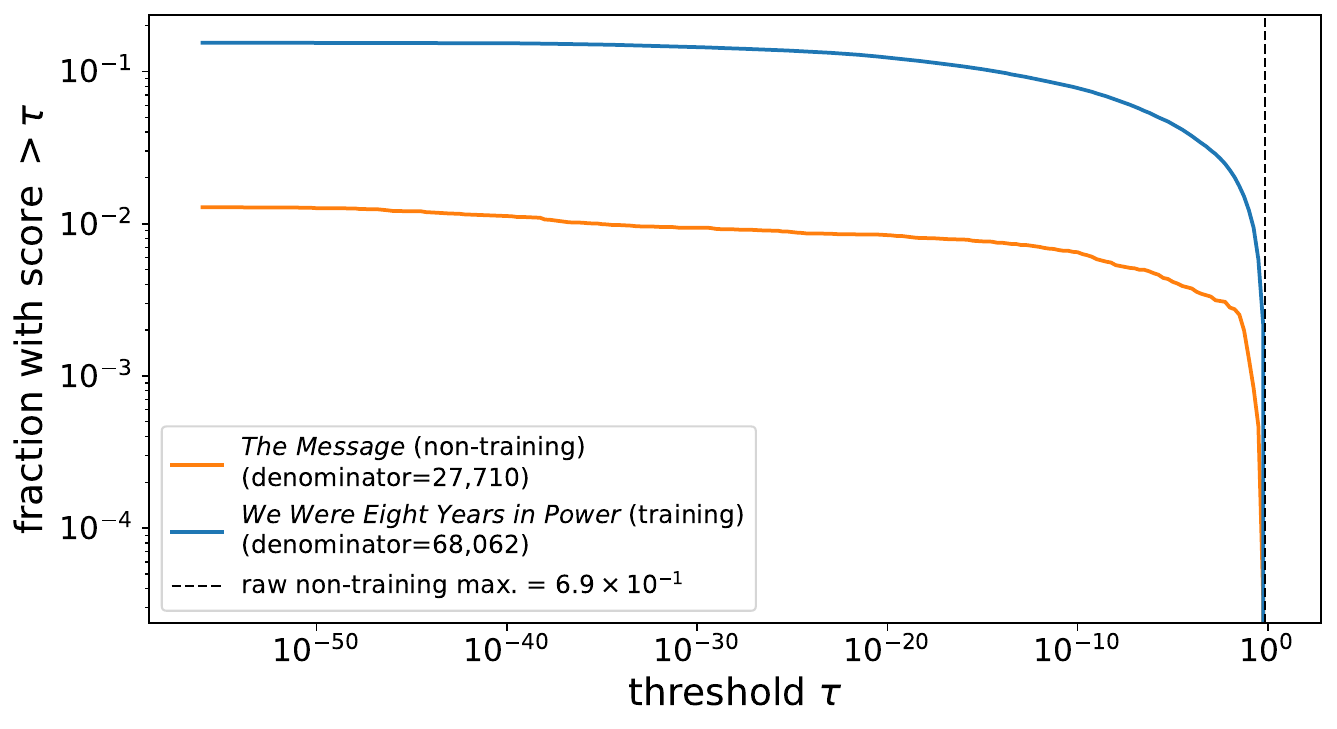}\hfill
  \includegraphics[width=0.49\linewidth]{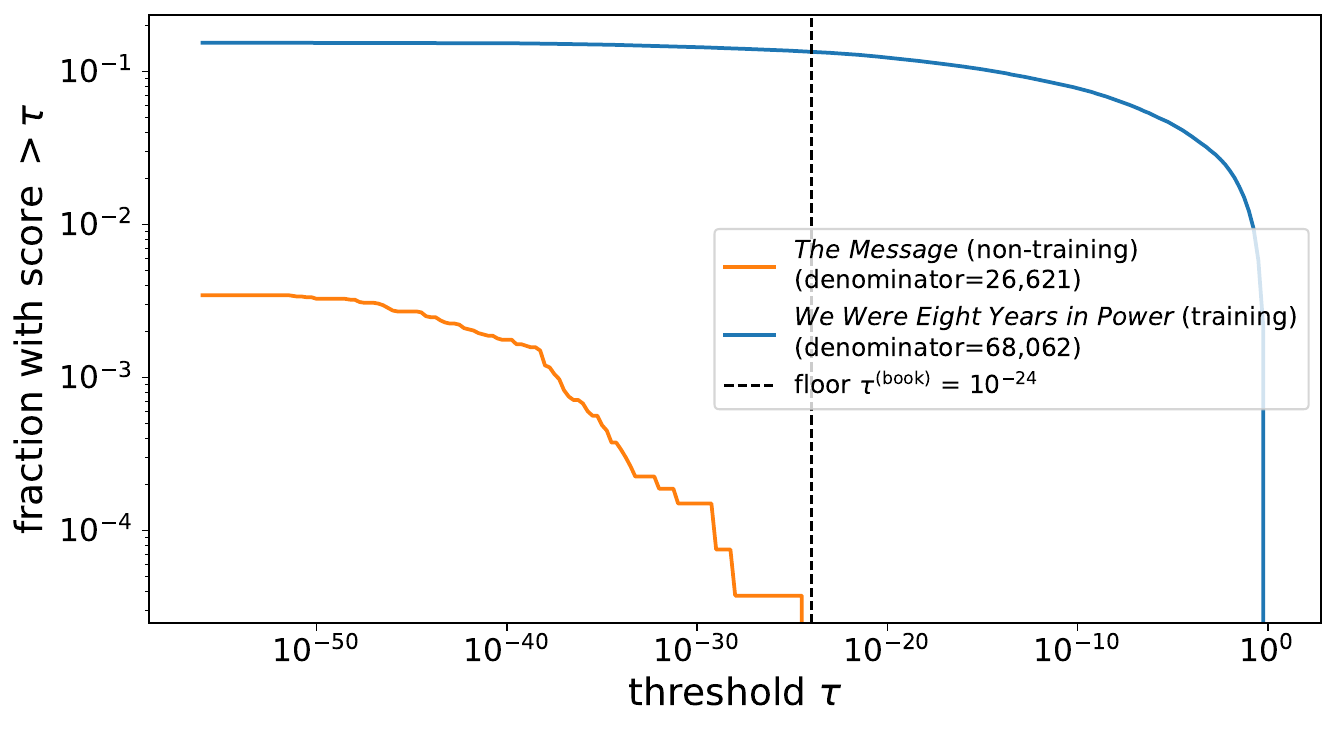}
  \caption{Ta-Nehisi Coates: \emph{The Message} (non-member) and \emph{We Were Eight Years in Power} (member).}
\end{subfigure}

\medskip

\begin{subfigure}{\linewidth}
  \centering
  \makebox[0.49\linewidth]{\small\bf before decontamination}\hfill
  \makebox[0.49\linewidth]{\small\bf after decontamination}\\[1pt]
  \includegraphics[width=0.49\linewidth]{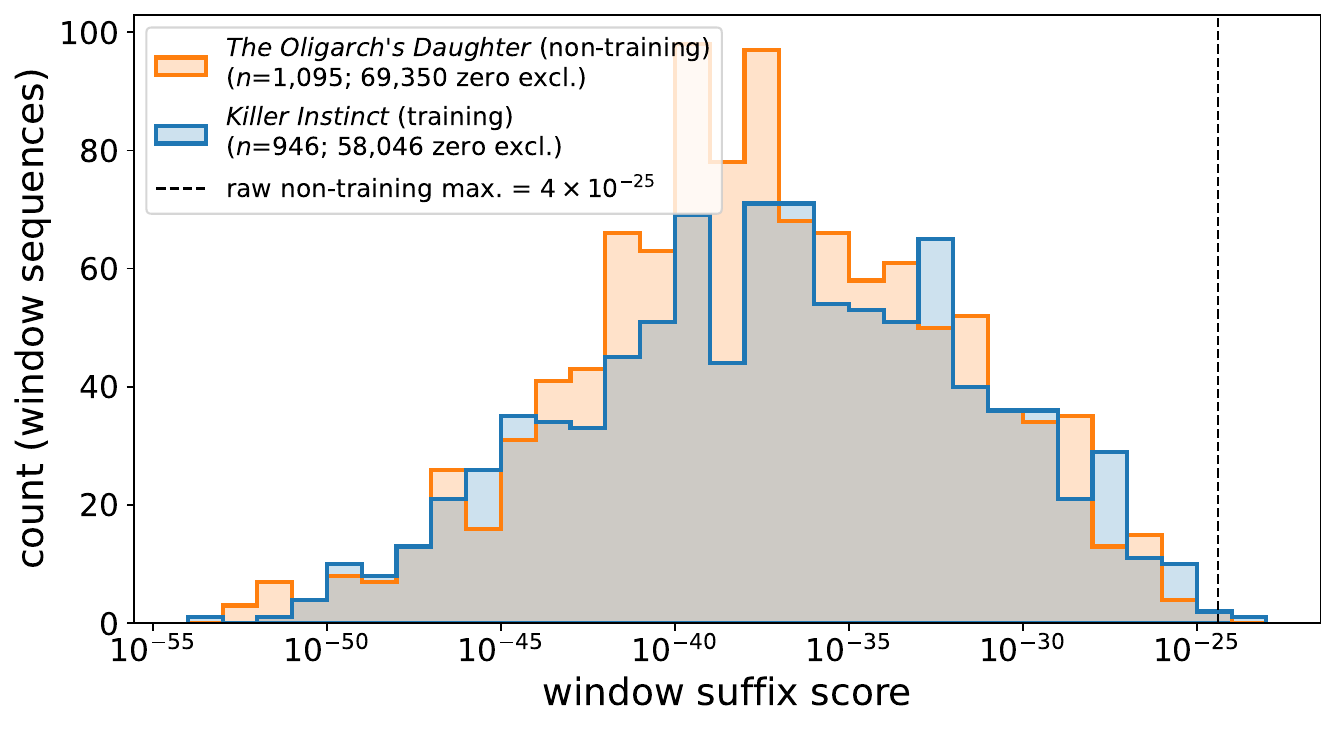}\hfill
  \includegraphics[width=0.49\linewidth]{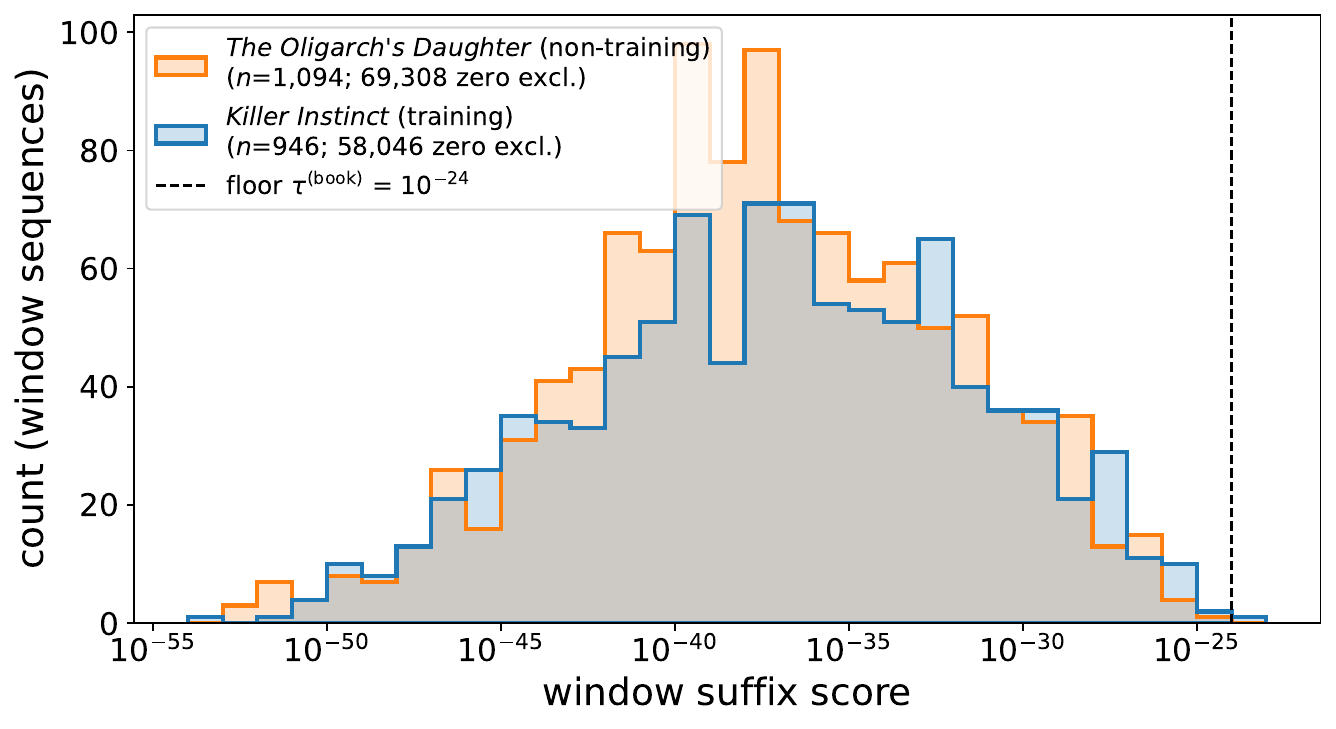}\\[2pt]
  \includegraphics[width=0.49\linewidth]{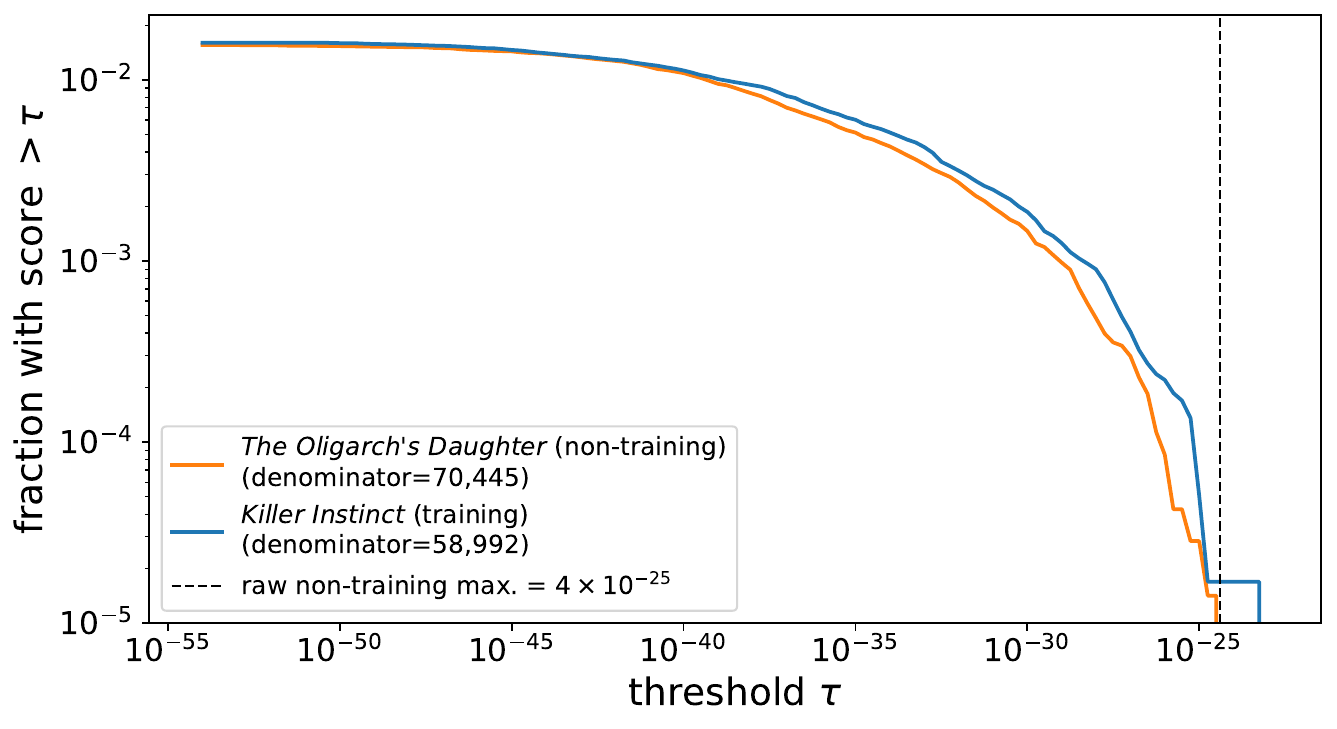}\hfill
  \includegraphics[width=0.49\linewidth]{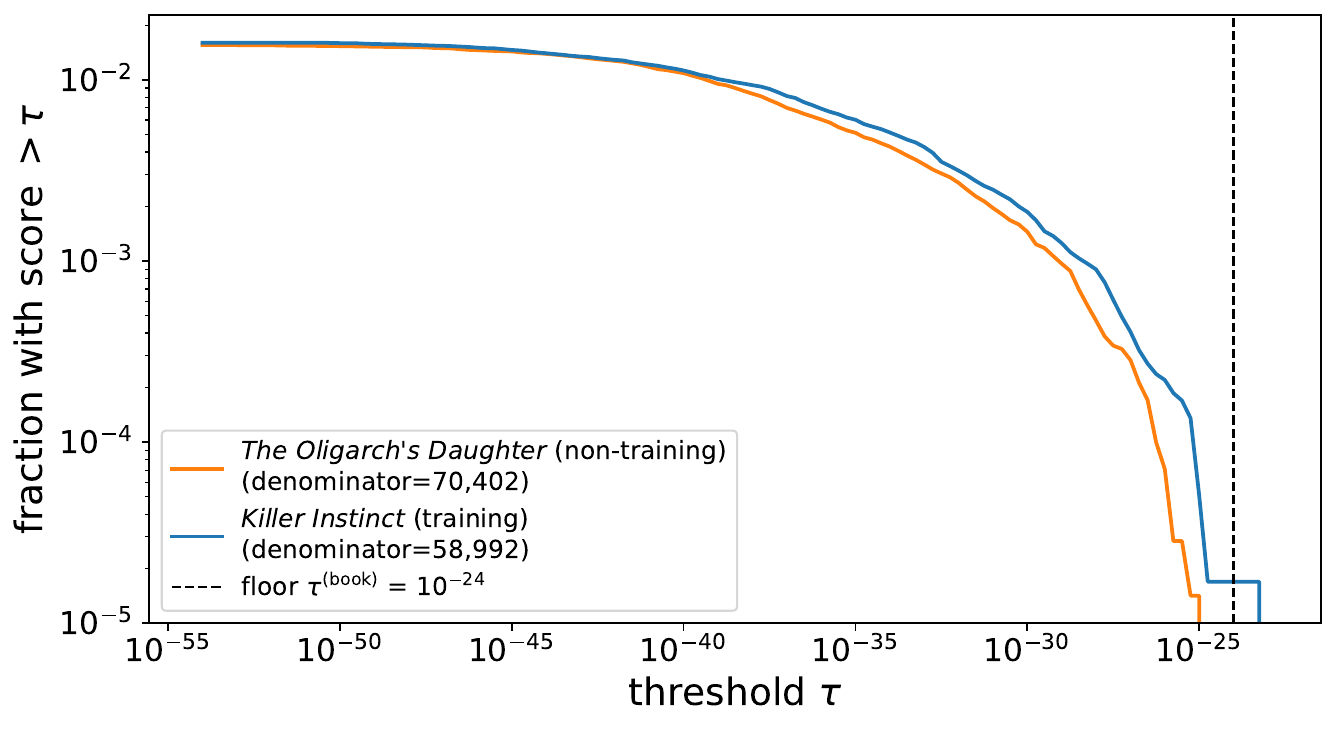}
  \caption{Joseph Finder: \emph{The Oligarch's Daughter} (non-member) and \emph{Killer Instinct} (member).}
\end{subfigure}
\caption{\textbf{Matched-control decontamination for the Coates and Finder pairs.}
The top row within each pair is the score histogram and the bottom row the complementary CDF, before and after verified reused passages are removed.
Note that the member and non-member distributions are indistinguishable for Finder books.}
\label{app:fig:decontam-coates-finder}
\end{figure}

Fig.~\ref{app:fig:decontam-brown-collins} shows how the non-member distributions shift after decontamination for Dan Brown and Suzanne Collins;
Fig.~\ref{app:fig:decontam-coates-finder} shows the same results for Ta-Nehisi Coates and Joseph Finder. 

\clearpage
\subsection{Additional results}\label{app:census:results}

\begin{figure}[h!]
  \centering
  \includegraphics[width=\linewidth]{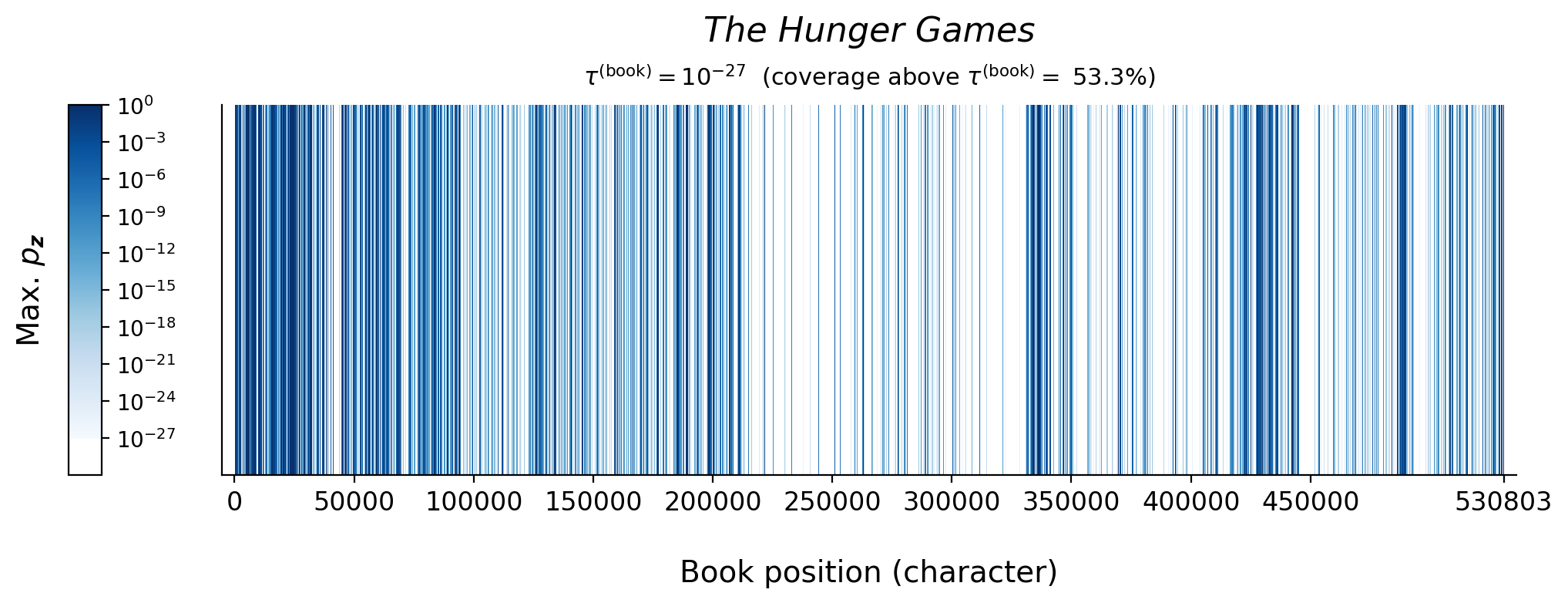}
  \caption{\textbf{Memorization across \emph{The Hunger Games}.}
  Each character is colored by the largest $50$-token suffix score among the overlapping windows whose suffix covers it.
  The minimum probability that counts as evidence of memorization comes from calibrating against the matched control book, \emph{Sunrise on the Reaping}.}
  \label{app:fig:heatmap-hunger-games}
\end{figure}

\paragraph{Where the signal occurs and how far it clears the floor.}
Beyond reporting coverage above the threshold, following \citet{cooper2025books}, we can plot where memorization occurs within a book using a heatmap.
Fig.~\ref{app:fig:heatmap-hunger-games} shows an example of this type of result for \emph{The Hunger Games} and Llama 3.1 70B, using verbatim probabilistic extraction, where the minimum probability that counts as extraction is just above the calibrated threshold.

\paragraph{Margin.}
In the main paper, we report results using coverage CCDFs. 
This is because the calibrated threshold makes a binary claim, but the strength of evidence above it varies.
That way, we can visualize how much coverage there is above the calibrated threshold, as well as how coverage changes as we increase probability above that threshold.
We quantify this with the margin (Eq.~\ref{app:eq:margin}), instantiated at each book's own calibrated threshold:
\begin{equation}
\label{app:eq:margin-book}
\Delta(\candseq;\tau^{(\mathrm{book})})
=
\log_{10}\frac{\score(\candseq)}{\tau^{(\mathrm{book})}}.
\end{equation}
Positive $\Delta$ means $\candseq$ clears the threshold; larger values place it farther from the strongest matched non-member that sets the calibrated threshold.
Fig.~\ref{fig:book-evidence-strength} in the main paper marks the coverage five orders of magnitude above each book's own $\tau^{(\mathrm{book})}$, which is coverage at $\Delta=5$.

\begin{figure}[b]
  \centering
  \begin{minipage}[c]{0.62\linewidth}
    \centering
    \includegraphics[width=\linewidth]{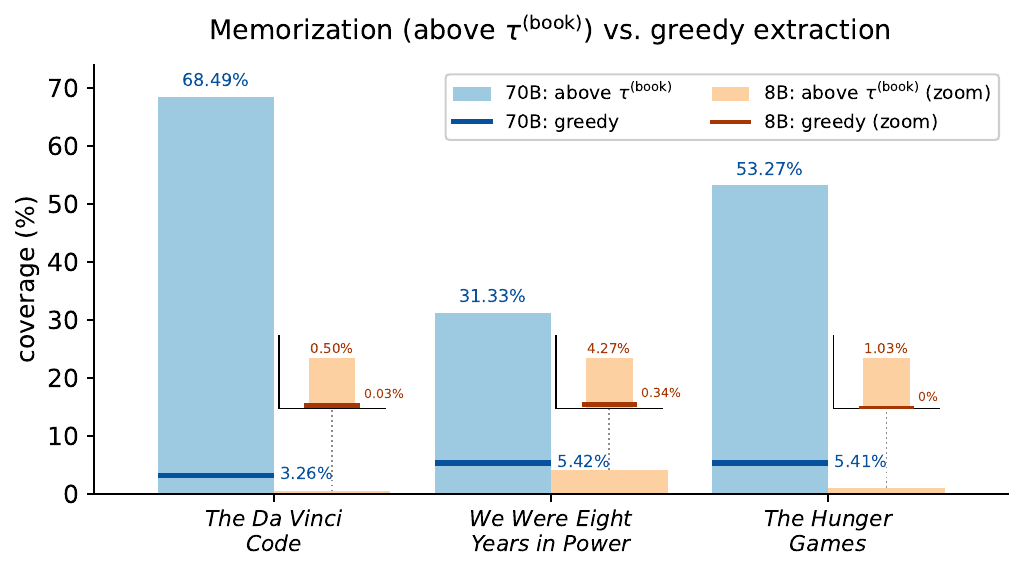}
  \end{minipage}\hfill
  \begin{minipage}[c]{0.35\linewidth}
    \centering
    \setlength{\tabcolsep}{2.5pt}
    \renewcommand{\arraystretch}{1.08}
    \begin{tabular}{@{}lrr@{}}
      \toprule
      \textbf{Training book} & \multicolumn{2}{c}{\textbf{$\tau^{(\mathrm{book})}$}} \\
      \cmidrule(lr){2-3}
       & \textbf{70B} & \textbf{8B} \\
      \midrule
      \emph{Da Vinci Code} & $10^{-21}$ & $10^{-22}$ \\
      \emph{Eight Years} & $10^{-24}$ & $10^{-36}$ \\
      \emph{Hunger Games} & $10^{-27}$ & $10^{-31}$ \\
      \emph{Killer Instinct} & $10^{-24}$ & $10^{-26}$ \\
      \bottomrule
    \end{tabular}
  \end{minipage}
  \caption{\textbf{Single-document census results.}
  For each book, we plot coverage for Llama 3.1 70B and 8B. The bars show probabilistic extraction coverage above the respective calibrated threshold for each pair. 
  We use lines on the bars to show greedy coverage for the same book and model. 
  \emph{Killer Instinct} is not plotted on the left because it carries effectively no memorization signal.}
  \label{app:fig:books-coverage}
\end{figure}

%% file: section/iclr/appendix/500-extractability.tex
\clearpage
\section{Extractability at a finite query budget}\label{app:extractability}

This appendix provides additional details on experiments concerning extractable memorization (\S\ref{sec:extractable-memorization}).
To start, we fully state the derivation of the sampling-budget accounting. 

\subsection{Query budget}\label{app:extractability:query}

Fix a prompt $\promptseq$ and let
\[
p_\candseq\coloneqq\channel(\candseq\mid\promptseq)
\]
be the per-query generation probability.
Across $n$ independent samples, the probability of generating $\candseq$ at least once is
\begin{equation}
\label{app:eq:at-least-one}
\Pr_{\mathrm{gen}}(n)=1-(1-p_\candseq)^n.
\end{equation}
For a target success probability $c = 1-\delta \in (0,1)$, the required integer budget is
\begin{equation}
\label{app:eq:budget-from-probability}
n(p_\candseq;c)
=
\left\lceil
\frac{\log(1-c)}{\log(1-p_\candseq)}
\right\rceil.
\end{equation}
Conversely, the smallest per-query probability that reaches confidence $c$ within a fixed budget $n$ is
\begin{equation}
\label{app:eq:probability-from-budget}
p_\candseq(n;c)=1-(1-c)^{1/n}.
\end{equation}
For small $p_\candseq$, $\log(1-p_\candseq)\approx-p_\candseq$, giving
\[
n(p_\candseq;c)\approx\frac{-\log(1-c)}{p_\candseq}.
\]
At $c=0.95$, the numerator is approximately $2.996$, so reaching $95\%$ confidence costs roughly three times the inverse per-query probability.
This is where the $3\times10^{21}$ samples number comes from in our discussion in \S\ref{sec:extractable-memorization}. 

In the remainder of this appendix, we provide budget views (analogous to Fig.~\ref{fig:davinci-realistic-extraction}) for both population conformal-test experiments and census experiments.
In all, we show results for $c=95\%$ (i.e., $\delta=5\%$). 
We then show additional results on deterministic realistic extraction of low-probability memorized sequences.

\subsection{Population budget views}\label{app:extractability:population}

We show two combined budget views for OLMo 2 models on Wikipedia. 
Both plots show rates at $1-\delta=c=95\%$. 

\begin{figure}[!ht]
\centering
\begin{subfigure}{\linewidth}
  \centering
  \includegraphics[width=\linewidth]{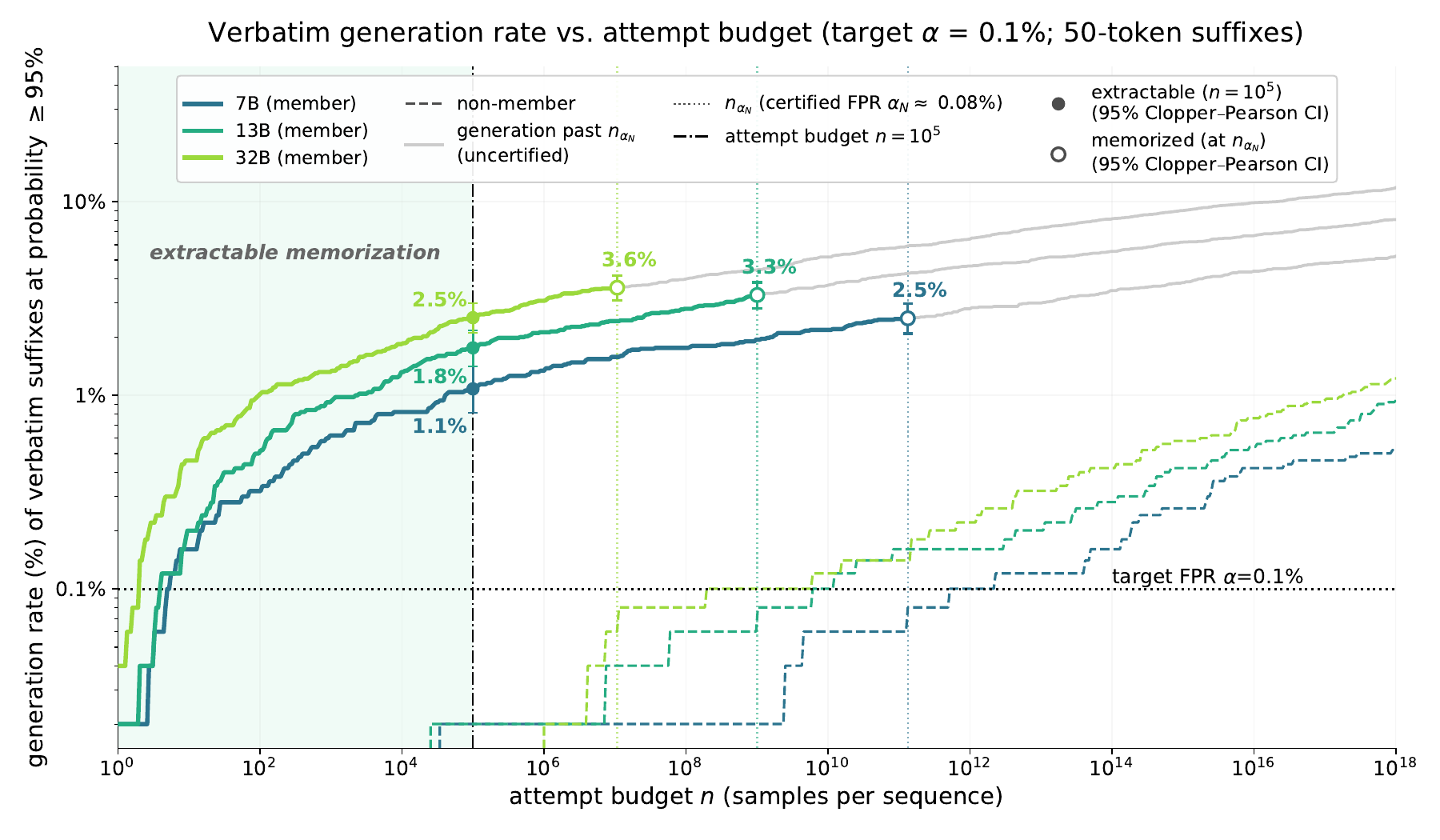}
  \caption{Member and non-member generation rates reachable at a given budget (certified $\alpha_N\approx0.0802\%$)}
  \label{app:fig:olmo-budget-raw}
\end{subfigure}

\smallskip

\begin{subfigure}{\linewidth}
  \centering
  \includegraphics[width=\linewidth]{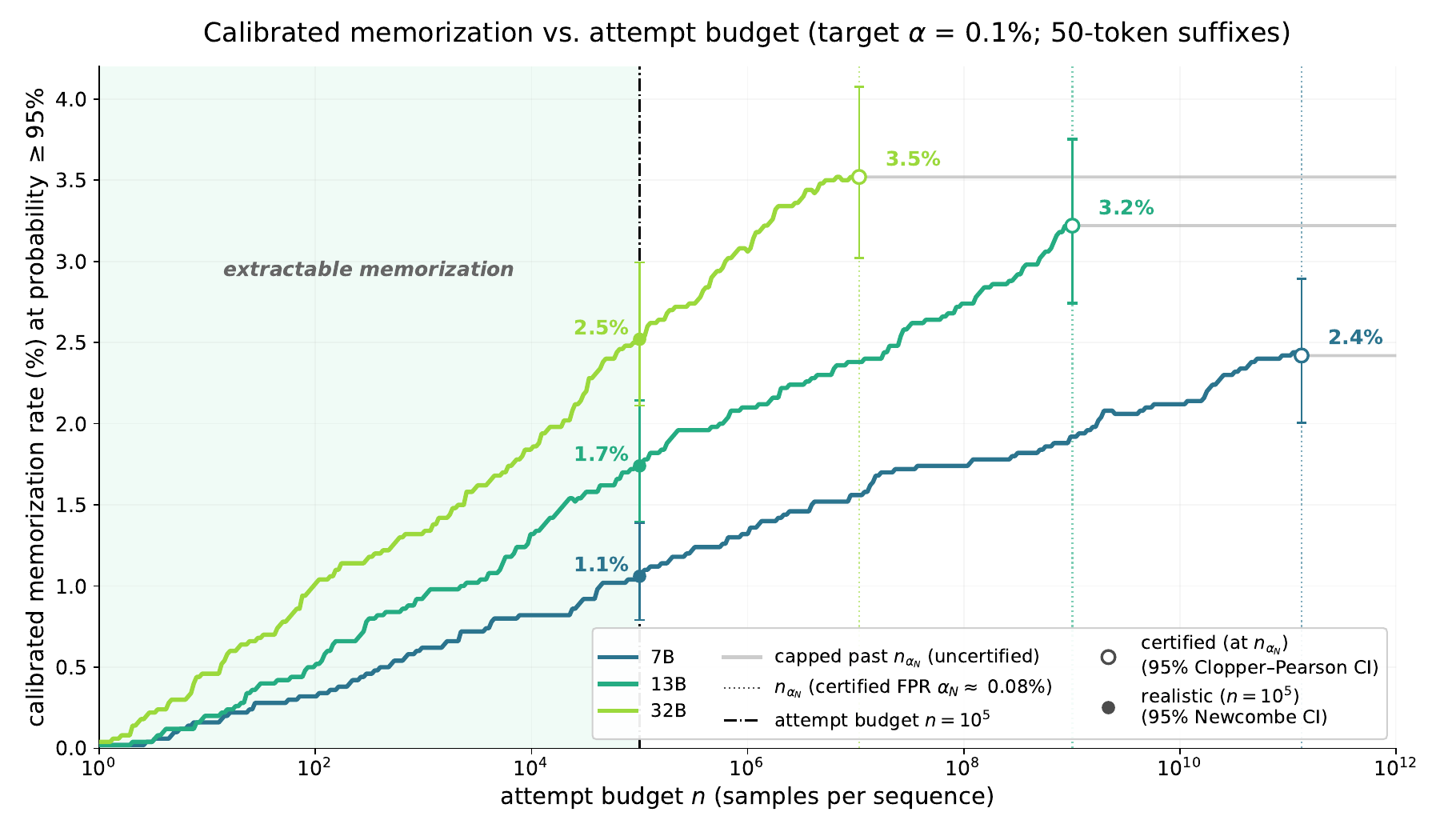}
  \caption{Calibrated memorization rate reachable at a given budget, at target FPR $\alpha=0.1\%$ (certified $\alpha_N\approx0.0802\%$)}
  \label{app:fig:olmo-budget-cmr}
\end{subfigure}
\caption{\textbf{Two budget-level views of OLMo~2 on Wikipedia.}
Both plots use $50$-token prefixes, $50$-token suffixes, and require at least $95\%$ probability of one successful generation within the stated number of independent samples.}
\label{app:fig:olmo-budget}
\end{figure}

In Fig.~\ref{app:fig:olmo-budget-raw}, we show the budget view for 7B, 13B, and 32B for the raw member and non-member generation rates.
The $y$-axis shows these rates, and so we can treat the non-member generation rate as the FPR.
We indicate a reference line for where the budget crosses a target FPR of $\alpha=0.1\%$;
the level the calibrated threshold actually certifies is $\alpha_N = 4/4989 \approx 0.0802\%$, just below that target $\alpha$.
These budget figures calibrate on the decontaminated control pool ($N=4988$; Fig.~\ref{app:fig:population-contamination}), which gives $K=\lfloor 0.001\times 4989\rfloor=4$.
The results in the main paper use the full pool ($N=5000$), for which $\alpha_N = 5/5001 \approx 0.09998\%$.
The mint-green shaded region marks budgets up to $n=10^5$, our illustrative realistic budget.
For each model, the hollow dot marks the member generation rate at $n_{\alpha_N}=\ceil{\log(0.05)/\log(1-\tau_{\alpha_N})}$ (Eq.~\ref{app:eq:budget-from-probability}):
the budget at which a sequence whose per-query probability equals the calibrated threshold is generated at least once with probability $95\%$.
The filled dot marks the rate at the realistic extraction budget $n=10^5$.
Both dots carry exact Clopper--Pearson intervals.
At a fixed budget, whether a sequence clears that budget is determined by its own per-query probability, so it's not a random event;
the only randomness is which sequences we sampled.
The count of sampled sequences that clear it is therefore binomial, and the Clopper--Pearson interval is exact for binomial proportions.

In Fig.~\ref{app:fig:olmo-budget-cmr}, we show the corresponding calibrated memorization rates (Eq.~\ref{eq:emp-cmr}) for the results in Fig.~\ref{app:fig:olmo-budget-raw}, where both rates count only sequences that score above $\tau_{\alpha_N}$.
Note that the two sets of annotated dots have different intervals because they subtract different objects.
At $n_{\alpha_N}$, the subtracted term is the certified level $\alpha_N$;
this is a constant that carries no sampling variance, so the interval is the member Clopper--Pearson interval shifted down by $\alpha_N$ (the same width, just translated down).
In contrast, at $n=10^5$, the subtracted amount is the measured non-member rate at that budget.
That rate is an empirical proportion over a finite control sample, so a different draw of controls would have produced a different value.
Both terms in the difference therefore carry sampling variance, so the interval has to account for both.
Newcombe's independent-proportions interval takes a Wilson score interval for each rate and combines them into a single interval for the difference.

\clearpage
\subsection{Census budget views}\label{app:extractability:census}

\begin{figure}[!t]
\centering
\begin{subfigure}{0.92\linewidth}
  \centering
  \includegraphics[width=\linewidth]{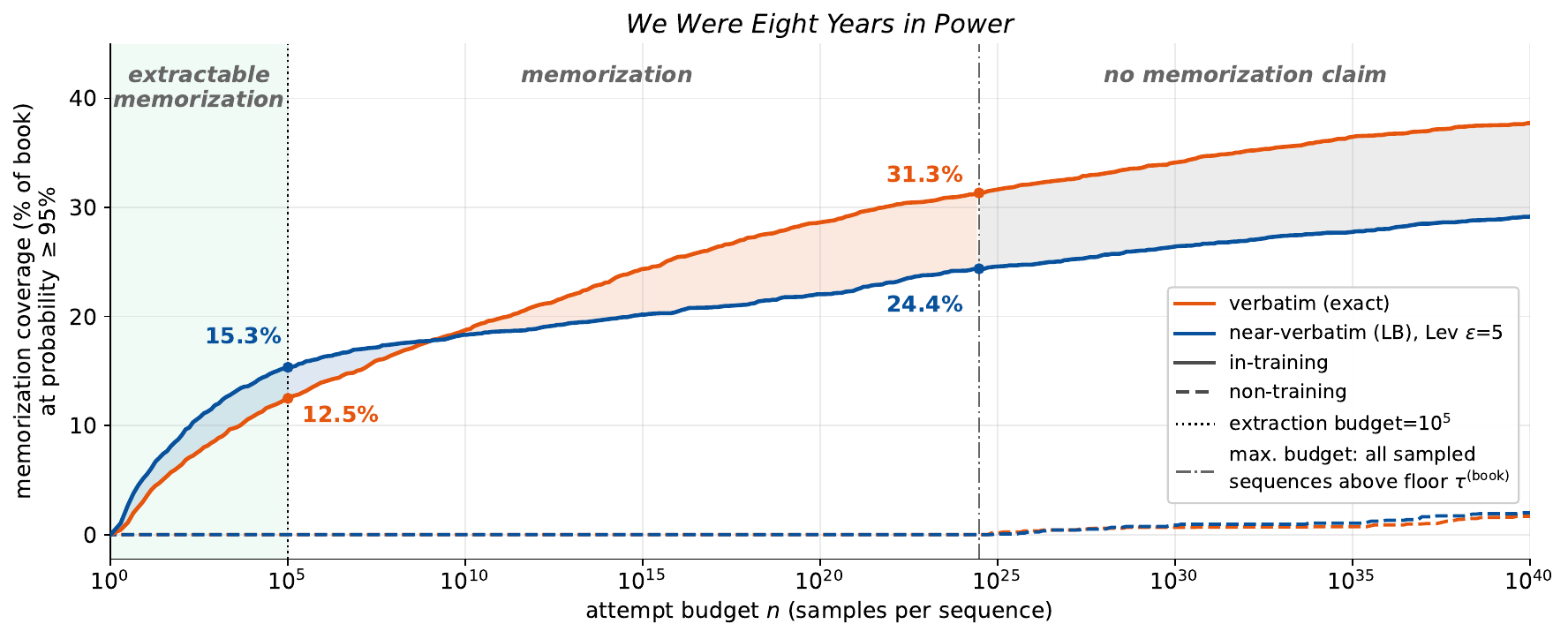}
  \caption{\emph{We Were Eight Years in Power}; verbatim and near-verbatim use $\tau^{(\mathrm{book})}=10^{-24}$.}
\end{subfigure}

\smallskip

\begin{subfigure}{0.92\linewidth}
  \centering
  \includegraphics[width=\linewidth]{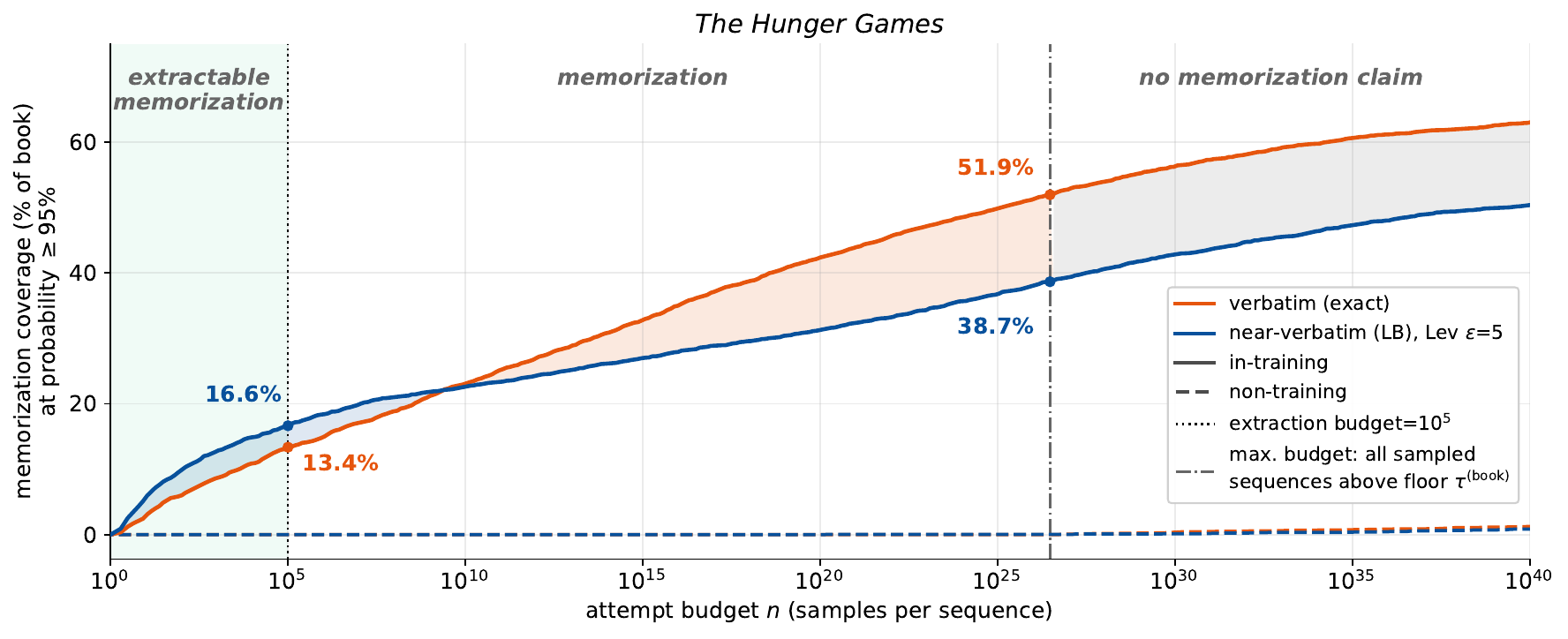}
  \caption{\emph{The Hunger Games}; the plot uses the more conservative near-verbatim $\tau^{(\mathrm{book})}=10^{-26}$.}
\end{subfigure}
\caption{\textbf{Coverage against attempt budget for two additional Llama~3.1 70B book censuses.}
The non-member reference curves rise above zero only after the budget associated with the matched threshold, $\tau^{(\mathrm{book})}$.
We calibrate $\tau^{(\mathrm{book})}$ separately for each extraction procedure, so it can produce different results for verbatim and near-verbatim for the same pair of books.}
\label{app:fig:additional-book-budgets}
\end{figure}

We provide results for \emph{We Were Eight Years in Power} and \emph{The Hunger Games} in Fig.~\ref{app:fig:additional-book-budgets}.
These complement the results for \emph{The Da Vinci Code} in Fig.~\ref{fig:davinci-realistic-extraction}
We again omit results for \emph{Killer Instinct}, since we detect effectively no memorization in our experiments for that book.
For near-verbatim probabilistic extraction, we use the same sliding-window procedure to produce prompts and candidates and the same decoding policy (top-$k$, $k=40$) as for the verbatim results;
what differs is the scoring, which uses the efficient lower-bound algorithm from \citet{cooper2026nv}, with Levenshtein-distance pruning enabled with $\varepsilon=5$.

\clearpage
\subsection{Deterministic recovery}\label{app:extractability:deterministic}

Random sampling is not the only way to decode outputs.
\citet{cooper2026nv} use a beam-search-based algorithm to return a deterministic pool of $800$ continuations per candidate, with cost comparable to approximately $20$ samples.
For these experiments, we run the baseline algorithm, which performs no pruning to guide the beam search to within the $\varepsilon$-ball of the candidate sequence, and we post-process the results to filter to those within Levenshtein distance of $5$ tokens from the candidate. 
That is, we check the returned pool for verbatim and near-verbatim recovery, then compare the recovered sequences with what can be sampled inside the illustrative budget.\looseness=-1

\begin{figure}[!h]
\centering
\begin{minipage}[c]{0.58\linewidth}
  \centering
  \includegraphics[width=\linewidth]{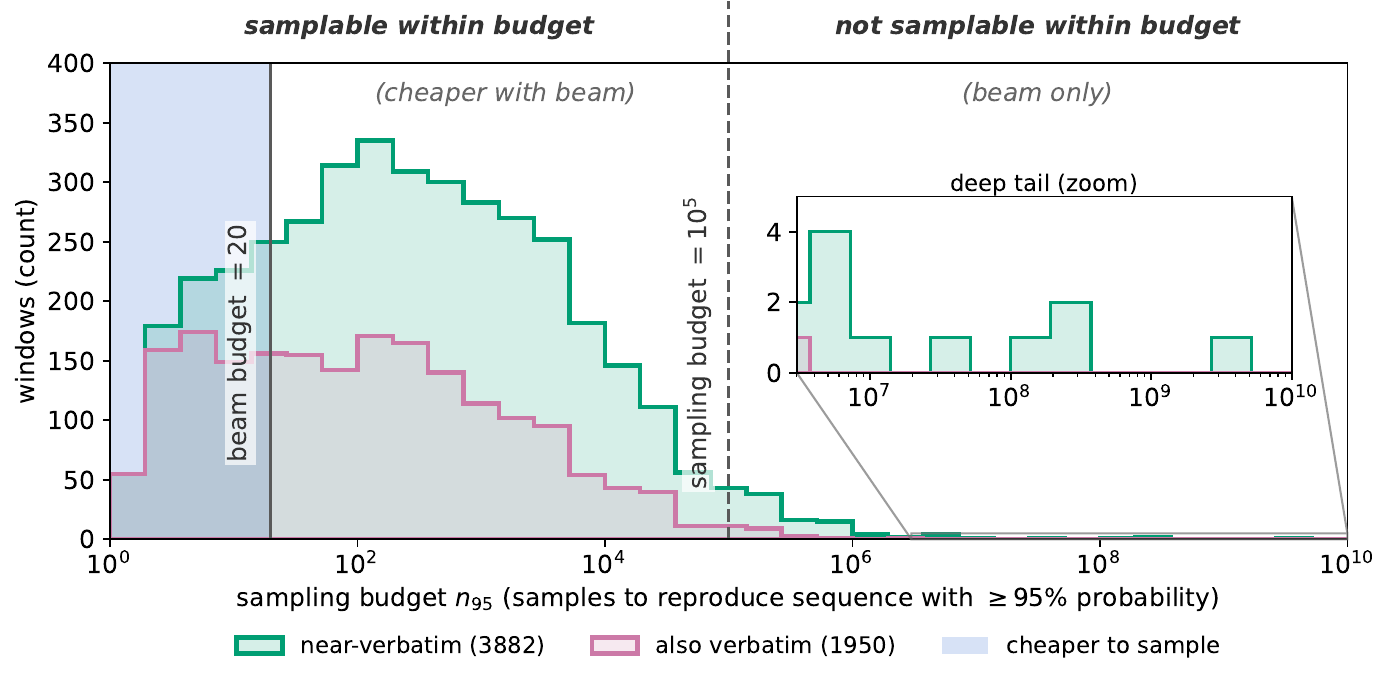}
\end{minipage}
\hfill
\begin{minipage}[c]{0.39\linewidth}
  \centering
  \footnotesize
  \setlength{\tabcolsep}{2pt}
  \renewcommand{\arraystretch}{1.08}
  \begin{tabular}{@{}lrr@{}}
  \toprule
  & \multicolumn{2}{c}{\textbf{Coverage (\%)}}\\
  \cmidrule(lr){2-3}
  \textbf{Recovered in} & \textbf{verbatim} & \textbf{near-verbatim}\\
  \midrule
  top-$1$ decode & 6.1 & 13.6\\
  top-$20$ beams & 10.1 & 16.9\\
  top-$800$ pool & 10.4 & 17.1\\
  \addlinespace
  \shortstack[l]{not samplable\\within $10^5$} & 1.8 & 2.3\\
  \bottomrule
  \end{tabular}
\end{minipage}
\vspace{-.25cm}
\caption{\textbf{Deterministic recovery for \emph{The Da Vinci Code}.}
The near-verbatim lower-bound algorithm surfaces sequences within a computational budget comparable to roughly $20$ independent samples; some of the recovered memorized content wouldn't be sampled within $10^5$ attempts.
Verbatim recovery requires the exact target, while near-verbatim requires only a sequence within Levenshtein distance of $5$ tokens of it (near-verbatim recovery subsumes verbatim recovery). 
``Cheaper to sample'' means the sequence's probability is high enough that random sampling reaches it in fewer draws than the roughly $20$ passes the beam-search algorithm costs.
The table reports coverage for different subsets of the pool of sequences returned by the algorithm. 
For instance, we get $6.1\%$ verbatim coverage from looking only at the top-$1$ sequence returned in the $800$-sequence pool for each candidate.
The last row shows the coverage that sampling would not reach within $10^5$ attempts.\looseness=-1}
\label{app:fig:deterministic-recovery}
\vspace{-.2cm}
\end{figure}

\begin{figure}[!h]
\vspace{-.3cm}
  \centering
  \includegraphics[width=0.65\linewidth]{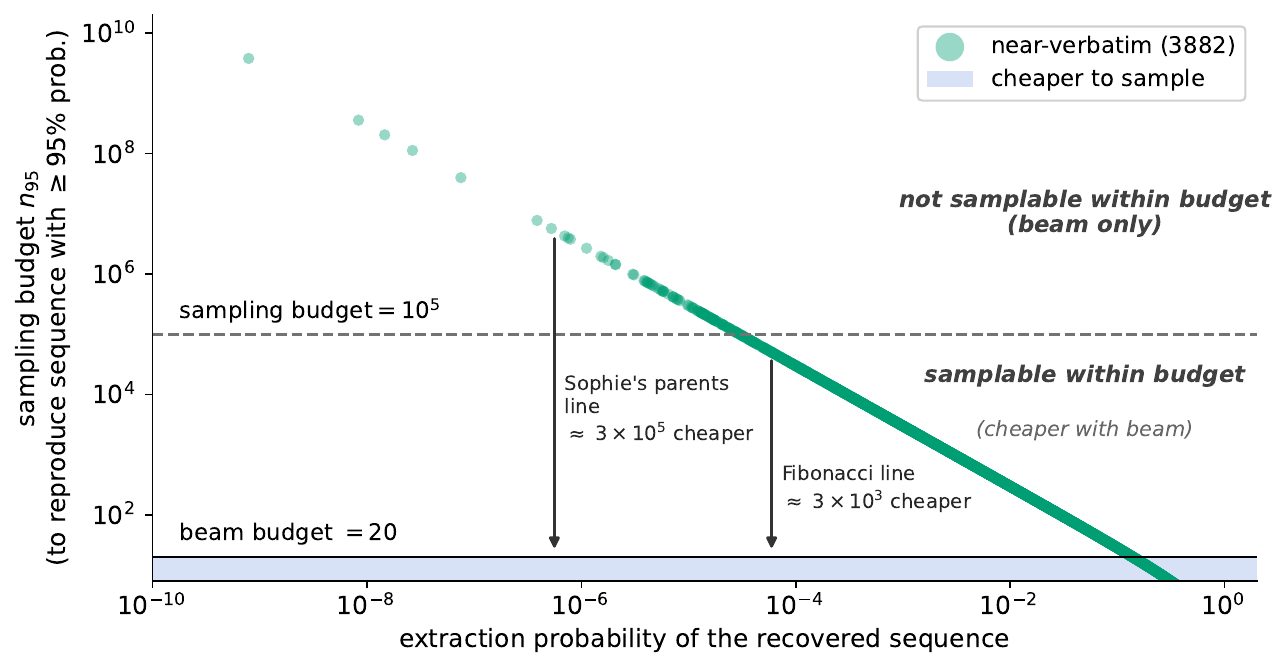}
  \vspace{-.25cm}
  \caption{\textbf{Sampling cost vs.\ deterministic near-verbatim recovery.}
  Comparing the budget required to sample a memorized sequence from \emph{The Da Vinci Code} with the cost of surfacing it using the beam-search procedure.
  The two annotated sequences show how much cheaper it is to recover them with this algorithm.
  The plotted points use the lower-bound near-verbatim $\varepsilon$-ball probability.\looseness=-1}
  \label{app:fig:beam-cost}
\end{figure}

Fig.~\ref{app:fig:deterministic-recovery} shows how many recovered windows sit at each sampling budget.
The $x$-axis is $n_{95}$, the number of random samples needed to reproduce a sequence with probability $95\%$, and the $y$-axis is the count. 
The near-verbatim results show every window the algorithm recovered within Levenshtein distance of $5$, and the verbatim results are the subset that also recovered the exact target (both series are plotted at the near-verbatim probability, so verbatim nests inside near-verbatim in every bin).
Counts are per sliding window rather than per distinct passage. 
The gray-blue region marks budgets below the beam's own cost of roughly $20$ passes, where random sampling is the cheaper way to obtain a sequence.
Fig.~\ref{app:fig:beam-cost} provides an alternate view of the same results, limited to near-verbatim extraction.
Each dot is one recovered window, placed at its extraction probability and the sampling budget $n_{95}$ that probability implies;
the flat line is the beam's fixed cost, so the vertical drop from a dot down to that line is how much cheaper the deterministic procedure was at extraction.

%% file: section/iclr/appendix/550-assumptions.tex
\clearpage
\section{Assumptions used throughout the paper}\label{app:assumptions}

This appendix collects the assumptions behind our results in one place.
Everything here is stated elsewhere in the paper, in the context of where it's used.
We organize and repeat everything here for readability.
For each assumption, we say what it requires for the conformal test in our population experiments (\S\ref{sec:test}, \S\ref{sec:experiments:population}) and what (if anything) it requires for the census (\S\ref{sec:experiments:census}).

\custompar{Matched controls}
\begin{itemize}[leftmargin=0.5cm]
\item \emph{Conformal test.}
The conformal test assumes that the candidate's score and the control scores are exchangeable under the matched null (\S\ref{sec:test}, \S\ref{app:conformal-proof}).
That is, a non-member candidate's score looks like one more draw from the control scores' distribution.
This requires that the controls be selected by the same rule that would select the candidate if it were a non-member;
that they share the candidate's source, format, length, and other score-relevant features;
and that they actually be non-members (\S\ref{app:sec:matching}).
Exchangeability doesn't require the controls to be independent of one another or of the candidate;
our experiments draw the controls i.i.d., which implies exchangeability, and the Beta law and the intervals discussed below additionally require independence (\S\ref{app:conformal-proof}).
Matching is a best effort, and \S\ref{app:sec:matching} discusses its practical limits;
contamination of the controls (i.e., the control sample actually contains members) can only make the test more conservative (\S\ref{app:sec:population:contamination}).

\item \emph{Census.}
We run experiments on pairs of books.
The member book comes from Books3, and is known to be in Llama~3.1's training data.
The matched control is a post-cutoff book by the same author;
matching on author is our best effort at exchangeability (\S\ref{sec:experiments:census}).
Contamination has to be handled differently here, because $\tau^{(\mathrm{book})}$ is set via the single maximum control score.
One highly memorized, contaminated passage in the control book (e.g., a very popular quote) can make $\tau^{(\mathrm{book})}$ close to $1$, so we decontaminate the control's high-scoring tail.
We don't cull the in-training book, since we want to take a census that covers the entire book (\S\ref{app:census:calibration}).
\end{itemize}

\custompar{A fixed procedure}
\begin{itemize}[leftmargin=0.5cm]
\item \emph{Conformal test.}
The target FPR $\alpha$ and the complete extraction procedure (model $\params$, decoding policy $\dec$, prompt-selection rule $\promptrule$, score $\score$, and the form of $\verify$) are fixed before any control or candidate score is observed; the threshold $\tau_{\alpha_N}$ is set from the controls (\S\ref{sec:test}, \S\ref{app:conformal-proof}).
\item \emph{Census.}
The same requirement of a fixed procedure holds.
In our experiments, we use two procedures:
in one set, the sliding-window verbatim probabilistic extraction of \citet{cooper2025books} under top-$k$ decoding ($k=40$);
in another, the same sliding-window procedure, scored with the lower-bound algorithm for near-verbatim probabilistic extraction from \citet{cooper2026nv} (also with $k=40$).
Both are applied identically to the member book and its matched control (\S\ref{sec:experiments:census}, \S\ref{app:census}, \S\ref{app:extractability:census}).
However, the census differs from the conformal test with respect to what gets calibrated from the control and what is claimed.
No population is sampled, so there's no $\alpha$ and no FPR to certify.
Instead, the control book sets $\tau^{(\mathrm{book})}$, the measured maximum on the decontaminated control scores.
As a result, by construction, there are no false positives among the control windows.
But this shouldn't be confused with a certified FPR of $0$, since there is no notion of an FPR here (\S\ref{app:census:calibration}).
\end{itemize}

\custompar{Ties and atoms}
\begin{itemize}[leftmargin=0.5cm]
\item \emph{Conformal test.}
Prop.~\ref{prop:conformal-validity} doesn't require the score to be continuous.
If the $N+1$ scores don't have ties, the marginal FPR equals $\alpha_N$ exactly;
ties can only make the test conservative (\S\ref{sec:test}, \S\ref{app:conformal-proof}).
Continuity matters for the Beta law of the population FPR $\alpha^*$ (Prop.~\ref{prop:alpha-star-law}), which also assumes that the scores are i.i.d.
Our population experiments draw the controls i.i.d.\ (\S\ref{app:conformal-proof}).
Some of the decoding policies we use produce atoms:
under top-$k$, a candidate that isn't reachable from the top-$k$ tokens at every step has score exactly $0$;
greedy decoding results in a binary score (\S\ref{app:sec:population:policy}).
Lemma~\ref{lem:alpha-star-atom} extends the Beta law to scores with atoms, including the atom at $0$ that top-$k$ decoding produces (\S\ref{app:sec:conformal:alpha-star:atoms}).
In our experiments, the nonzero control scores are all distinct (\S\ref{app:sec:conformal:alpha-star:practice}).
Under greedy decoding, $\alpha^*$ isn't Beta-distributed and we don't draw the Beta band (\S\ref{app:sec:conformal:alpha-star:practice}).
\item \emph{Census.}
Not applicable.
$\tau^{(\mathrm{book})}$ is a rounded-up maximum, not a certified level (\S\ref{app:census:calibration}), so there's no FPR for ties to affect (and, accordingly, no Beta law).
\end{itemize}

\custompar{Finite controls}
\begin{itemize}[leftmargin=0.5cm]
\item \emph{Conformal test.}
$N$ controls can certify FPRs only in steps of $1/(N+1)$: $\alpha_N=K/(N+1)$ with $K=\lfloor\alpha(N+1)\rfloor$ (Eq.~\ref{eq:alpha-n}).
The smallest nonzero certified FPR is $1/(N+1)$, reached at $K=1$, where the threshold is the largest control score;
setting $\alpha<1/(N+1)$ gives $K=0$ and the test can never reject (\S\ref{sec:test}, \S\ref{app:conformal-proof}).
\item \emph{Census.}
Not applicable, since there's no notion of a certified FPR (see above).
\end{itemize}

\custompar{Marginal guarantee}
\begin{itemize}[leftmargin=0.5cm]
\item \emph{Conformal test.}
The guarantee of Prop.~\ref{prop:conformal-validity} is marginal: it averages over control draws (\S\ref{app:sec:conformal:alpha-star:conditional}).
For the specific control sample we draw, the calibrated threshold has a population FPR $\alpha^*$ that changes with the draw, and the marginal guarantee doesn't say how far that $\alpha^*$ sits from $\alpha_N$ (\S\ref{app:sec:conformal:alpha-star}).
Saying anything about it requires the i.i.d.\ assumption above, under which $\alpha^*\sim\mathrm{Beta}(K,N-K+1)$ with mean $\alpha_N$ (Prop.~\ref{prop:alpha-star-law});
Cor.~\ref{cor:alpha-star-conditional} turns this into a guarantee conditional on the drawn control sample.
Our population figures have two sources of randomness:
the member sample and the control draw.
The Clopper--Pearson intervals in our figures cover the randomness in the member sample, holding the calibrated threshold fixed;
they deliberately leave out the randomness in the control draw, since the plotted quantity is defined with the constant $\alpha_N$, not the unknown $\alpha^*$ (\S\ref{app:sec:population:intervals}).
For the second source of randomness, the Beta law gives the distribution of $\alpha^*$ across control draws, which is fixed before any control is drawn and isn't a confidence interval (\S\ref{app:sec:conformal:notation}).
\item \emph{Census.}
Not applicable, since there's no notion of a certified FPR (see above).
\end{itemize}

\custompar{Many candidates, one control sample}
\begin{itemize}[leftmargin=0.5cm]
\item \emph{Conformal test.}
The guarantee of Prop.~\ref{prop:conformal-validity} is per candidate:
for each candidate, the test either claims memorization or makes no claim, and nothing in the guarantee requires the candidates to be independent of one another (\S\ref{app:sec:conformal:multiplicity}).
One control sample can therefore serve to make calibrated claims for many candidates, and each claim stands on its own.
Reusing the sample does make the claims dependent on one another, since every candidate is tested against the same threshold $\tau_{\alpha_N}$.
For a continuous score, the outcomes for two non-member candidates drawn independently of each other and of the controls are positively correlated, at exactly $1/(N+2)$~\citep{bates2021testing}.
The per-candidate guarantee doesn't by itself bound the familywise error rate or the false discovery rate over many candidates (\S\ref{sec:test});
\S\ref{app:sec:conformal:multiplicity} discusses both, and we don't study either bound in this work.
\item \emph{Census.}
Every window of the member book is compared against the same $\tau^{(\mathrm{book})}$, and the windows overlap heavily (\S\ref{sec:experiments:census}).
Coverage is the fraction of the book's characters spanned by windows above $\tau^{(\mathrm{book})}$ (Eq.~\ref{eq:coverage}).
\end{itemize}

\custompar{Intervals}
\begin{itemize}[leftmargin=0.5cm]
\item \emph{Conformal test.}
In our Wikipedia experiments, we draw candidates and controls uniformly at random within their date ranges, so each is an i.i.d.\ sample (\S\ref{sec:experiments:population}).
The member rate above a threshold is then a binomial proportion, and the intervals we plot on it are exact Clopper--Pearson intervals (\S\ref{app:sec:population:intervals}).
The calibrated rate subtracts the constant $\alpha_N$, so its interval is the Clopper--Pearson interval translated down by $\alpha_N$, with the same width.
We instead use Newcombe's independent-proportions interval when the subtracted quantity is estimated from the controls (the greedy in-sample floor, or the non-member rate at a fixed budget);
for the same candidates scored by two models, we use Newcombe's paired-proportions interval.
All plotted intervals are $95\%$ and pointwise:
each is a statement about one plotted point, not about a curve as a whole.
\item \emph{Census.}
Every window of each book is scored, so the coverage we report is the exact value for that book rather than an estimate, and none of our census results carries an interval (\S\ref{app:census}).
\end{itemize}

\custompar{Budget views}
\begin{itemize}[leftmargin=0.5cm]
\item \emph{Conformal test.}
The budget views convert a per-query generation probability into the number of attempts needed to generate the candidate at least once with probability $c$, which we set to $95\%$ throughout (Eq.~\ref{app:eq:at-least-one}; \S\ref{app:extractability:query}).
For a fixed prompt, each attempt is an independent sample from the decoding policy, so the budget follows directly from $p=\score(\candseq)$.
The population budget $n_{\alpha_N}$ is computed from the calibrated threshold $\tau_{\alpha_N}$, so it inherits the assumptions above (\S\ref{app:extractability:population}).
\item \emph{Census.}
The census budget views use $\tau^{(\mathrm{book})}$ in the same way, calibrated separately for each extraction procedure, so the verbatim and near-verbatim results for the same book pair can use different thresholds (\S\ref{app:extractability:census}).
\end{itemize}

%% file: section/iclr/appendix/600-prior-work.tex
\section{Extended discussion of prior work}\label{app:prior-work}

This appendix provides an extended discussion of relevant prior work. 
We first discuss in greater detail how existing extraction methods are instances of our definition of an extraction procedure (Def.~\ref{app:def:extraction-procedure}; \S\ref{app:sec:prelim:prior}).
We then revisit two well-cited definitions of extraction (\S\ref{app:prior-work:matched}) and examine the validity of memorization claims in some recent work (\S\ref{app:prior-work:validity}) in light of matched comparisons.
Finally, we describe similarities and differences between our conformal test for memorization and membership inference attacks (\S\ref{app:prior-work:mia}).

\input{section/iclr/appendix/610-prior-methods}
\input{section/iclr/appendix/620-prior-defs}
\input{section/iclr/appendix/630-prior-validity}
\input{section/iclr/appendix/640-prior-mia}

%% file: section/iclr/appendix/610-prior-methods.tex
\subsection{Prior work on extraction, in these terms}\label{app:sec:prelim:prior}

Existing extraction methods are instances of Def.~\ref{app:def:extraction-procedure}.
For verbatim \newterm{discoverable extraction}~\citep{lee2022dedup,carlini2023quantifying},  $\params$ is an open-weight non-instruction-tuned model, $\dec$ is greedy decoding, and $\promptrule$ returns a singleton:
$\promptseq$ (Eq.~\ref{app:eq:prompt-rule}) is the (typically $50$-token) natural prefix of $\candseq$.
The model is prompted with $\promptseq$ and greedily generates an $\eventlen$-token continuation $\outseq$.
Since $q_{\params,\greedy}(\cdot \mid \promptseq) \in \{0,1\}$, the score reduces to the binary match indicator $\score(\candseq) = \1[\outseq = \candseq]$ (Eq.~\ref{app:eq:candidate-score}), and $\verify$ deems $\candseq$ memorized when it equals $1$.
The relaxation of discoverable extraction, \newterm{probabilistic (discoverable) extraction}~\citep{hayes2025measuringmemorizationlanguagemodels}, follows the same setup, but makes $\dec$ stochastic (e.g., top-$k$) and replaces the binary score with $\score(\candseq) = \channel(\candseq \mid \promptseq)$ (Eq.~\ref{app:eq:generation-prob}), computed exactly using teacher forcing rather than estimated with sampling.
\citet{hayes2025measuringmemorizationlanguagemodels} leave $\verify$ fairly informal; 
as a sanity check, they run one experiment to show that the same model generates unseen test data at rates orders of magnitude lower than training data. 
By contrast, \citet{cooper2025books} run a suite of experiments that score non-training data; 
these experiments determine a conservative global threshold $\tau_{\min}$, where $\verify$ evaluates $\1[\channel(\candseq \mid \promptseq) \geq \tau_{\min}]$.\looseness=-1

Both discoverable extraction and its probabilistic relaxation  extend naturally to the near-verbatim setting, in which 
$\verify$ considers extraction successful when there is a match to any sequence in the $\tol$-ball $\ball{\tol}{\candseq}$ (Eq.~\ref{eq:ball}).
For near-verbatim discoverable extraction, the score is $\1[\distance(\outseq, \candseq) \leq \tol]$, and $\verify$ deems memorization when it equals $1$~\citep{ippolito-etal-2023-preventing}. 
For near-verbatim probabilistic extraction, the near-verbatim extraction probability is the total mass on the $\tol$-ball around $\candseq$, 
\(
\channel(\ball{\tol}{\candseq} \mid \promptseq) 
\).
The set of continuations within a $\tol$-ball around $\candseq$ is combinatorially large, so there is no teacher-forced analogue like there is for the verbatim computation (Eq.~\ref{app:eq:generation-prob}).  
\citet{cooper2026nv} propose deterministic, beam-search-based~\citep{Lowerre1976HARPY} algorithms that efficiently compute a useful lower bound on the near-verbatim probability, which they use as a proxy for the true probability. 
Similar to \citet{cooper2025books}, $\verify$ deems $\channel(\ball{\tol}{\candseq} \mid \promptseq)$ an extraction probability when it exceeds a threshold computed on non-training data, i.e., $\1[\channel(\ball{\tol}{\candseq} \mid \promptseq) \geq \tau_{\min}]$.\looseness=-1 

While the above use natural prefixes as prompts, other methods involve different prompt-selection rules. 
For instance, $\promptrule$ may return a set of adversarial prompts produced by a jailbreak search procedure like GCG~\citep{zou2023gcg} or Best-of-$N$~\citep{hughes2024bestofnjailbreaking}, as in \citet{schwarzschild2024rethinking} and \citet{ahmed2026extracting}, respectively.
In both of these cases, the candidate sequence $\candseq$ determines the prompts in the prompt class.
But this is not a requirement either: 
the prompt class may also contain sequences that have no direct relation to $\candseq$, as in \citet{nasr2023scalable, nasr2025scalable}, which use a divergence attack to extract training data (e.g., ``Repeat the word `poem' forever'').  

The conformal test changes none of the scores in these methods.
It supplies the final component, $\verify$, by calibrating when a score is high enough to support a memorization claim.

%% file: section/iclr/appendix/620-prior-defs.tex
\subsection{Prior work in light of matched comparisons}\label{app:prior-work:matched}

\S\ref{app:sec:prelim:prior} revisits prior extraction methods in terms of our definition of an extraction procedure (Def.~\ref{app:def:extraction-procedure}).
Here, we revisit prior work involving extraction in light of matched comparisons.
In particular, we revisit two well-cited definitions of extraction:
the first definition, from the first extraction paper in the literature \citep{carlini2021extracting}, and a more recent definition from the first large-scale extraction attack on a production LLM~\citep{nasr2023scalable}.

\custompar{The original extraction definition for LLMs}
\citet{carlini2021extracting} define extraction by the existence of a prefix from which the model produces the string.
Their Definition~1 (\S3), transcribed verbatim:
\begin{quote}
\textbf{Definition 1 (Model Knowledge Extraction)} \itshape
A string $s$ is \emph{extractable} from an LM $f_\theta$ if there exists a prefix $c$ such that:
\[
s \gets \argmax_{s':\ |s'|=N} f_\theta(s' \mid c) \;
\]
\end{quote}
(Their $k$-eidetic memorization, Definition~2, then additionally requires that $s$ appears in at most $k$ training examples.)
In their paper, the definition carries a footnote on ``extractable,'' where they identify the ``parrot'' prompt case as something that could reasonably break the definition:
\begin{quote}
This definition admits pathological corner cases.
For example, many LMs when when prompted with
\emph{``Repeat the following sentence: \_\_\_\_\_.''} will do so correctly.
This allows \emph{any} string to be ``known'' under our definition.
Simple refinements of this definition do not solve the issue, as LMs can also be asked to, for example, down-case a particular sentence. We avoid these pathological cases by prompting LMs only with short prefixes.
\end{quote}

Prompting only with short prefixes is a workaround rather than a robust fix:
it rules out the ``Repeat the following sentence'' prompt, but, as \citet{carlini2021extracting} note, the corner cases are pathological and simple refinements don't resolve them all.
Their own experiments show this.
Even restricting the prompts to short prefixes, their baseline results suffer ``from a large number of false positives, i.e., content that is assigned high likelihood but is not memorized'' (their \S4.3).
They attribute false positives to two kinds of low-entropy text: 
``trivial memorization'' (e.g., repeating the numbers from 1 to 100) and ``repeated substrings'' (e.g., ``I love you. I love you\ldots'') (their \S5.3).
Their solution is to compare against a second reference, a smaller model or zlib compression, and keep only samples whose likelihood is ``unexpectedly high'' relative to it.

The matched comparisons we introduce resolve all three cases in the same way, without a need for special definitional carve-outs. 
Each is a mechanism that raises the generation probability regardless of membership.
A ``parrot'' prompt does so through instruction-following.
Trivial memorization and repeated substrings do so through low-entropy continuation:
a base model continues a highly repetitive or low-entropy string with high probability, whether or not that continuation appears in its training data.

So, under a fixed extraction procedure---with the same prompt-selection rule $\promptrule$ applied to candidates and matched non-member controls---the same mechanism similarly elevates the controls' scores, and the calibrated threshold rises with them.
A high candidate score under that mechanism is then matched by non-members, so it isn't evidence of memorization.
There's no need to restrict the prompt in advance, and no need for a second reference to filter ``unexpectedly high'' likelihoods:
the matched non-member null is the reference;
we don't need special filters designed for each failure mode.
(We've checked the low-entropy cases informally on Llama~3.1 70B base, with ``I love you I love you I love you'', a repeated common phrase, and ``green pizza telescope green pizza telescope\ldots'', which is repeated nonsense. 
Both are produced with high probability regardless of training-data membership, so a matched control scores as high as the candidate.
We haven't run this at scale.)

\custompar{Extraction with an unrestricted prompt}
\citet{nasr2023scalable} instead place no restriction on the prompts used to elicit extraction.
They allow for any prompt an adversary can construct without direct access to the training data.
Their Definition~1 (\S3.1):
\begin{quote}
\textbf{Definition 1 (Extractable memorization).} \itshape
Given a model with a generation routine $\mathsf{Gen}$, an example $\bm{x}$ from the training set $\mathbb{X}$ is \underline{extractably memorized} if an adversary (without access to $\mathbb{X}$) can construct a prompt $\bm{p}$ that makes the model produce $\bm{x}$ (i.e.,  $\mathsf{Gen}(\bm{p})=\bm{x}$).
\end{quote}
The generality of this definition is deliberate:
a footnote to the definition says it's meant to be ``a general definition that encompasses attacks in this work and our own,'' including prompts unrelated to the candidate, like their divergence attack (\S\ref{app:sec:prelim:prior}).
As a result, the whole family of ``parrot'' prompts satisfies this definition, as well.

Matched comparisons can similarly fix the issues in this definition, without narrowing it.
The definition can still allow any prompt the adversary constructs, so long as the same prompt-selection rule is also run on matched non-member controls.
If that rule also makes matched non-members appear ``extracted,'' then a hit on a training example isn't evidence of memorization.

%% file: section/iclr/appendix/630-prior-validity.tex
\subsection{Validity in prior work}\label{app:prior-work:validity}

As we show in this work, a memorization claim is valid when the candidate's score clears a threshold calibrated on matched non-members under the same extraction procedure (Box~\ref{box:extraction-claim:main}; \S\ref{sec:test}; \S\ref{app:sec:conformal}; \S\ref{app:census});
without running a baseline on non-members, it's impossible to know if a candidate's high score could be due to more general predictability.

Running matched comparisons like this is consistent with the field's preference for reporting memorization claims at low FPRs, even at the cost of false negatives (\S\ref{app:prior-work:mia}).
This posture is implicit in the first extraction paper in the literature, which treated false positives as a serious problem its attack had to fix~\citep{carlini2021extracting}, and was made explicit in later work on membership inference~\citep{carlini2022membership}.

Prior work has generally treated matched comparisons, when it runs them at all, as informal sanity checks (\S\ref{sec:intro})~\citep{lee2022dedup,carlini2023quantifying,hayes2025measuringmemorizationlanguagemodels,pappu2025rlhf,schwarzschild2024rethinking,cooper2025books,cooper2026nv}.
Below, we examine memorization claims in some recent work in light of matched comparisons.

\custompar{Short candidates without matched comparisons}
As our results show, a short verbatim match when generating training data shouldn't be treated as sufficient evidence for carrying a valid memorization claim on its own.
At short candidate lengths, matched non-members can also be generated by capable models at very high rates, indicating many false positives (\S\ref{sec:experiments:population}, \S\ref{app:sec:population:length}).
A lot of recent work nonetheless reports extraction using short matches, and does so without running a matched comparison~\citep{chang-etal-2023-speak,karamolegkou2023copyrightviolationslargelanguage,wei2025interrogatingllmdesignfair,liu2026alignmentwhackamolefinetuning}.
These claims rest on member generation rates, which can't separate memorization from general predictability on their own (\S\ref{sec:takeaways}).
\citet{liu2026alignmentwhackamolefinetuning} is a recent example.
Its headline metric, book memorization coverage ($\mathrm{bmc}@5$), counts every run of \emph{five} or more consecutive words as evidence for successful extraction, where a ``word'' is an NLTK (Natural Language Toolkit) \texttt{wordpunct} token in their released code.
Five NLTK words is far below the field's standard minimum of $50$ tokens (approximately $37$ readable words)~\citep{carlini2023quantifying,nasr2023scalable,nasr2025scalable,hayes2025measuringmemorizationlanguagemodels}.
\citet{cooper2026whack} discusses this in detail.

\custompar{Sanity checks}
Some work does run negative controls, but not ones that match the claims being made.
\citet{schwarzschild2024rethinking} validate ACR with what they call ``sanity checks that we hope any useful definition of memorization to pass'':
$100$ random token strings of $3$ to $17$ tokens, and sentences from Associated Press articles published after the training cutoff, neither of which their procedure ever compresses.
On that basis, they write that ``while the GCG algorithm can elicit any output, we never observe compression of non-training data. This suggests that our method is robust against false positives.''
But these controls run only on Pythia models, which the paper discloses as a limitation, whereas the headline results run on Llama-2-7B-chat and Phi-1.5 with no control at all;\footnote{While the models for their headline results don't have publicly disclosed training datasets, enough is publicly disclosed about their training (e.g., Llama 2 models on Books3~\citep{cooper2025books}, the training cutoffs) to run on non-member sequences.
}
and neither control is matched to the famous quotes the claims are about.
A matched comparison would run the same procedure, on the same model, on matched non-training text.
We ran one check on Llama~2 13B and were successful in generating non-training data (\S\ref{app:sec:intro:acr}).\footnote{As noted in \S\ref{app:sec:intro:acr}, we identified other issues in the ACR implementation that required us to change the GCG success condition in order to reliably register success on training data, not just non-training data.
}

\custompar{Prompt leakage}
Another problem in some recent work is that the prompts used to elicit memorized training data can themselves supply words that get counted as a match in generated outputs.
For instance, nothing in GCG prevents the optimized prompt from leaking the target.
When we ran ACR, we found that the optimized prompts often contained words from the target itself, on both training targets and on the post-cutoff quotes (\S\ref{app:sec:intro:acr}).
Leakage is also potentially responsible for some of the short-span results that go into computing $\mathrm{bmc}@5$ in \citet{liu2026alignmentwhackamolefinetuning}.
Each prompt is a detailed plot summary of the excerpt under test;
such prompts supply the passage's characters, events, their order, and distinctive words and short phrases.
The authors note this, and add a trimming step that removes contiguous five-word overlaps between an output and its prompt.
But it's possible that short five-word matches in outputs could be reassembled from words the prompt's summary supplied, even if the prompt doesn't contain those short matches contiguously.
This is similar to what we observe for the non-training Jalen Brunson quote in our ACR control (\S\ref{app:sec:intro:acr}).
This type of leakage would survive the paper's trimming procedure.
The paper doesn't run controls on non-training data to test for potential false positives (either with respect to counting very short spans or prompt leakage).
\citet{cooper2026whack} discusses this in detail.

\custompar{Validity issues in claims in the opposite direction}
The above addresses work that risks over-claiming memorization, due to omitting matched comparisons to assess potential false positives.
Other recent work runs the other direction;
it claims that extraction is unreliable evidence of memorization, because models can verbatim complete text they weren't explicitly trained on~\citep{liu2025forged,morris2025howmuch}.
In their experiments, neither \citet{liu2025forged} nor \citet{morris2025howmuch} removes near-duplicates of training data from the sequences it treats as non-members;
such near-duplicates are functionally members of the training data (Eq.~\ref{eq:membership-indicator}, \citet{lee2022dedup}), so a verbatim completion (especially of famous passages, like quotes from the U.S. Constitution~\citep{liu2025forged}) quite reasonably reflects memorization rather than being a counterexample to the validity of extraction (\S\ref{app:sec:matching}).
And regardless of this issue, the conclusion is too strong:
valid extraction claims are always relative to a specific extraction procedure and baseline, so a negative result under one procedure says nothing about calibrated claims under another (\S\ref{sec:takeaways}).

In each of these cases, the fix is the same:
when running extraction experiments, one needs to run a matched comparison to make a valid claim (Box~\ref{box:extraction-claim:main}).
That is what makes a memorization claim valid at the low FPR the field prioritizes (\S\ref{app:prior-work:mia}), and it's also what the conformal test and the matched census in this paper are constructed to do.

%% file: section/iclr/appendix/640-prior-mia.tex
\subsection{Membership inference}\label{app:prior-work:mia}

Our conformal test is a \newterm{membership inference} test~\citep{shokri2017membership}.
Its null hypothesis is that the candidate $\candseq$ is a matched non-member, and rejecting the null says that $\candseq$ is a member that the model memorized (\S\ref{sec:test}).
Membership inference predates work on memorization and extraction in language models, and the strongest \newterm{membership inference attacks (MIAs)} rely on reference models~\citep{shokri2017membership,carlini2022membership}.
Reference-model MIAs ultimately ask the same overarching question as our test:
the null hypothesis is that the model under evaluation was trained without sequence $\candseq$.
The likelihood-ratio attack (LiRA) of \citet{carlini2022membership} estimates the distribution of the model's score (e.g., likelihood) on $\candseq$ under two sets of reference models---trained with $\candseq$ and trained without it---by training many models on random samples of the data distribution and fitting a Gaussian to each set's scores;
the MIA rejects the null when the likelihood ratio of the observed score under the two fitted distributions exceeds a calibrated threshold.
(The offline variant of the attack uses only the reference models trained without $\candseq$, and rejects when the observed score falls in the upper tail of that set's fitted distribution.)

\custompar{Data versus models}
Our test and reference-model MIAs differ in what the null reference is composed of.
The conformal test compares $\score(\candseq)$ against the scores of matched non-member \emph{data}, under \emph{one fixed model};
reference-model MIAs compare the score of the model under evaluation on \emph{a given sequence} $\candseq$ against the scores of reference \emph{models} trained with and without it.
Everything discussed in this subsection has to do with this difference in the null hypotheses that the two tests use to address the same question: whether the model memorized its training data.
(Membership inference in general doesn't require that the memorized training data can be \emph{generated} from the model;
that requirement is specific to extraction~\citep{carlini2021extracting}.)

\custompar{Testing known members}
In extraction experiments (including ours), the candidates are often known members.
The membership half of a rejection of the null hypothesis is then trivial, but the memorization part of the claim remains.
A score that clears the calibrated threshold is too high to be explained by general predictability, so the candidate \emph{behaves like a member}:
its score is one that matched non-members produce only at the certified FPR (\S\ref{app:sec:conformal:multiplicity}).
Members whose scores fall below the threshold are of course still members;
the test just makes no memorization claim about them.

\custompar{Exchangeability over data versus over models}
Each test rests on an exchangeability assumption, and what has to be exchangeable depends on how the null is set up.
For the conformal test, the candidate $\candseq$ must be exchangeable with the controls:
selected by the same rule and scored under the same fixed model, so that a non-member candidate's score would look like one more draw from the control scores (\S\ref{sec:test}, \S\ref{app:conformal-proof}).
For a reference-model MIA, the model under evaluation must be exchangeable, under the null, with the reference models trained without $\candseq$.
Those reference models estimate the null distribution only if the model under evaluation is one more draw from the same family of models;
otherwise, the tail probability the offline test computes isn't the FPR for that model.
(In the online variant, the reference models trained with $\candseq$ play the same role for the alternative; with the threshold set against null draws, they govern the attack's power rather than its FPR.)
This exchangeability assumption is latent in the foundational work on MIAs~\citep{carlini2022membership}, but LiRA (and other reference-model MIAs like RMIA~\citep{zarifzadeh2024rmia}) depends on it regardless.\looseness=-1

In both MIAs and our conformal test, exchangeability is a best effort in practice (\S\ref{app:sec:matching}).
But where the effort is located differs. 
For the conformal test, we choose the selection rule and draw the controls ourselves, and the model is the same fixed object for candidate and controls.
Distribution shift or contamination of the controls degrades the quality of the match (\S\ref{app:sec:matching}, \S\ref{app:sec:population:contamination}).
In reference-model MIAs, an attacker trains many models to estimate the distribution that the model under evaluation is assumed to be a draw from.
In practice, training has various sources of non-determinism~\citep{hayes2025strongmia,jebreel2026revisiting}, and not all of it is random~\citep{cooper2022nondeterminism}:
for instance, hardware non-determinism~\citep{pham2020variance,zhuang2022randomness} can compromise exchangeability.\looseness=-1

\custompar{Rank-based vs.\ parametric calibration}
The exchangeability assumption also determines what kind of calibration the null supports.
If the model under evaluation is exchangeable with $N$ reference models trained without $\candseq$, then thresholding at an empirical quantile of their scores has a finite-sample false-positive guarantee, just as the conformal test does, with a floor of order $1/N$.
\citet{ye2022enhanced}'s Attack R follows this setup: 
its threshold for $\candseq$ is the $\alpha$-percentile of $\candseq$'s loss over reference models trained on i.i.d.\ samples of the population data, which they describe as approximately guaranteeing FPR $\alpha$.
(Their construction is also where the exchangeability assumption comes closest to being stated among the papers discussed in this appendix: 
the reference models are explicitly i.i.d.\ draws, and what remains implicit is that the model under evaluation is another such draw.)
\citet{carlini2022membership} don't threshold at the empirical quantile; 
they instead fit a Gaussian to each set of reference models' scores and compute the test statistic from the fitted distributions rather than from the scores' ranks.\footnote{``By design, the exact test of Ye et al.\ can at best target an FPR of $1/N$ with $N$ shadow models. 
    It is thus inapplicable in the setting we consider here (256 shadow models, and a target FPR of 0.1\%)'' (internal citations omitted)~\citep{carlini2022membership}.
}
The Gaussian is a parametric assumption:
with $256$ reference models an empirical quantile can't resolve levels below $1/256\approx0.4\%$, but the fitted distribution has a tail at every level, so the statistic extrapolates past the data and the attack can be run at $0.1\%$ and below.
A false-positive level taken from the fitted tail is then only as good as the fit.
In contrast, the conformal test uses the $1/(N+1)$ grid that $N$ controls can certify, with marginal FPR at most $\alpha_N$ (\S\ref{sec:test}), and assumes nothing about the shape of the control distribution;
certifying a smaller floor requires more controls, not extrapolation (\S\ref{sec:experiments:population}).

\custompar{ROC curves and AUC}
MIA results are commonly reported as ROC curves and summarized by AUC~\citep{carlini2022membership}.
Our sweep (Fig.~\ref{fig:olmo-calibrated-rate}) isn't an ROC.
Even when the score can reach an FPR of $1$ (e.g., under $T=1$ sampling, where every suffix has non-zero probability; \S\ref{app:sec:population:policy}), no threshold ever certifies an FPR of $0$ (since $N$ controls certify levels only down to $1/(N+1)$; see \S\ref{sec:test}).
And there's no true-positive rate (TPR) for the memorization part of the claim, since which members are memorized isn't known (Def.~\ref{def:cmr}).
An ROC curve needs both at every operating point: 
an FPR that runs from $0$ to $1$, and a TPR to plot against.
(AUC then integrates the curve over that whole span.)

\custompar{The field's posture for making memorization claims at low FPR}
In this paper, we privilege a low FPR, even at the cost of false negatives (under-counting true instances of memorization).
This posture is inherited from foundational work on memorization. 
\citet{carlini2022membership} evaluate LiRA attacks by their true-positive rate at FPRs of $0.1\%$ and below.
\citet{carlini2022membership} also make this point about extraction directly:
\begin{quote}
to extract training data it is far more important for attacks to have a low false positive rate than high average success, as false positives are fare [sic] more costly than false negatives (internal citations omitted).
\end{quote}
It's from this lineage that we draw our paper's name~\citep{carlini2022membership}, and this conservative posture runs through both of our experimental settings.
For the population experiments, the conformal test is a hypothesis test whose calibrated threshold limits false positives to a certified level, and the low-FPR end of the sweep is where one would operate in practice (\S\ref{sec:test}, \S\ref{sec:experiments:population}).
The census has no FPR to certify (\S\ref{sec:experiments:census}), but it inherits the same conservatism:
$\tau^{(\mathrm{book})}$ is set at the largest score any control window attains.
This is also why we wrote this paper.
Recent work in the extraction literature studies settings that risk false positives, but often doesn't run the appropriate control baselines.
Without those baselines, there's no way to tell how much false positives affect the validity of memorization claims;
such work may have high false positives, which is contrary to what the field accepts when making memorization claims (\S\ref{app:prior-work:validity}).